\documentclass[lettersize,journal]{IEEEtran}
\usepackage{amsmath,amsfonts}
\usepackage{algorithmic}
\usepackage{array}
\usepackage[caption=false,font=normalsize,labelfont=sf,textfont=sf]{subfig}
\usepackage{textcomp}
\usepackage{stfloats}
\usepackage{url}
\usepackage{verbatim}
\usepackage{graphicx}
\usepackage{cite}
\usepackage{amsthm} 

\newtheorem{lemma}{Lemma}
\newtheorem{proposition}{Proposition}
\usepackage{bm}
\usepackage[colorlinks,linkcolor=black]{hyperref}
\usepackage{caption}
\usepackage{color}
\usepackage[table]{xcolor}
\usepackage{makecell}
\usepackage[ruled]{algorithm2e}
\usepackage{multirow}
\usepackage{siunitx}
\usepackage{array}
\usepackage{booktabs} 
\usepackage{multirow}
\usepackage{gensymb}
\usepackage{colortbl}
\usepackage{lineno}
\usepackage{color}
\usepackage[table]{xcolor}
\usepackage{makecell}
\usepackage[ruled]{algorithm2e}
\usepackage{cite}
\usepackage {amssymb}
\usepackage{fontawesome5}

\begin{document}

\title{Efficient and Deterministic Search Strategy Based on Residual Projections for Point Cloud Registration with Correspondences}

\author{Xinyi~Li,
        Hu~Cao,
        Yinlong~Liu\textsuperscript{\faEnvelope[regular]},
        Xueli~Liu,
        Feihu~Zhang, and~Alois~Knoll,~\IEEEmembership{Fellow,~IEEE}
\thanks{Xinyi Li, Hu Cao, and Alois Knoll are with Chair of Robotics, Artificial Intelligence and Real-time Systems, TUM School of Computation, Information and Technology, Technical University of Munich, Munich 85748, Germany (e-mail: super.xinyi@tum.de; hu.cao@tum.de; knoll@in.tum.de).}
\thanks{Yinlong Liu is with State Key Laboratory of Internet of Things for Smart City (SKL-IOTSC), University of Macau, Macau 999078, China (e-mail: YinlongLiu@um.edu.mo).}
\thanks{Xueli Liu is with Eye $\&$ Ent Hospital of Fudan University, Shanghai 200031, China (e-mail: liuxueli@fudan.edu.cn).}
\thanks{Feihu Zhang is with School of Marine Science and Technology, Northwestern Polytechnical University, Xi’an 710072, China (e-mail: feihu.zhang@nwpu.edu.cn).}
\thanks{\textsuperscript{\faEnvelope[regular]}Corresponding author: Yinlong Liu.}

}

\maketitle

\begin{abstract}
Estimating the rigid transformation between two LiDAR scans through putative 3D correspondences is a typical point cloud registration paradigm. Current 3D feature matching approaches commonly lead to numerous outlier correspondences, making outlier-robust registration techniques indispensable. Many recent studies have adopted the branch and bound (BnB) optimization framework to solve the correspondence-based point cloud registration problem globally and deterministically. Nonetheless, BnB-based methods are time-consuming to search the entire 6-dimensional parameter space, since their computational complexity is exponential to the solution domain dimension in the worst-case. To enhance algorithm efficiency, existing works attempt to decouple the 6 degrees of freedom (DOF) original problem into two 3-DOF sub-problems, thereby reducing the search space. In contrast, our approach introduces a novel pose decoupling strategy based on residual projections, decomposing the raw registration problem into three sub-problems. Subsequently, we embed interval stabbing into BnB to solve these sub-problems within a lower two-dimensional domain, resulting in efficient and deterministic registration. Moreover, our method can be adapted to address the challenging problem of simultaneous pose and registration. Through comprehensive experiments conducted on challenging synthetic and real-world datasets, we demonstrate that the proposed method outperforms state-of-the-art methods in terms of efficiency while maintaining comparable robustness.
\end{abstract}


\section{Introduction}
\IEEEPARstart{R}{igid} point cloud registration is a core and fundamental problem in the field of 3D vision and robotics with a wide range of applications, such as autonomous driving\cite{xia2021soe,10070382,im2024omni}, 3D reconstruction\cite{blais1995registering,10296850}, and simultaneous localization and mapping (SLAM)\cite{cattaneo2022lcdnet,xia2023lightweight}. Given the source and target point clouds in different coordinate systems, it aims to estimate the 6 degrees of freedom (DOF) transformation in $\mathbb{SE}(3)$ to align the two point clouds best. The 6-DOF transformation includes both 3-DOF rotation in $\mathbb{SO}(3)$ and 3-DOF translation in $\mathbb{R}^3$.

\begin{figure*}
  \centering
  \begin{tabular}{  c  c  c  c  c}
  \multirow{5}*{\rotatebox{90}{{\makecell[c]{Correspondence-based \\ (FPFH descriptor)}}}} 
    & \begin{minipage}[b]{0.22\textwidth}
		\centering
		\raisebox{-.5\height}{\includegraphics[width=0.8\linewidth]{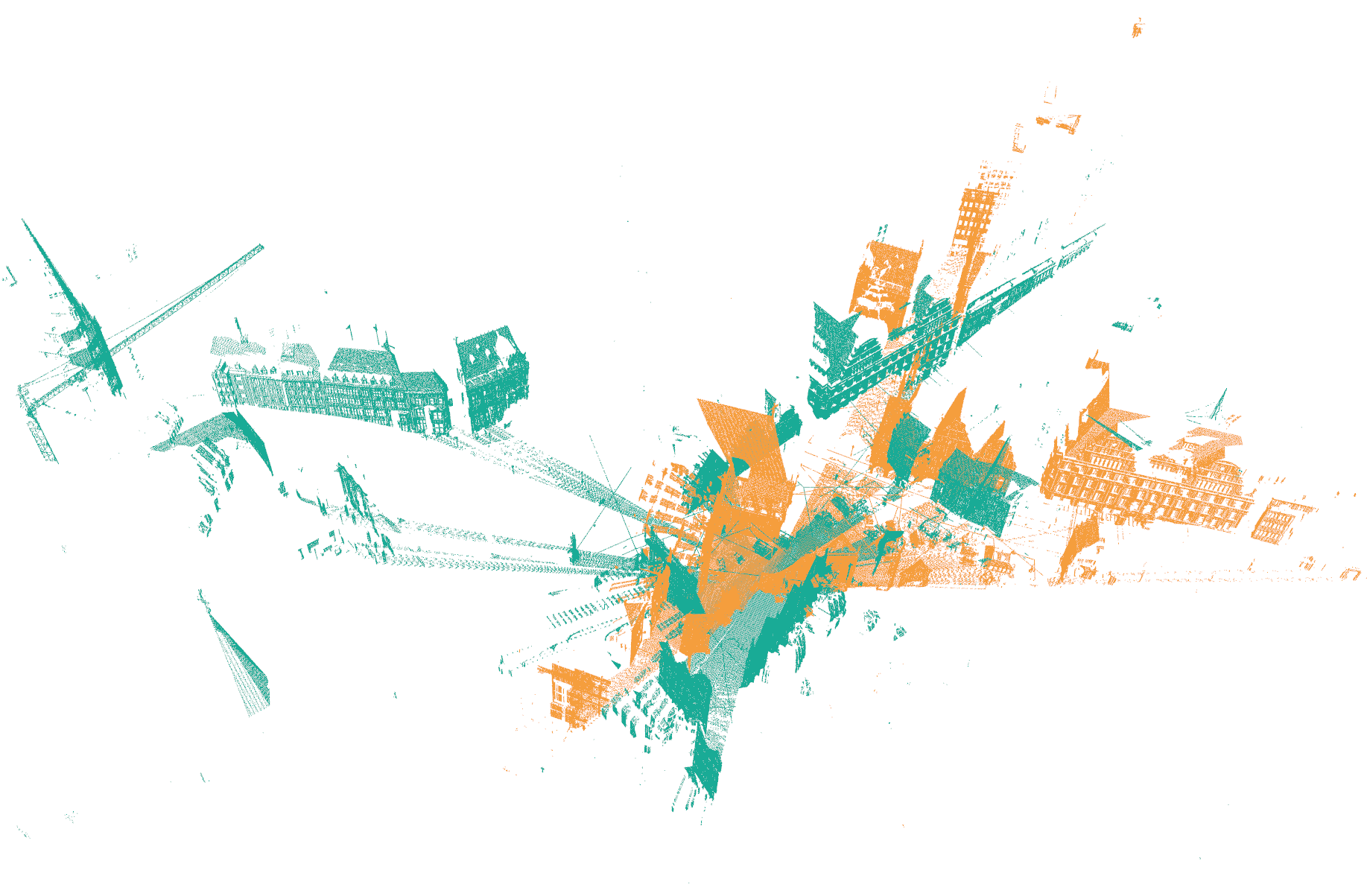}}
	\end{minipage}
    & \begin{minipage}[b]{0.22\textwidth}
		\centering
		\raisebox{-.5\height}{\includegraphics[width=0.8\linewidth]{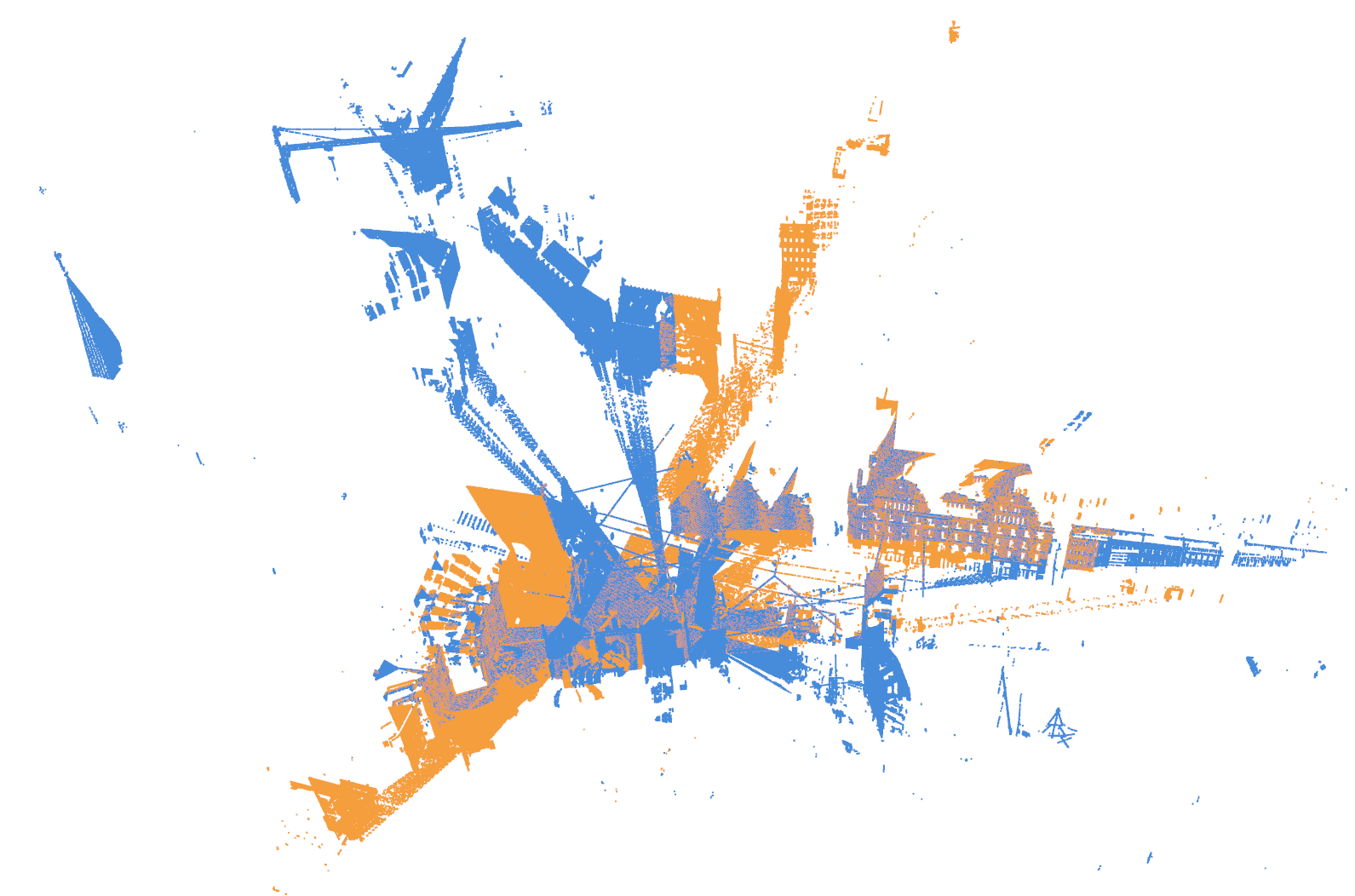}}
	\end{minipage}
    & \begin{minipage}[b]{0.22\textwidth}
		\centering
		\raisebox{-.5\height}{\includegraphics[width=0.8\linewidth]{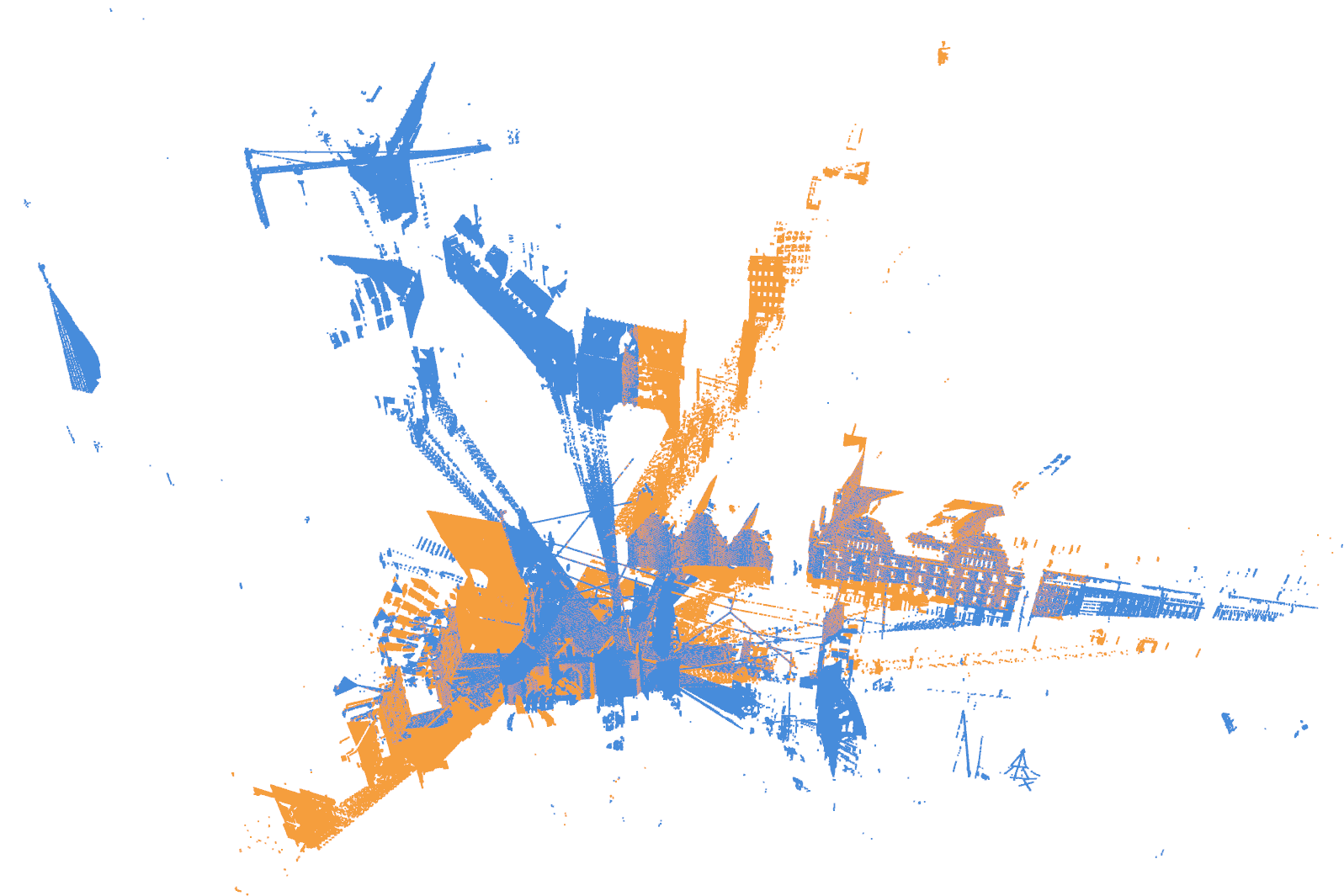}}
	\end{minipage}
    & \begin{minipage}[b]{0.22\textwidth}
		\centering
		\raisebox{-.5\height}{\includegraphics[width=0.8\linewidth]{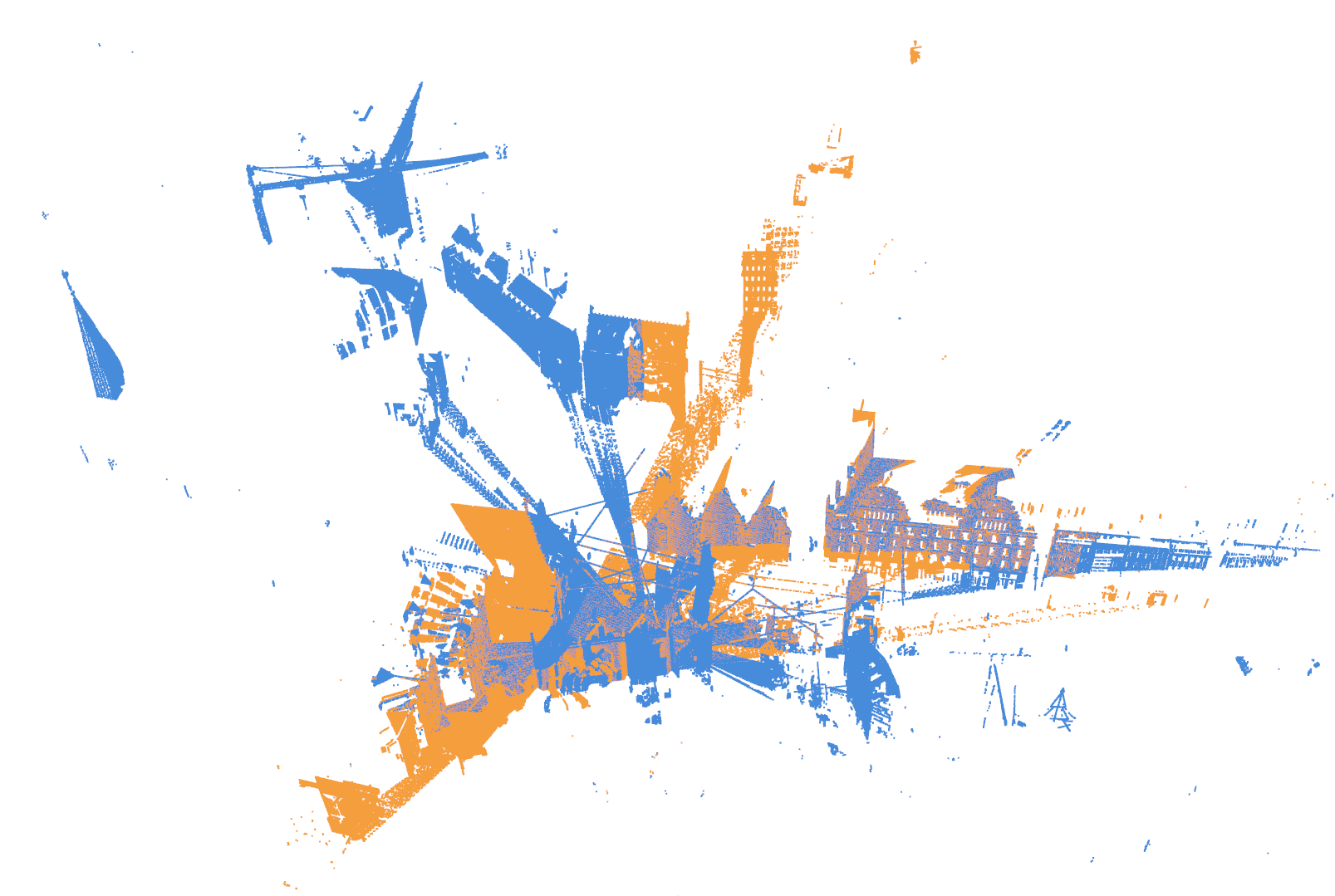}}
	\end{minipage}
    \\
    ~ & \footnotesize{(a-1) Initial ($94.20\%$ outlier)} & \footnotesize{(a-2) GORE\cite{bustos2017guaranteed}} & \footnotesize{(a-3) TEASER\cite{yang2020teaser}} & \footnotesize{(a-4) Ours}
     \\ 
    ~ & \footnotesize - & \footnotesize {$E_{\bm{R}}=0.041\degree$} & \footnotesize {$E_{\bm{R}}=0.069\degree$} & \footnotesize {$E_{\bm{R}}=\textbf{0.004}\degree$}
    \\ 
    ~ & \footnotesize - & \footnotesize {$E_{\bm{t}}=0.055\si{\metre}$} & \footnotesize {$E_{\bm{t}}=0.067\si{\metre}$} & \footnotesize {$E_{\bm{t}}=\textbf{0.042}\si{\metre}$}
    \\ 
    ~ & \footnotesize - & \footnotesize {${time}=403.1$s} & \footnotesize {${time}=1.357$s} & \footnotesize {${time}=\textbf{0.307}$s}
    \\ 
    
\multirow{5}*{\rotatebox{90}{{\makecell[c]{Correspondence-based \\ (FPFH descriptor)}}}} 
    & \begin{minipage}[b]{0.22\textwidth}
		\centering
		\raisebox{-.5\height}{\includegraphics[width=0.8\linewidth]{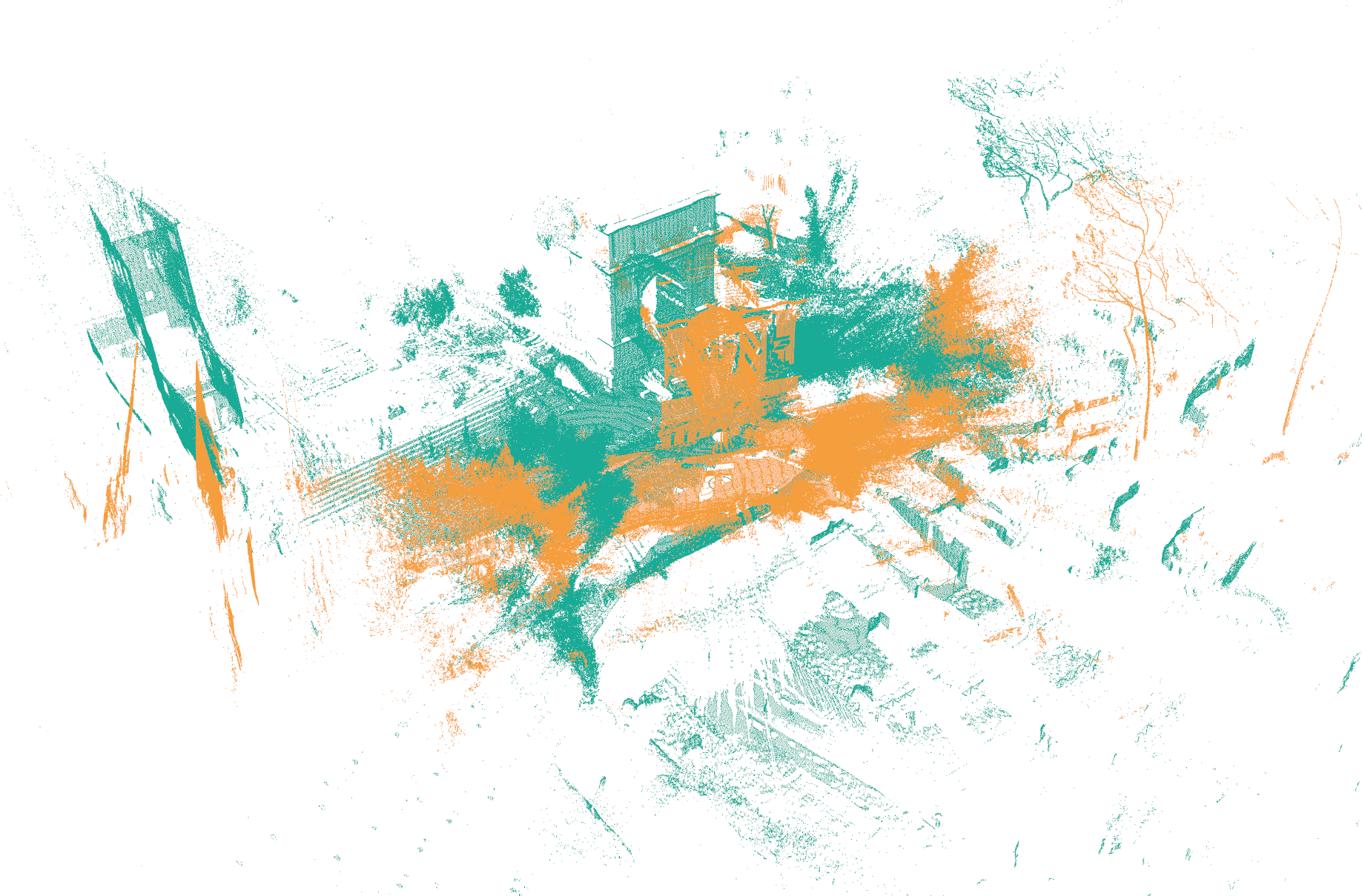}}
	\end{minipage}
    & \begin{minipage}[b]{0.22\textwidth}
		\centering
		\raisebox{-.5\height}{\includegraphics[width=0.8\linewidth]{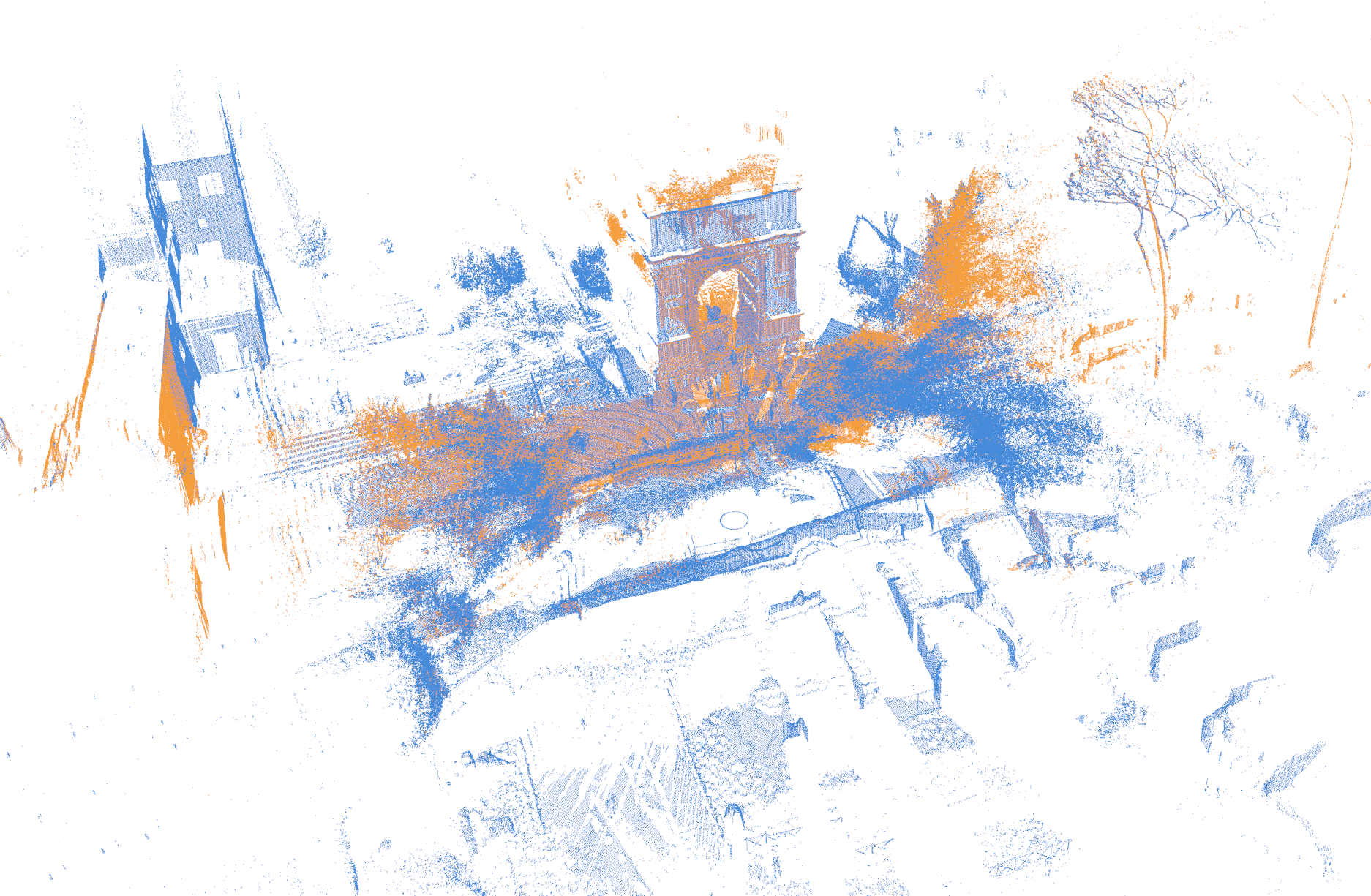}}
	\end{minipage}
    & \begin{minipage}[b]{0.22\textwidth}
		\centering
		\raisebox{-.5\height}{\includegraphics[width=0.8\linewidth]{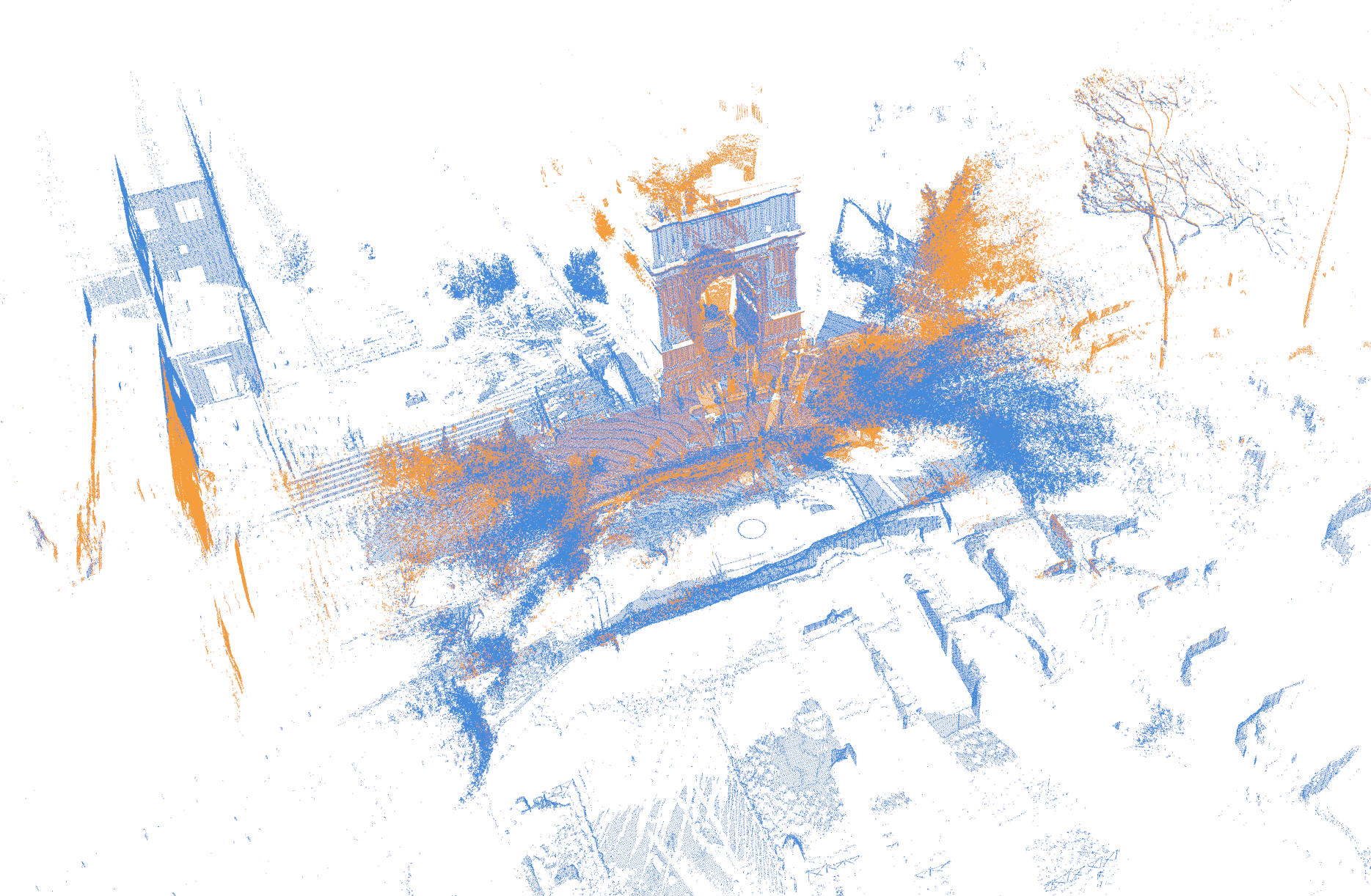}}
	\end{minipage}
    & \begin{minipage}[b]{0.22\textwidth}
		\centering
		\raisebox{-.5\height}{\includegraphics[width=0.8\linewidth]{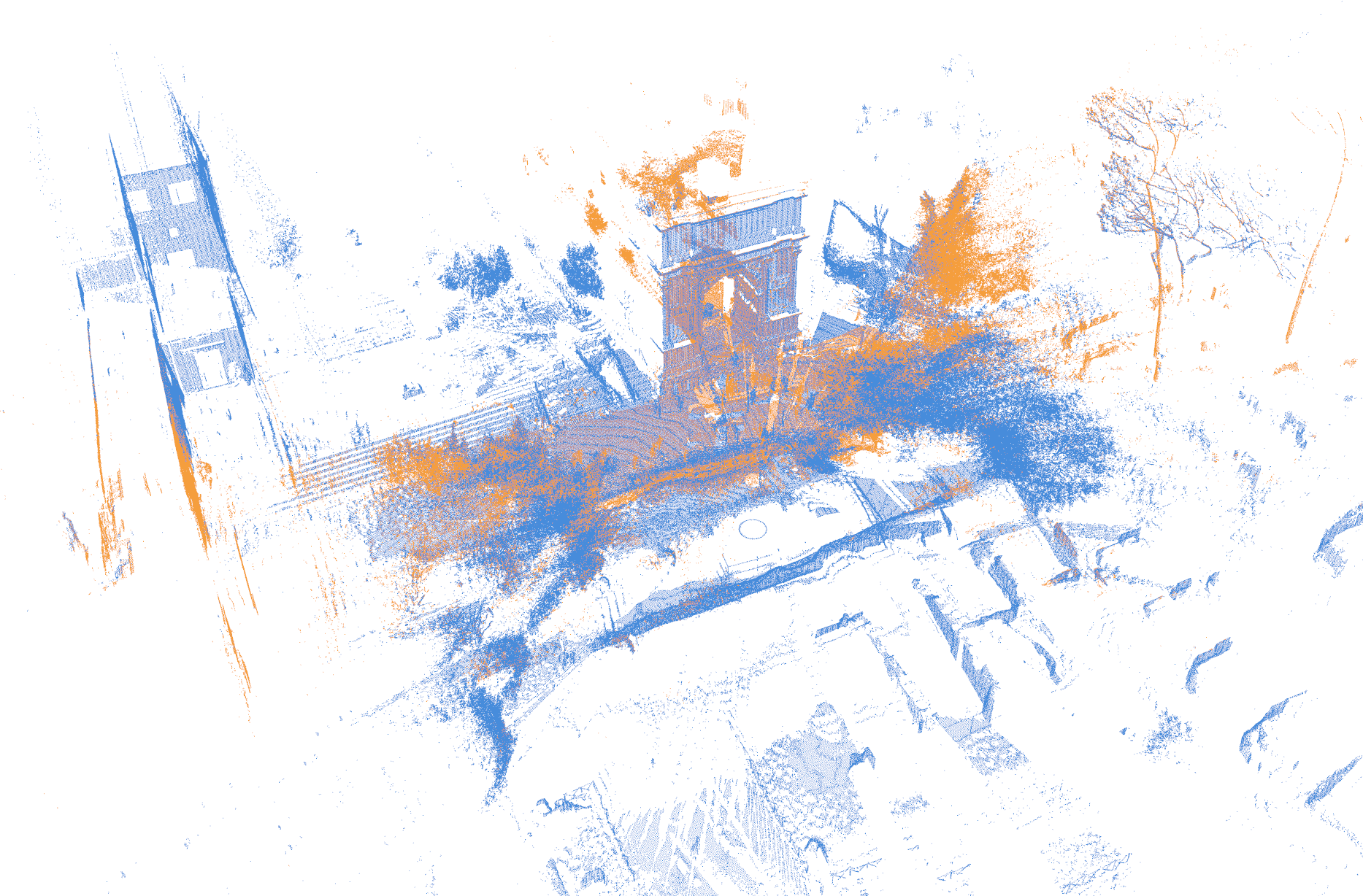}}
	\end{minipage}
    \\
    ~ & \footnotesize{(b-1) Initial ($98.45\%$ outlier)} & \footnotesize{(b-2) GORE\cite{bustos2017guaranteed}} & \footnotesize{(b-3) TEASER\cite{yang2020teaser}} & \footnotesize{(b-4) Ours}
     \\ 
    ~ & \footnotesize - & \footnotesize {$E_{\bm{R}}=\textbf{0.055}\degree$} & \footnotesize {$E_{\bm{R}}=0.197\degree$} & \footnotesize {$E_{\bm{R}}=0.104\degree$}
    \\ 
    ~ & \footnotesize - & \footnotesize {$E_{\bm{t}}=\textbf{0.018}\si{\metre}$} & \footnotesize {$E_{\bm{t}}=0.031\si{\metre}$} & \footnotesize {$E_{\bm{t}}=\textbf{0.018}\si{\metre}$}
    \\ 
    ~ & \footnotesize - & \footnotesize {${time}=687.7$s} & \footnotesize {${time}=5.917$s} & \footnotesize {${time}=\textbf{1.436}$s}
    \\ 

  \multirow{5}*{\rotatebox{90}{{\makecell[c]{Correspondence-based \\ (FCGF descriptor)}}}} 
    & \begin{minipage}[b]{0.2\textwidth}
		\centering
		\raisebox{-.5\height}{\includegraphics[width=0.8\linewidth]{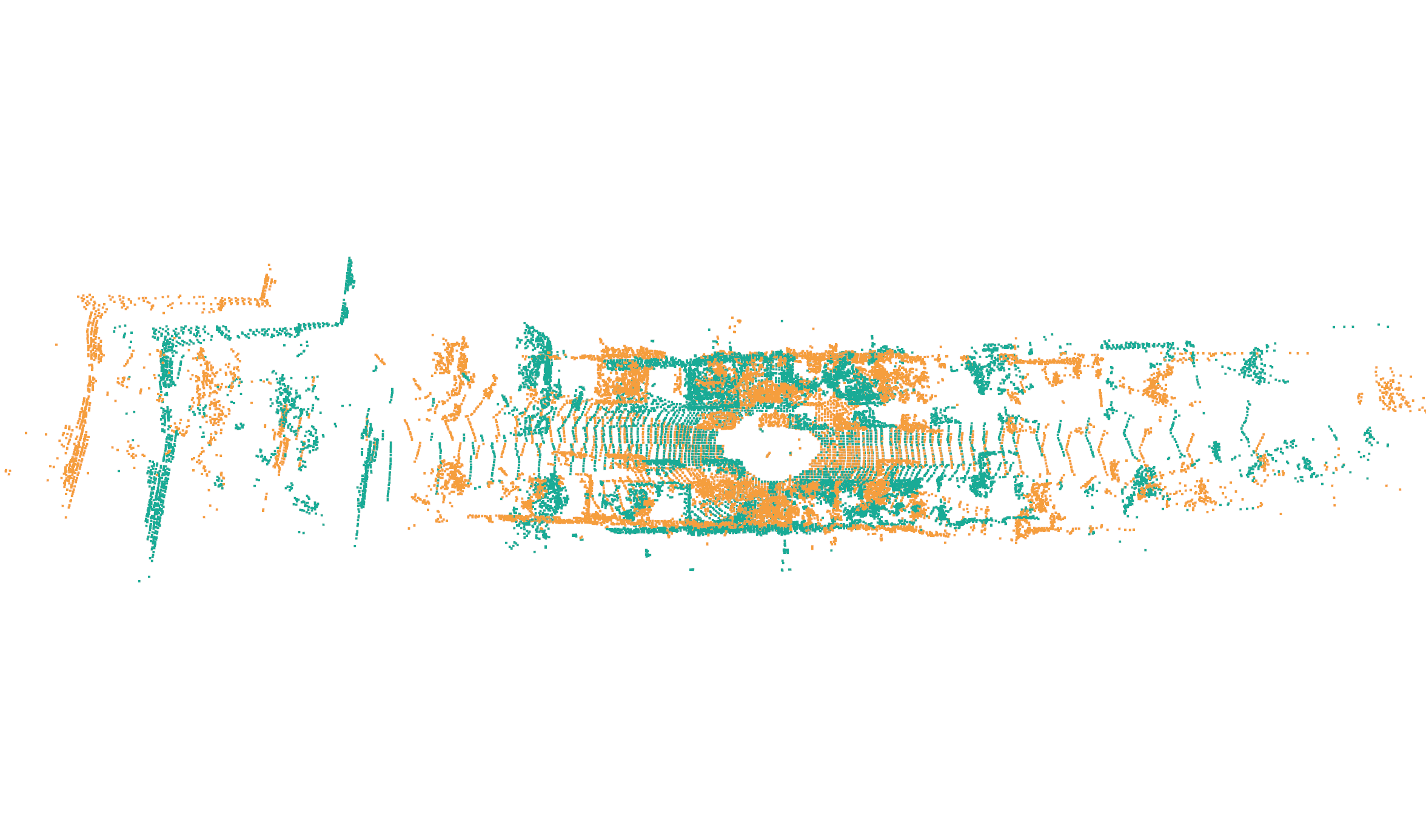}}
	\end{minipage}
    & \begin{minipage}[b]{0.2\textwidth}
		\centering
		\raisebox{-.5\height}{\includegraphics[width=0.8\linewidth]{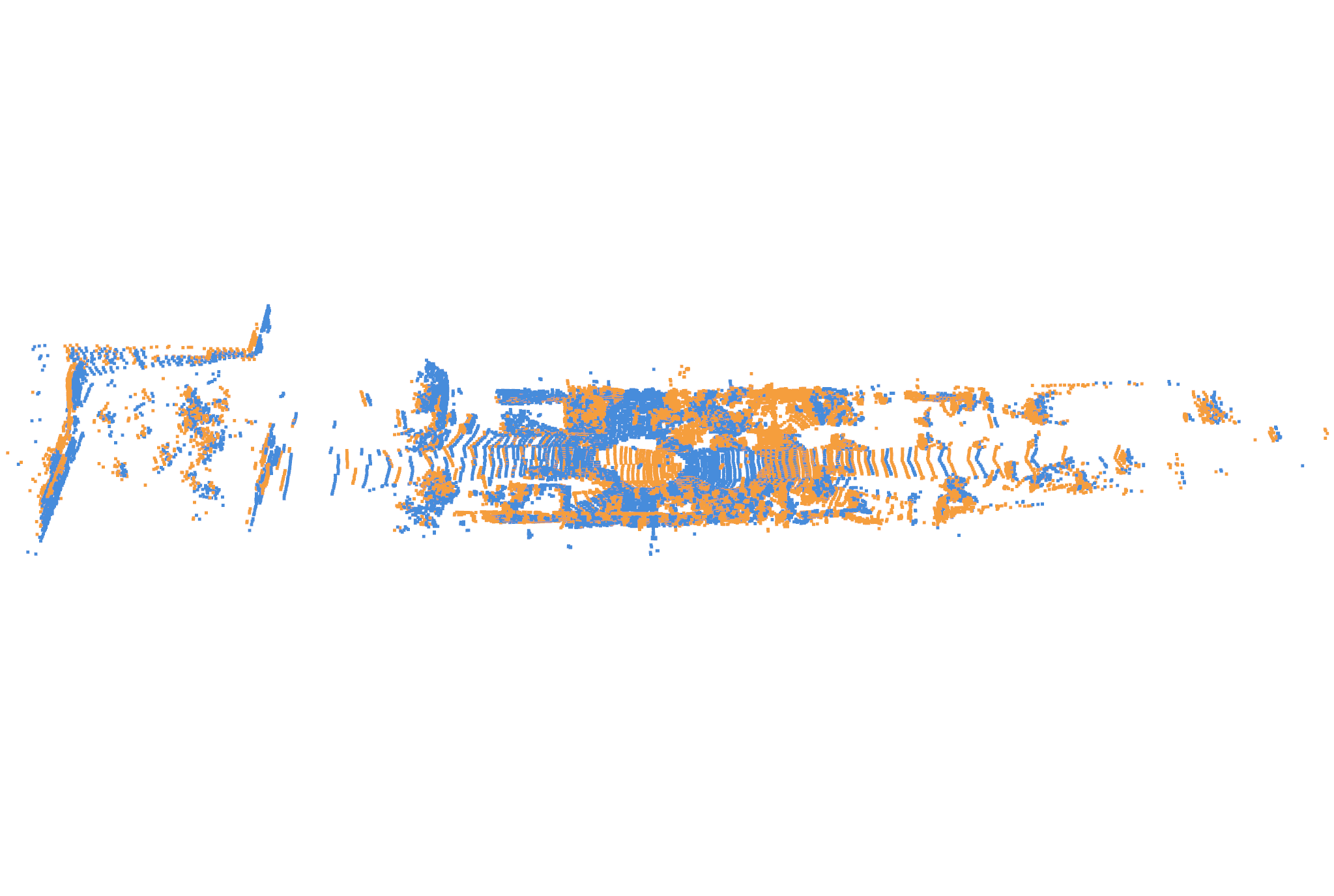}}
	\end{minipage}
    & \begin{minipage}[b]{0.2\textwidth}
		\centering
		\raisebox{-.5\height}{\includegraphics[width=0.8\linewidth]{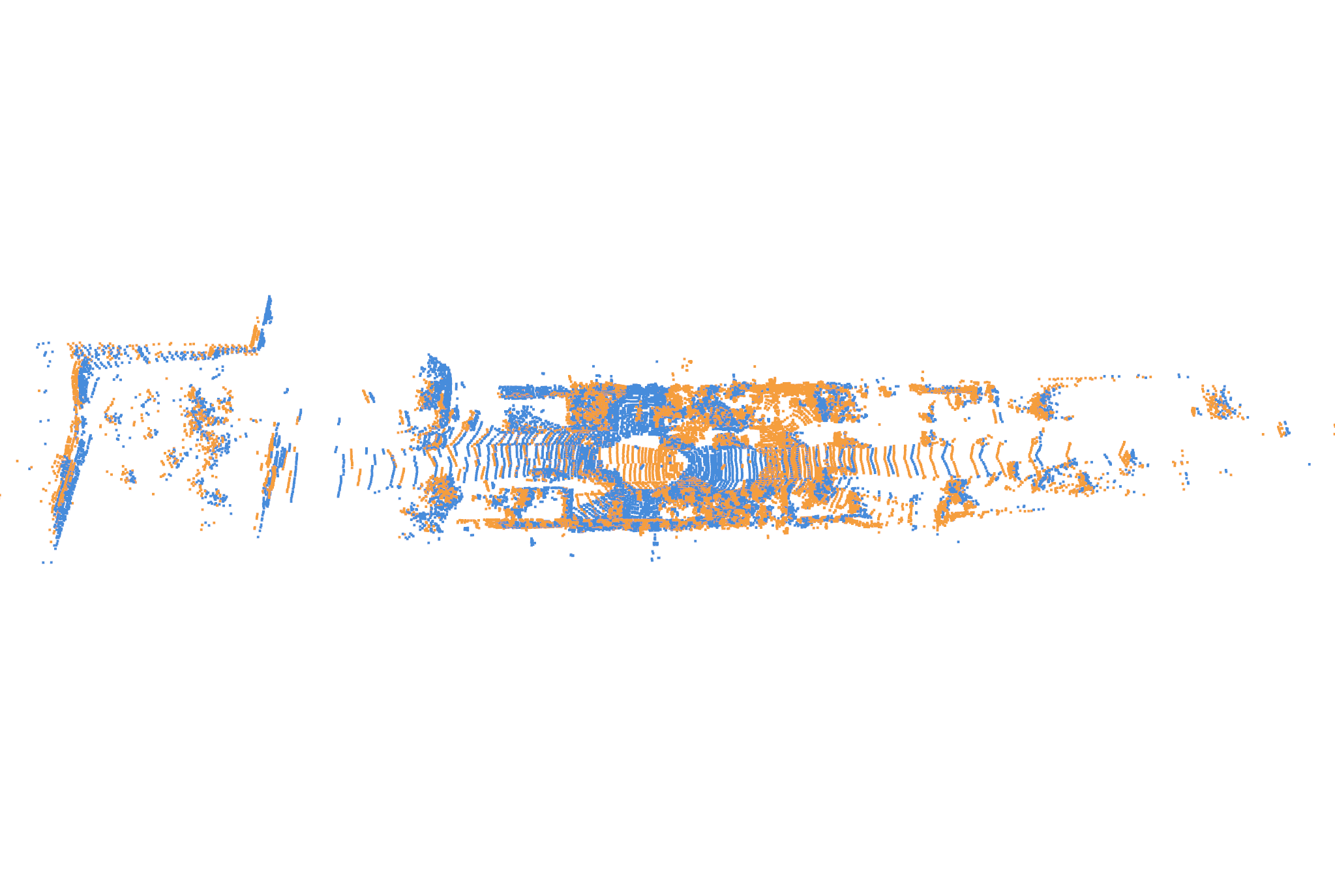}}
	\end{minipage}
    & \begin{minipage}[b]{0.2\textwidth}
		\centering
		\raisebox{-.5\height}{\includegraphics[width=0.8\linewidth]{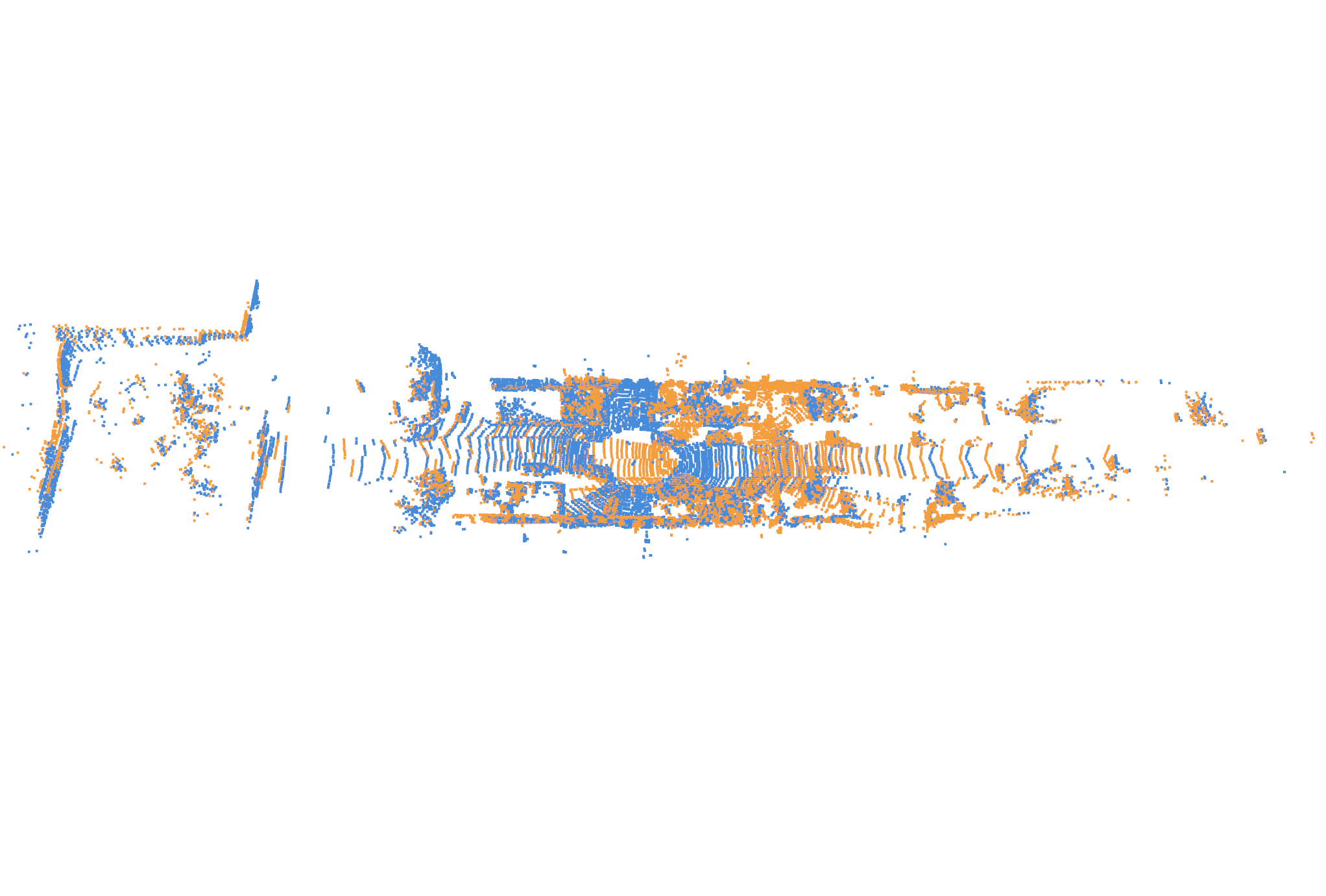}}
	\end{minipage}
    \\
    ~ & \footnotesize{(c-1) Initial ($73.73\%$ outlier)} & \footnotesize{(c-2) FGR\cite{zhou2016fast}} & \footnotesize{(c-3) GC-RANSAC\cite{barath2021graph}} & \footnotesize{(c-4) Ours}
     \\ 
    ~ & \footnotesize - & \footnotesize {$E_{\bm{R}}=0.220\degree$} & \footnotesize {$E_{\bm{R}}=0.294\degree$} & \footnotesize {$E_{\bm{R}}=\textbf{0.212}\degree$}
    \\ 
    ~ & \footnotesize - & \footnotesize {$E_{\bm{t}}=0.521\si{\metre}$} & \footnotesize {$E_{\bm{t}}=0.510\si{\metre}$} & \footnotesize {$E_{\bm{t}}=\textbf{0.374}\si{\metre}$}
    \\ 
    ~ & \footnotesize - & \footnotesize {${time}=1.258$s} & \footnotesize {${time}=1.576$s} & \footnotesize {${time}=\textbf{0.562}$s}
    \\ 
    
  \multirow{5}*{\rotatebox{90}{Correspondence-free}} 
    & \begin{minipage}[b]{0.15\textwidth}
		\centering
		\raisebox{-.5\height}{\includegraphics[width=0.8\linewidth]{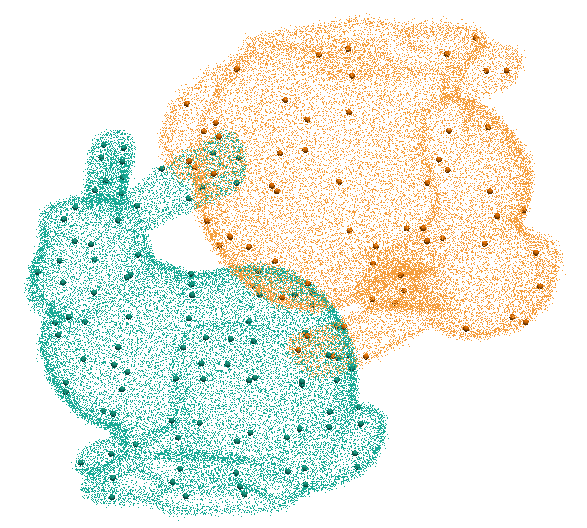}}
	\end{minipage}
    & \begin{minipage}[b]{0.15\textwidth}
		\centering
		\raisebox{-.5\height}{\includegraphics[width=0.8\linewidth]{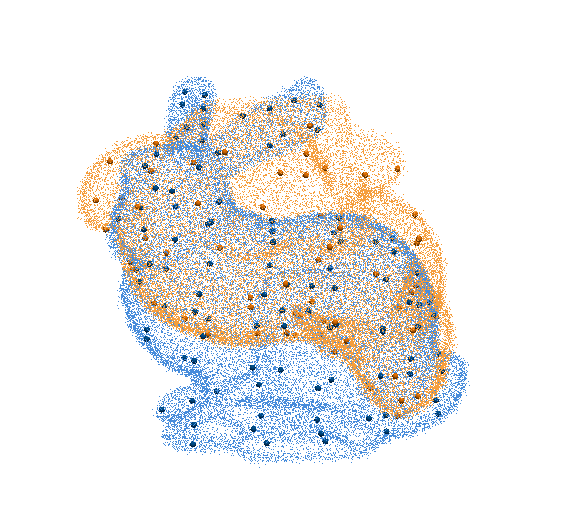}}
	\end{minipage}
    & \begin{minipage}[b]{0.15\textwidth}
		\centering
		\raisebox{-.5\height}{\includegraphics[width=0.8\linewidth]{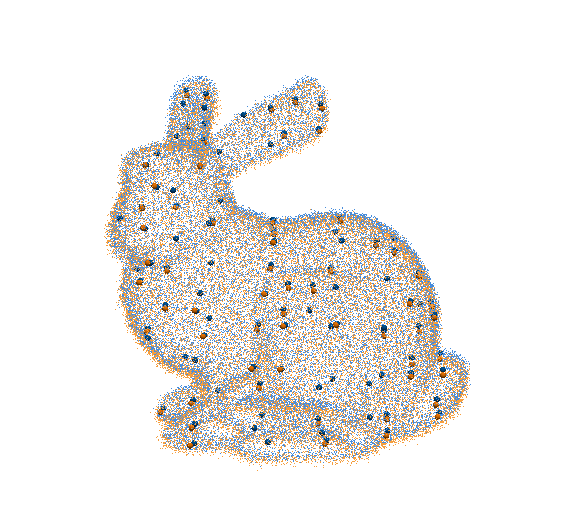}}
	\end{minipage}
    & \begin{minipage}[b]{0.15\textwidth}
		\centering
		\raisebox{-.5\height}{\includegraphics[width=0.8\linewidth]{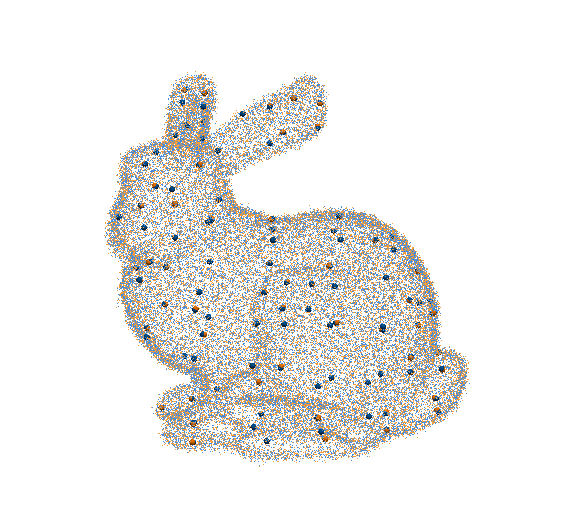}}
	\end{minipage}
    \\
    ~ & \footnotesize{(d-1) Initial ($60\%$ overlap)} & \footnotesize{(d-2) ICP\cite{121791}} & \footnotesize{(d-3) GoICP\cite{7368945}} & \footnotesize{(d-4) Ours}
     \\ 
    ~ & \footnotesize - & \footnotesize {$E_{\bm{R}}=173.7\degree$} & \footnotesize {$E_{\bm{R}}=2.092\degree$} & \footnotesize {$E_{\bm{R}}=\textbf{0.061}\degree$}
    \\ 
    ~ & \footnotesize - & \footnotesize {$E_{\bm{t}}=0.240\si{\metre}$} & \footnotesize {$E_{\bm{t}}=0.017\si{\metre}$} & \footnotesize {$E_{\bm{t}}=\textbf{0.004}\si{\metre}$}
    \\ 
    ~ & \footnotesize - & \footnotesize {${time}=\textbf{0.069}$s} & \footnotesize {${time}=80.40$s} & \footnotesize {${time}=2.524$s}
    \\ 
  \end{tabular}
  \caption{The proposed method can efficiently address the rigid registration problem in different scenarios with high outlier rates or low overlap. The input point clouds are selected from (a) Bremen dataset\cite{borrmann2013thermal}, (b) ETH dataset\cite{theiler2014keypoint}, (c) KITTI dataset\cite{geiger2012we}, and (d) Bunny dataset\cite{curless1996volumetric}, respectively. The source point cloud is green, the target point cloud is yellow, and the aligned point cloud is blue. Compared with state-of-the-art (SOTA) correspondence-based methods, the proposed method achieves significant performance in terms of robustness and efficiency. Besides, the proposed method also can solve the SPCR problem efficiently and robustly.}
  \label{fig_1}
\end{figure*}

Current 3D feature matching approaches have achieved satisfactory development. However, severe outlier correspondences are still inevitable either for handcrafted or learning-based descriptors\cite{rusu2009fast,choy2019fully,bai2020d3feat,huang2021predator,ao2021spinnet}. Several paradigms have been extensively developed to implement robust registration, of which the \textit{consensus maximization} (a.k.a. \textit{inlier set maximization}) is inherently robust to outliers without smoothing or trimming to change the objective function\cite{li2009consensus,campbell2018globally}. Random sample consensus (RANSAC) is the most popular heuristic method for solving the consensus maximization problem of correspondence-based registration. However, RANSAC and its variants are non-deterministic and only generate satisfactory solutions with a certain probability due to the random sampling mechanism\cite{le2019deterministic,barath2021graph}.

More recently, many global and deterministic methods based on the \textit{branch and bound} (BnB) framework have been applied to solve the point cloud registration problem with optimality guarantees\cite{hartley2009global,7368945,campbell2016gogma,7381673,straub2017efficient,bustos2017guaranteed,liu2018efficient,chen2022deterministic}. Nonetheless, the computational complexity of BnB optimization is exponential to the dimension of the solution domain in the worst-case. Most studies address the issue by jointly searching for the optimal solution in $\mathbb{SE}(3)$\cite{7368945,campbell2016gogma,7381673}. In order to improve the algorithm efficiency, one direction is utilizing the known gravity directions measured by inertial measurement units (IMUs) to reduce the dimension of the parameter space to 4-dimension\cite{cai2019practical,li2023fast,liu2023absolute}. Another direction for reducing the problem dimension is to decompose the original problem into two 3-DOF sub-problems by leveraging the geometric properties\cite{straub2017efficient,liu2018efficient,li2018fast,yang2020teaser,9485090,chen2022deterministic}. Typically, two unique categories of features are employed for pose decoupling, i.e., rotation invariant features (RIFs)\cite{liu2018efficient,9485090} and translation invariant measurements (TIMs)\cite{jiao2021deterministic,yang2020teaser}. Nonetheless, the pairwise features increase the number of input data quadratically, resulting in limited efficiency gains. Furthermore, a more efficient strategy is proposed based on the rotation decomposition, which decouples 6-DOF transformation into \textit{i)} (2+1)-DOF, i.e., 2-DOF rotation axis and 1-DOF of translation along the axis, and \textit{ii)} (1+2)-DOF, i.e., the remaining 1-DOF rotation and 2-DOF translation\cite{chen2022deterministic}.

In this paper, we propose an efficient and deterministic search strategy based on residual projections for the rigid registration problem, in which a novel pose decoupling strategy is introduced. Specifically, we decouple the 6-DOF original problem into three search sub-problems by projecting the residuals based on the \textit{Chebyshev distance}, i.e., $L_\infty$ residual\cite{sim2006removing,4385722}, on the coordinate axes. We then define the consensus maximization objective function for each sub-problem and apply a BnB-based optimization method to search for the solution globally and deterministically while obtaining the consensus set. A novel polynomial-time upper bound is derived based on the \textit{interval stabbing} technique\cite{bustos2017guaranteed,cai2019practical,peng2022arcs} for the proposed objective. The proposed BnB algorithm searches for three 2-DOF rotation matrix components individually. Meanwhile, three 1-DOF translation projections on the coordinate axes are implicitly estimated by interval stabbing. After solving these three sub-problems, we can obtain the coarse solution to the 6-DOF registration problem, as well as the final consensus set. Finally, in order to get refined and valid results, the final rotation and translation are re-estimated by using singular value decomposition (SVD) on the estimated consensus set. 

Contrary to existing methods that search in the three-dimensional domain via BnB, the proposed method allows searching only in the two-dimensional parameter space, thus enhancing the computational efficiency, as shown in Fig.~\ref{fig_1}. In addition, the proposed method requires no initialization of the translation domain, which is challenging to accurately determine in different practical scenarios. Therefore, it avoids the problems that would arise when the translation domain is not initialized correctly. Notably, we can also partially verify if the solution is valid by checking whether the coarse solutions of the three sub-problems are orthogonal before SVD. This is because rotation matrices are inherently orthogonal, with a determinant of $1$.

The main contributions of this paper can be summarized as follows:
\begin{itemize}
\item We propose a novel pose decoupling strategy based on the $L_\infty$ residual projections. Compared with existing methods, our approach searches for the solution in a lower-dimensional parameter space, thereby improving search efficiency.

\item We propose a novel deterministic BnB-based search method for the decoupled sub-problems. The specific upper bound is derived based on the \textit{interval stabbing} technique, allowing a further dimensionality reduction of the branching space.

\item Due to its significant robustness, the proposed method can be extended to solve the challenging SPCR problem. We adapt the proposed upper bound to the SPCR objective by \textit{interval merging} technique. 

\end{itemize}

The rest of this paper is organized as follows: The next section addresses the related work in two directions and discusses the innovations of the proposed method. Section \ref{Problem Formulation} illustrates the problem formulation of our proposed method. Section \ref{method} demonstrates the principle and details of our method. Section \ref{Experiments} presents extensive experimental results on both synthetic and real-world datasets. Finally, Section \ref{Conclusion} gives a conclusion.

\section{Related Work}
There are two paradigms for the rigid point cloud registration problem regarding whether putative correspondences are given, i.e., correspondence-based registration and simultaneous pose and correspondence registration.

\subsection{Correspondence-Based Registration}
The correspondence-based registration comprises two steps: \textit{i)} extract the 3D keypoints and build putative correspondences by matching 3D feature descriptors, and \textit{ii)} estimate the 6-DOF transformation based on the given correspondences. During the feature matching process, previous works commonly extracted hand-crafted\cite{rusu2009fast,guo2016comprehensive} or learning-based\cite{gojcic2019perfect,choy2019fully,huang2021predator,ao2021spinnet,wang2022you} 3D descriptors. They then utilized nearest neighbor matcher\cite{lowe2004distinctive} to generate putative correspondences. Despite the significant performance improvements achieved by learning-based descriptors\cite{xia2023casspr,xia2024text2loc}, establishing a completely outlier-free correspondence set remains challenging\cite{chen2022sc2}. Therefore, robust transformation estimators are indispensable. 

Consensus maximization is one of the most popular paradigms to address the robust registration problem, of which heuristic RANSAC\cite{fischler1981random} is the most representative. Recently, several RANSAC-based variants\cite{li2020gesac,barath2021graph,sun2021ransic,barath2021marginalizing} were proposed by introducing novel sampling strategies or local optimization methods. For instance, Graph-cut RANSAC\cite{barath2021graph} (GC-RANSAC) introduced the graph-cut algorithm to improve the local optimization performance. Nonetheless, RANSAC-based methods are non-deterministic and generate a correct solution only with a certain probability due to the essence of random sampling\cite{le2019deterministic}.

Given this context, numerous deterministic and robust methods have been proposed, most of which rely on the globally-optimal BnB framework\cite{bustos2017guaranteed,cai2019practical,chen2022deterministic,li2023fast}. Parra and Chin\cite{bustos2017guaranteed} proposed a guaranteed outlier removal (GORE) method, which leverages geometrical bounds to prune outliers and guarantees that eliminated correspondences are not the inlier. GORE converts the 6-DOF registration problem to a 3DOF rotational registration problem and then utilizes BnB to maximize the inlier set. Similar to GORE, Cai \textit{et al.}\cite{cai2019practical} presented a deterministic pre-processing method to prune outliers for the 4-DOF terrestrial LiDAR registration problem. More recently, Chen \textit{et al.}\cite{chen2022deterministic} introduced an efficient decomposition scheme for the 6-DOF rigid registration, which decouples the 6-DOF original problem into a (2+1)-DOF sub-problem and a (1+2)-DOF sub-problem. These two 3-DOF sub-problems are then sequentially solved by BnB-based search methods. On the other hand, a representative method of the M-estimation paradigm is FGR\cite{zhou2016fast}. This method formulates the registration problem using the Geman-McClure objective function and then combines graduated non-convexity (GNC) to solve the problem. Combining the ideas of outlier removal and M-estimation, Yang \textit{et al.}\cite{yang2020teaser} proposed TEASER, a certifiable and deterministic approach. TEASER leverage TIMs to decouple the 6-DOF transformation search problem into a 3-DOF rotation search sub-problem followed by a 3-DOF translation search sub-problem. As mentioned before, existing decoupling-based methods typically decompose the original 6-DOF problem into two 3-DOF sub-problems. Inspired by this observation, our study aims to develop a novel pose decoupling strategy that allows us to search for solutions in a lower-dimensional domain.

In addition, recent works have incorporated deep learning techniques for robust transformation estimation, such as 3DRegNet\cite{pais20203dregnet}, DGR\cite{choy2020deep}, PointDSC\cite{bai2021pointdsc}, DHVR\cite{lee2021deep}, and DetarNet\cite{chen2022detarnet}. The pose decoupling strategy is also widely used in these methods. However, they typically require a substantial volume of training data, which may not always be readily available in real-world applications. Moreover, their robustness to the unseen data remains limited. As such, this paper focuses on geometric methods for robust registration. 

\subsection{Simultaneous Pose and Correspondence Registration}

The SPCR problem is more challenging since the transformation and correspondences need to be estimated simultaneously. A typical algorithm is ICP\cite{121791}, an expectation-maximization (EM) type method. However, ICP is susceptible to local minima and is greatly influenced by the initialization of the transformation. Another series of noise-robust methods represent the point cloud as Gaussian mixture models (GMMs) to build robust objective functions based on the probability density\cite{myronenko2010point,jian2010robust}. Although all these methods can efficiently converge to an optimum when they are initialized well, they can not provide any optimality guarantees. 

Another line for the SPCR problem is to estimate the globally-optimal solution without initialization, which is commonly based on the BnB framework\cite{7368945,7381673,campbell2016gogma,straub2017efficient,liu2018efficient}. Go-ICP\cite{7368945} is the first practical globally-optimal SPCR approach that employs the nested BnB structure to jointly search for the solution over the 6-dimensional domain. To improve efficiency, Straub \textit{et al.}\cite{straub2017efficient} proposed a decoupling method based on surface normal distributions that decomposes the 6-DOF registration problem into the separate 3-DOF rotation and 3-DOF translation sub-problems. Liu \textit{et al.}\cite{liu2018efficient} introduced the RIFs to enable sequential estimations of the 3-DOF translation and the 3-DOF rotation. By contrast, the proposed method not only presents a novel decoupling strategy but also can be extended to solve the SPCR problem by adjusting the bound functions. The detailed formulation will be given in Section \ref{Challenge}.

In recent studies, researchers also have developed learning-based end-to-end registration approaches\cite{wang2019deep,10018266,10319695,10168979,10185932,10475373} to solve partial-overlap problems. These approaches conduct feature extraction and transformation estimation in a single-forward pass. Although learning-based methods have achieved remarkable performance improvements, it is worth noting that their generalization capabilities may not always be reliable.

\section{Problem Formulation}\label{Problem Formulation}
\subsection{Inlier Set Maximization}
Given the source point cloud \(\mathcal{P}\) and the target point cloud \(\mathcal{Q}\), a set of putative correspondences $\mathcal{K}=\left\{(\bm{p}_i,\bm{q}_i)\right\}_{i=1}^N$ is extracted by matching points between $\mathcal{P}$ and $\mathcal{Q}$, where \(\bm{p}_i,\bm{q}_i\in\mathbb{R}^3\), and $N$ is the correspondences number. The proposed method aims to estimate the rigid transformation between the source and target point clouds. Specifically, the 6-DOF transformation matrix \(\bm{T}\in \mathbb{SE}(3)\) is formed by the 3-DOF rotation matrix \(\bm{R}\in \mathbb{SO}(3)\) and the 3-DOF translation vector \(\bm{t}\in\mathbb{R}^3\). The rotation matrix \(\bm{R}\) is an orthogonal matrix in which the columns and rows are orthogonal vectors, i.e., $\bm{R}\bm{R}^\mathrm{T}=\bm{I}$, with a determinant of $1$. Formally, we adopt the inlier set maximization formulation for the robust registration problem:
\begin{equation}
    \bm{T}^*=\mathop{\arg\max}_{\bm{T}\in \mathbb{SE}(3)}E\left(\bm{T}(\mathcal{P}),\mathcal{Q}\right),
\end{equation}
where $E$ is the objective function for calculating the cardinality of the inlier set. 

Different from existing approaches\cite{7368945,7381673,bustos2017guaranteed,cai2019practical,chen2022deterministic} that commonly employ the $L_2$ residual to measure the alignment, we apply the \textit{Chebyshev distance}, i.e., $L_\infty$ residual\cite{sim2006removing,4385722,liu2018efficient}, to build the robust objective function. Therefore, considering the presence of noise, we estimate the rotation and translation that maximize the objective:
\begin{equation}
E(\bm{R},\bm{t}|\mathcal{K},\epsilon)= \sum_{i=1}^{N} \mathbb{I}\left(\left\| \bm{R}\bm{p}_i+\bm{t}-\bm{q}_i \right\|_\infty \leq\epsilon\right),\label{qie}
\end{equation}
where $\mathbb{I}(\cdot)$ is the indicator function that returns $1$ if the input condition is true and $0$ otherwise, $\left\|\cdot\right\|_{\infty}$ denotes the $L_\infty$-norm, and $\epsilon$ is the inlier threshold.

\subsection{Residual Projections and Pose Decoupling}

\begin{figure}
\centering
\includegraphics[width=1\columnwidth]{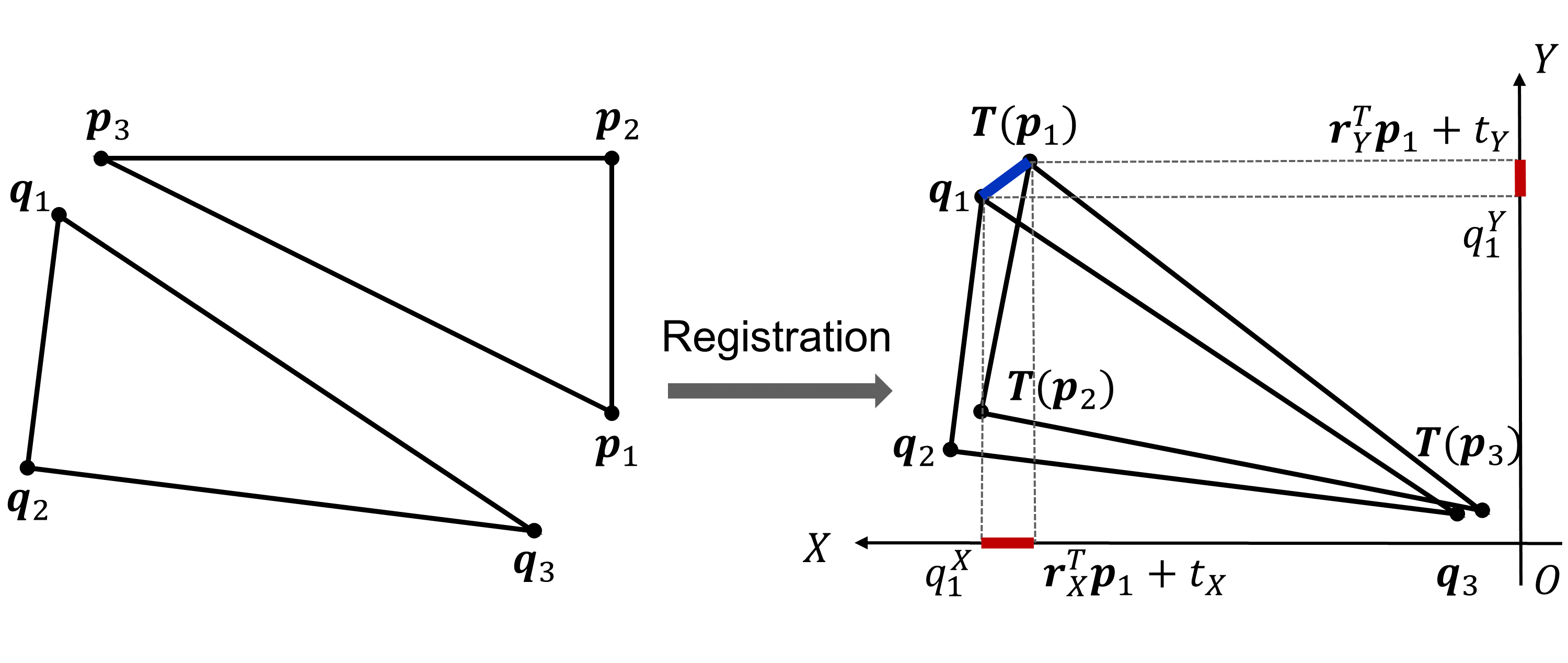}
\caption{A toy 2D registration example to demonstrate $L_\infty$ residual projection. Specifically, $\left\{(\bm{p}_i,\bm{q}_i)\right\}_{i=1}^3$ is the set of input correspondences, $\bm{r}_j$, $j=X, Y$, is the transpose of each row of the rotation matrix, and $t_j$, $j=X, Y$, is the component of the translation vector. The red line segments represent the projections of the residual on the coordinate axes $X$ and $Y$, i.e., $\left| \bm{r}_j^\mathrm{T}\bm{p}_1+t_j-q_1^j \right|$, $j=X, Y$. The inlier constraint for $L_\infty$ residual indicates that $(\bm{p}_1,\bm{q}_1)$ is an inlier only if both residual projections on the coordinate axes are not larger than the inlier threshold.}
\label{fig_2}
\end{figure}

Mathematically, we apply the following definitions to derive the residual projections. First, we denote the rotation matrix as
\begin{equation}
\bm{R}\triangleq\begin{bmatrix} r_{X1} & r_{X2} & r_{X3}\\
r_{Y1} & r_{Y2} & r_{Y3}\\
r_{Z1} & r_{Z2} & r_{Z3}\end{bmatrix}
=\begin{bmatrix}\bm{r}_X & \bm{r}_Y & \bm{r}_Z \end{bmatrix}^\mathrm{T},
\end{equation}
where $\bm{r}_j=\left[r_{j1}, r_{j2}, r_{j3}\right]^\mathrm{T}$, $j=X,Y,Z$, is the transpose of each row of the rotation matrix. The translation vector is  
\begin{equation}
\bm{t}\triangleq\left[t_X, t_Y, t_Z\right]^\mathrm{T}.
\end{equation}

Given the definitions of $\bm{R}$ and $\bm{t}$, according to the definition of Chebyshev distance, the inlier constraint in the objective function (\ref{qie}) can be rewritten as
\begin{subequations}
\begin{align}
    &\left\| \bm{R}\bm{p}_i+\bm{t}-\bm{q}_i \right\|_\infty \leq\epsilon
    \\
    \Leftrightarrow &   \left\| \begin{bmatrix}\bm{r}_X ^\mathrm{T}\\ \bm{r}_Y ^\mathrm{T}\\ \bm{r}_Z ^\mathrm{T}\end{bmatrix}\bm{p}_i+
    \begin{bmatrix}
        t_X\\t_Y\\t_Z
    \end{bmatrix}
    -
\begin{bmatrix}
{q}_i^X\\
{q}_i^Y \\
{q}_i^Z\\
\end{bmatrix}
    \right\|_\infty\leq\epsilon
    \\
    \Leftrightarrow&\mathop{\max}\left\{
    \begin{aligned}    
&\left| \bm{r}_X^\mathrm{T}\bm{p}_i+t_X-q_i^X \right|,
        \\
&\left| \bm{r}_Y^\mathrm{T}\bm{p}_i+t_Y-q_i^Y \right|,
        \\
&\left| \bm{r}_Z^\mathrm{T}\bm{p}_i+t_Z-q_i^Z \right|
    \end{aligned}
    \right\}\leq\epsilon
    \\
\Leftrightarrow&\left\{
    \begin{aligned}    
&\left| \bm{r}_X^\mathrm{T}\bm{p}_i+t_X-q_i^X \right|\leq\epsilon,
        \\
&\left| \bm{r}_Y^\mathrm{T}\bm{p}_i+t_Y-q_i^Y \right|\leq\epsilon,
        \\
&\left| \bm{r}_Z^\mathrm{T}\bm{p}_i+t_Z-q_i^Z \right|\leq\epsilon
    \end{aligned}
    \right.
    \\
\Leftrightarrow&\left\{
    \begin{aligned}    
&\mathbb{I}\left(\left| \bm{r}_X^\mathrm{T}\bm{p}_i+t_X-q_i^X \right|\leq\epsilon\right)=1,
        \\
&\mathbb{I}\left(\left| \bm{r}_Y^\mathrm{T}\bm{p}_i+t_Y-q_i^Y \right|\leq\epsilon\right)=1,
        \\
&\mathbb{I}\left(\left| \bm{r}_Z^\mathrm{T}\bm{p}_i+t_Z-q_i^Z \right|\leq\epsilon\right)=1
    \end{aligned}
    \right.    
\end{align}
\end{subequations}
where $\bm{q}_i\triangleq\left[q_i^X, q_i^Y, q_i^Z\right]^\mathrm{T}$, and $\left| \bm{r}_j^\mathrm{T}\bm{p}_i+t_j-q_i^j \right|$, $j=X,Y,Z$, are projections of the $i$-th residual on the coordinate axes, as shown in Fig.~\ref{fig_2}. Then we can set $\mathbb{I}\left(\left| \bm{r}_j^\mathrm{T}\bm{p}_i+t_j-q_i^j \right| \leq\epsilon\right)=\mathcal{L}_i^j$. Therefore, the objective function (\ref{qie}) can be reformulated as
\begin{equation}
E(\bm{R},\bm{t}|\mathcal{K},\epsilon)= \sum_{i=1}^{N} \mathbb{I}\left(\mathcal{L}_i^X\land\mathcal{L}_i^Y\land\mathcal{L}_i^Z\right),\label{obj}
\end{equation}
where $\land$ is the logical \textit{AND} operation. 

Geometrically, the objective function (\ref{obj}) indicates that, given an arbitrary correspondence $(\bm{p}_i,\bm{q}_i)$ and the inlier threshold $\epsilon$, only when the residual projections on the $X$, $Y$, and $Z$ coordinate axes are not larger than $\epsilon$, $(\bm{p}_i,\bm{q}_i)$ is an inlier, as shown in Fig.~\ref{fig_2}. Notably, these three conditions are equally independent. Accordingly, we may reduce the original constraint in Eq.~(\ref{obj}) as three separate constraints, i.e., $\mathcal{L}_i^X$, $\mathcal{L}_i^Y$, and $\mathcal{L}_i^Z$. 

In this way, the original search problem for the transformation in $\mathbb{SE}(3)$ can be decoupled into three sub-problems. The inlier set maximization objective for each sub-problem can be
\begin{equation}
E_j(\bm{r}_j,t_j|\mathcal{K},\epsilon)= \sum_{i=1}^{N} \mathbb{I}\left(\left| \bm{r}_j^\mathrm{T}\bm{p}_i+t_j-q_i^j \right| \leq\epsilon\right), j=X,Y,Z.\label{objr}
\end{equation}
In other words, we reformulate the $L_\infty$ residual-based objective function in the form of residual projections. Then we decompose the joint constraint into three independent constraints to decouple the original registration problem into three sub-problems, i.e., $\max E_X(\bm{r}_X,t_X|\mathcal{K},\epsilon)$, $\max E_Y(\bm{r}_Y,t_Y|\mathcal{K},\epsilon)$, and $\max E_Z(\bm{r}_Z,t_Z|\mathcal{K},\epsilon)$. The following section will introduce a step-wise search strategy to solve these three sub-problems.

\section{Step-wise Search Strategy Based on Branch and Bound}\label{method}
Branch and bound (BnB) is an algorithm framework for global optimization\cite{scholz,horst2013global}. Specifically, to design the BnB-based algorithm, two main aspects need to be addressed: \textit{i)} how to parameterize and branch the solution domain, and \textit{ii)} how to efficiently calculate the upper and lower bounds. Then, the BnB-based algorithm can recursively divide the solution domain into smaller spaces and prune the sub-branches by upper and lower bounds until convergence. Compared with existing methods, our key idea is adopting interval stabbing to enable the BnB searching in a lower-dimensional domain.

\begin{figure*}
\centering
\includegraphics[width=0.85\textwidth]{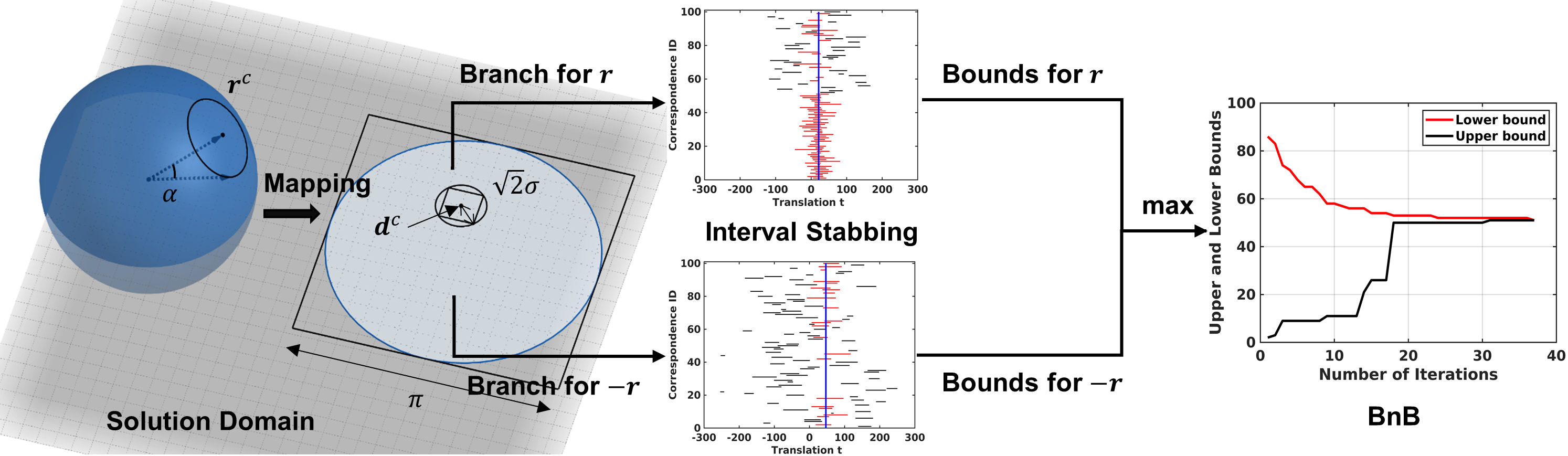}
\caption{The solution domain before and after \textit{exponential mapping}, and the pipeline of our proposed BnB algorithm. The original solution domain of the vector $\bm{r}$ is a \textit{unit sphere} in the 3D Euclidean space. The exponential mapping method maps the unit sphere to two identical 2D disks, representing the solution domains of $\bm{r}$ and $-\bm{r}$, respectively. We can only branch one 2D-disk domain during each iteration, followed by the calculation of upper and lower bounds for each sub-branch. The proposed BnB algorithm converges until the optimal solution $\bm{r}^*$ is found, and the optimal $t^*$ is found by \textit{interval stabbing} simultaneously. In the visualization results of interval stabbing, the black line segments are the candidate intervals of each correspondence, and the red line segments are the intervals crossed by the blue probe with the \textit{max-stabbing number}. The probe position is the \textit{max-stabbing position}.}
\label{fig_3}
\end{figure*}

\subsection{Parametrization of Solution Domain}\label{Parametrization}
\subsubsection{Rotation}

For each sub-problem of the objective function (\ref{objr}), the unknown-but-sought vector $\bm{r}_j$ (denoted by $\bm{r}$ in this section) is on the surface of a \textit{unit sphere} (denoted by $\mathbb{S}^{2}$). Then we divide the unit sphere into two unit hemispheres ($\mathbb{S}^{2+}$ and $\mathbb{S}^{2-}$) to represent the parameter spaces of the “positive” vector $\bm{r}$ and the “negative” vector $-\bm{r}$. The “upper” hemisphere is defined as 
\begin{equation}
    \mathbb{S}^{2+}=\left\{\bm{r} | \bm{r}^\mathrm{T}\bm{r}=1,r_3\geq 0\right\},
\end{equation}
where $\bm{r}\triangleq[r_1,r_2,r_3]^\mathrm{T}$ is a unit vector in $\mathbb{R}^{3}$. Geometrically, since these two hemispheres are centrally symmetric, the “lower” hemisphere is $\mathbb{S}^{2-}$ which can be seen as $-\mathbb{S}^{2+}$. In order to parametrize $\mathbb{S}^{2+}$ and $\mathbb{S}^{2-}$ minimally, we introduce the \textit{exponential mapping}\cite{liu2020globally,liu2022globally} technique to map a 3-dimensional hemisphere to a 2-dimensional disk efficiently. Specifically, given a vector $\bm{r}\in\mathbb{S}^{2+}$, it can be represented by a corresponding point $\bm{d}\in\mathbb{R}^{2}$ in the 2D disk, i.e.,
\begin{equation}
\bm{r}^\mathrm{T}=\left[\sin(\gamma)\hat{\bm{d}}^\mathrm{T},\cos(\gamma)\right], \quad \text{and} \quad \bm{d}=\gamma\hat{\bm{d}},
\end{equation}
where $\gamma\in[0,\pi/2]$, and $\hat{\bm{d}}$ is a unit vector in $\mathbb{R}^{2}$. Notably, the range of $\gamma$ corresponds to $r_3\geq 0$, and its maximum corresponds to the radius of the 2D disk, i.e., $\pi/2$, as shown in Fig.~\ref{fig_3}. For a vector $-\bm{r}\in\mathbb{S}^{2-}$, we define another exponential mapping method,
\begin{equation}
-\bm{r}^\mathrm{T}=-\left[\sin(\gamma)\hat{\bm{d}}^\mathrm{T},\cos(\gamma)\right].
\end{equation}
Accordingly, the total solution domain (unit sphere) is mapped as two identical 2D disks, which represent the parameter spaces of $\bm{r}$ and $-\bm{r}$, respectively. Compared to the unit sphere representation within three parameters and a unit-norm constraint, the exponential mapping is a more compact representation within only two parameters\cite{liu2022globally}. Meanwhile, for ease of operation, a circumscribed square of the disk domain is initialized as the domain of $\bm{r}$ in the proposed BnB algorithm, and the domain of $-\bm{r}$ is relaxed in the same way. 

Further, we introduce the following lemma\cite{liu2020globally} about the exponential mapping between $\mathbb{S}^{2+}$ and $\mathbb{R}^{2}$.
\begin{lemma}\label{lemma1}
$\bm{r}_a,\bm{r}_b\in\mathbb{S}^{2+}$ are two vectors in the unit hemisphere, and $\bm{d}_a,\bm{d}_b\in\mathbb{R}^{2}$ are corresponding points in the 2D disk. Then we have 
\begin{equation}
\angle(\bm{r}_a,\bm{r}_b)\leq \left\| \bm{d}_a-\bm{d}_b \right\|.
\end{equation}
\end{lemma}
According to Lemma~\ref{lemma1}, we can obtain the following proposition.
\begin{proposition}\label{Proposition1}
Given a sub-branch of the square-shaped domain $\mathbb{D}$, its center is $\bm{d}^c\in\mathbb{R}^{2}$ and half-side length is $\sigma$. For $\forall\bm{d}\in\mathbb{D}$, we have
\begin{equation}
\angle\left(\bm{r},\bm{r}^c\right)\leq \left\| \bm{d}-\bm{d}^c \right\| \leq \sqrt{2}\sigma,\label{in}
\end{equation}
where $\bm{r}$ and $\bm{r}^c$ correspond to $\bm{d}$ and $\bm{d}^c$, respectively. 
\end{proposition}

Defining $\alpha\triangleq\max\angle\left(\bm{r},\bm{r}^c\right)$, we can obtain $\alpha\leq \sqrt{2}\sigma$ with Proposition~\ref{Proposition1}, as shown in Fig.~\ref{fig_3}. Geometrically, Proposition~\ref{Proposition1} indicates that one square-shaped sub-branch of the 2D disk domain is relaxed to a spherical patch of the 3D unit sphere. In addition, Lemma~\ref{lemma1} and Proposition~\ref{Proposition1} hold for both hemispheres $\mathbb{S}^{2+}$ and $\mathbb{S}^{2-}$. In this study, we apply Proposition~\ref{Proposition1} as one of the fundamental parts to derive our proposed bounds.

\subsubsection{Translation}
Estimating the translation component $t_j\in\mathbb{R}$, $j=X,Y,Z$ in the objective function (\ref{objr}) is a 1-dimensional problem. The translation is unconstrained, and it is not easy to estimate a suitable solution domain accurately in advance for various practical scenarios. Existing BnB-based approaches\cite{7368945,chen2022deterministic,liu2018efficient} commonly initialize the translation domain as a redundant space and search it exhaustively, leading to a significant decrease in efficiency. Meanwhile, if the translation domain is not initialized correctly, the algorithm may not find the optimal (correct) solution since the optimal solution may be excluded from the initial search domain.

In this study, we propose an \textit{interval stabbing}-based method to estimate the translation components without any prior information on the translation domain, which can efficiently reduce the solution domain and accordingly accelerate the BnB search. It also avoids the problems that may arise when the translation initialization is incorrect. The proposed method will be described thoroughly in the next sub-section.

\subsection{Interval Stabbing and Bounds}\label{Stabbing}

We first introduce the following lemma to derive the bounds for the objective function (\ref{objr}).

\begin{lemma}\label{lemma2}
Given an arbitrary consensus maximization objective $F(x|A)=\sum_{i=1}^{M}\mathcal{F}_i\left(x,a_i\right)$, where $x$ is the variable to be calculated, $A=\left\{a_i\right\}_{i=1}^M$ is the set of input measurements, and $\mathcal{F}_i(x,a_i)$ is an indicator function with a certain constraint. Then we have 
\begin{equation}
\mathop{\max}_{x}F(x|A)=\mathop{\max}_{x}\sum_{i=1}^{M}\mathcal{F}_i\left(x,a_i\right) \leq \sum_{i=1}^{M}\mathop{\max}_{x}\mathcal{F}_i\left(x,a_i\right).
\end{equation}
\end{lemma}

\begin{proof}
For the $i$-th input measurement $a_i$, we can obtain $\mathcal{F}_i\left(x,a_i\right) \leq \mathop{\max}_{x}\mathcal{F}_i\left(x,a_i\right) \leq 1$. Therefore, it is obvious that the maximum of $\sum_{i=1}^{M}\mathcal{F}_i\left(x,a_i\right)$ is not bigger than the sum of $\mathop{\max}_{x}\mathcal{F}_i\left(x,a_i\right)$.
\end{proof}

In this study, the upper and lower bounds are proposed as follows:
\begin{proposition}[Upper bound for $\mathbb{S}^{2+}$]\label{Proposition2}
Given a sub-branch of the square-shaped domain $\mathbb{D}$, whose center is $\bm{d}^c\in\mathbb{R}^{2}$ (corresponds to $\bm{r}_j^c\in\mathbb{S}^{2+}$) and half-side length is $\sigma$, the upper bound can be set as
\begin{subequations}
\begin{align}
&\overline{E}_j^+(\mathbb{D}) = \mathop{\max}_{t_j}\sum_{i=1}^{N}\mathbb{I}\left( t_j\in\left[t_j^{i-},t_j^{i+}\right]\right),\\
&t_j^{i-}= -\epsilon- \left\| \bm{p}_i \right\|\cos\left(\mathop{\max}\left\{\angle\left(\bm{r}_j^c,\bm{p}_i\right)-\sqrt{2}\sigma, 0\right\}\right) +q_i^j,\\
&t_j^{i+}= \epsilon- \left\| \bm{p}_i \right\|\cos\left(\mathop{\min}\left\{\angle\left(\bm{r}_j^c,\bm{p}_i\right)+ \sqrt{2}\sigma, \pi\right\}\right) +q_i^j.
\end{align}
\end{subequations}
\end{proposition}

\begin{proof}
The rigorous proof can be found in Appendix A.
\end{proof}

\begin{proposition}[Upper bound for $\mathbb{S}^{2-}$]\label{Proposition3}
Given a sub-branch of the square-shaped domain $\mathbb{D}$, whose center is $\bm{d}^c\in\mathbb{R}^{2}$ (corresponds to $-\bm{r}_j^c\in\mathbb{S}^{2-}$) and half-side length is $\sigma$, the upper bound can be set as
\begin{subequations}
\begin{align}
&\overline{E}_j^-(\mathbb{D}) = \mathop{\max}_{t_j}\sum_{i=1}^{N}\mathbb{I}\left( t_j\in\left[t_j^{i-},t_j^{i+}\right]\right),\\
&t_j^{i-}= -\epsilon+ \left\| \bm{p}_i \right\|\cos\left(\mathop{\min}\left\{\angle\left(\bm{r}_j^c,\bm{p}_i\right)+ \sqrt{2}\sigma, \pi\right\}\right) +q_i^j,\\
&t_j^{i+}= \epsilon+ \left\| \bm{p}_i \right\|\cos\left(\mathop{\max}\left\{\angle\left(\bm{r}_j^c,\bm{p}_i\right)-\sqrt{2}\sigma, 0\right\}\right) +q_i^j.
\end{align}
\end{subequations}
\end{proposition}

\begin{proof}
The proof is similar to Proposition~\ref{Proposition2}, which is simple enough that we omit it.
\end{proof}

Although the upper bounds in Proposition~\ref{Proposition2} and Proposition~\ref{Proposition3} are theoretically provided, we still need to find an appropriate method to compute them. Mathematically, the calculation of the upper bounds is a typical \textit{interval stabbing} problem\cite{de1997computational}. As shown in Fig.~\ref{fig_3}, the interval stabbing problem aims to find a probe (i.e., the blue line segment) that stabs the maximum number of intervals. There has been a deterministic and polynomial-time algorithm\cite{cai2019practical} to solve the interval stabbing problem. More details are given in~\cite{de1997computational,cai2019practical}. 

By utilizing the interval stabbing technique to compute the upper bounds, the proposed BnB algorithm only needs to search in a 2-dimensional solution domain, thereby improving the algorithm efficiency. Meanwhile, the translation projections $\{t_X,t_Y,t_Z\}$ are implicitly estimated by 1-dimensional interval stabbing without requiring the initialization of the translation domain. In other words, the interval stabbing approach returns not only the \textit{max-stabbing number} (i.e., the upper bound), but also the \textit{max-stabbing position} (i.e., the estimation of $t_j$).

To sum up, considering the total solution domain $\mathbb{S}^{2+}$ and $\mathbb{S}^{2-}$, we have the following proposition.
\begin{proposition}[Upper bound for $\mathbb{S}^{2}$]\label{Proposition4}
Given a sub-branch of the square-shaped domain $\mathbb{D}$, whose center is $\bm{d}^c\in\mathbb{R}^{2}$ and half-side length is $\sigma$, the upper bound of the objective function  (\ref{objr}) can be set as
\begin{equation}
\overline{E}_j(\mathbb{D}) = \mathop{\max}\left\{\overline{E}_j^+(\mathbb{D}), \overline{E}_j^-(\mathbb{D})\right\}.
\end{equation}
\end{proposition}
\begin{proof}
The maximum of these two upper bounds is not smaller than the maximum of the objective function (\ref{objr}). Therefore, $\overline{E}_j(\mathbb{D})$ is the final upper bound of the objective function (\ref{objr}).
\end{proof}

\begin{proposition}[Lower bound for $\mathbb{S}^{2}$]\label{Proposition5}
Given a sub-branch of the square-shaped domain $\mathbb{D}$, whose center is $\bm{d}^c\in\mathbb{R}^{2}$ and half-side length is $\sigma$, the lower bound of the objective function (\ref{objr}) can be set as
\begin{subequations}
\begin{align}
&\underline{E}_j(\mathbb{D}) = \mathop{\max}\left\{\underline{E}_j^+(\mathbb{D}), \underline{E}_j^-(\mathbb{D})\right\},\\
&\underline{E}_j^+(\mathbb{D}) = \sum_{i=1}^{N} \mathbb{I}\left(\left| \left(\bm{r}_j^c\right)^\mathrm{T}\bm{p}_i+\overline{t}_j^+-q_i^j \right| \leq\epsilon\right),\\
&\underline{E}_j^-(\mathbb{D}) = \sum_{i=1}^{N} \mathbb{I}\left(\left| -\left(\bm{r}_j^c\right)^\mathrm{T}\bm{p}_i+\overline{t}_j^--q_i^j \right| \leq\epsilon\right),
\end{align}
\end{subequations}
where $\overline{t}_j^+$ is the max-stabbing position of the upper bound for $\mathbb{S}^{2+}$, and $\overline{t}_j^-$ is the max-stabbing position of the upper bound for $\mathbb{S}^{2-}$.
\end{proposition}
\begin{proof}
The maximum of the objective function in the given sub-branch $\mathbb{D}$ should be no less than any objective value at a specific point. Therefore, $\underline{E}_j(\mathbb{D})$ is the lower bound of the objective function (\ref{objr}).
\end{proof}

\begin{figure}
\centering
  \begin{tabular}{ c  c }
    \begin{minipage}[b]{0.46\columnwidth}
		\centering
		\raisebox{-.5\height}{\includegraphics[width=\linewidth]{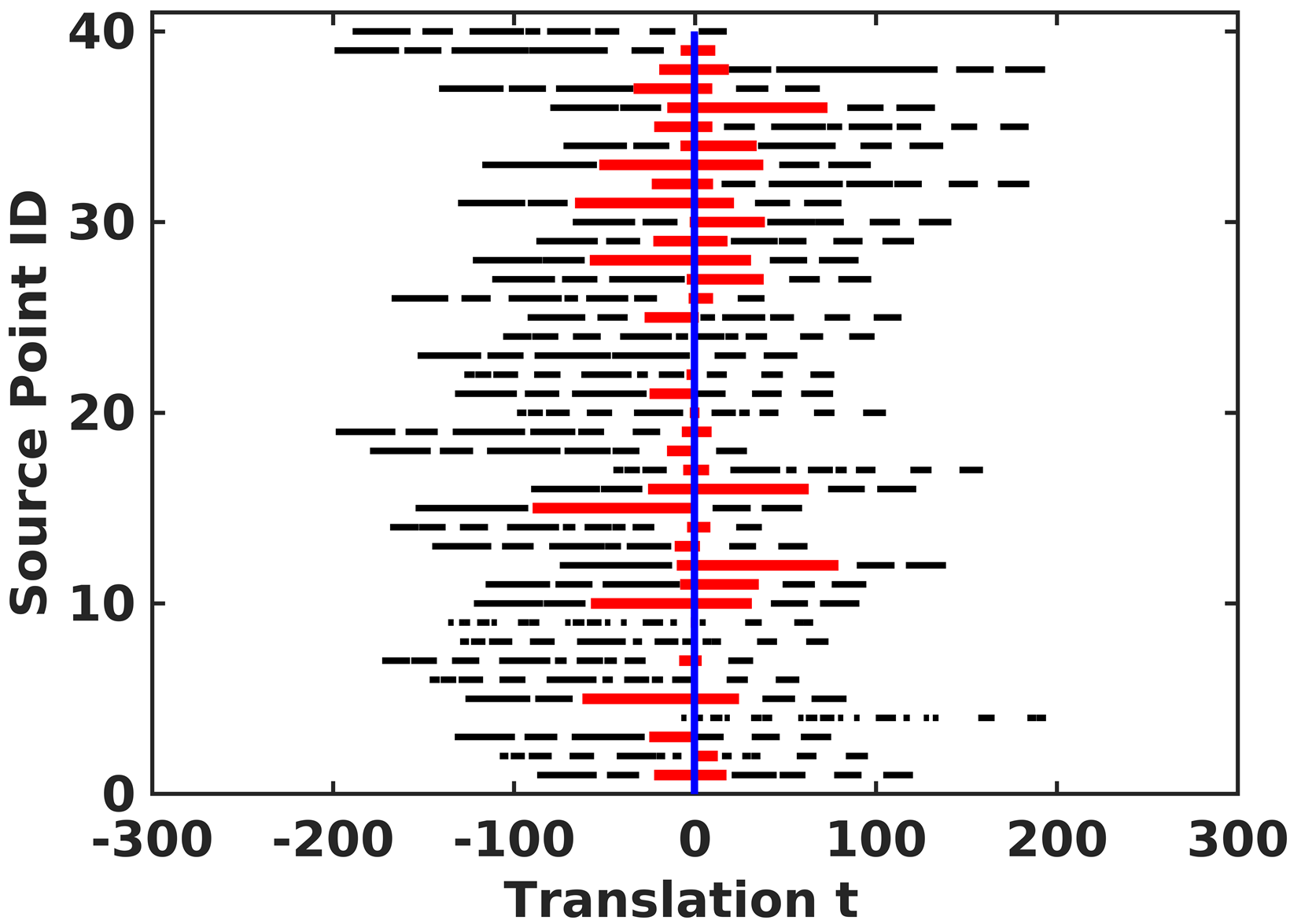}}
	\end{minipage}
    & \begin{minipage}[b]{0.46\columnwidth}
		\centering
		\raisebox{-.5\height}{\includegraphics[width=\linewidth]{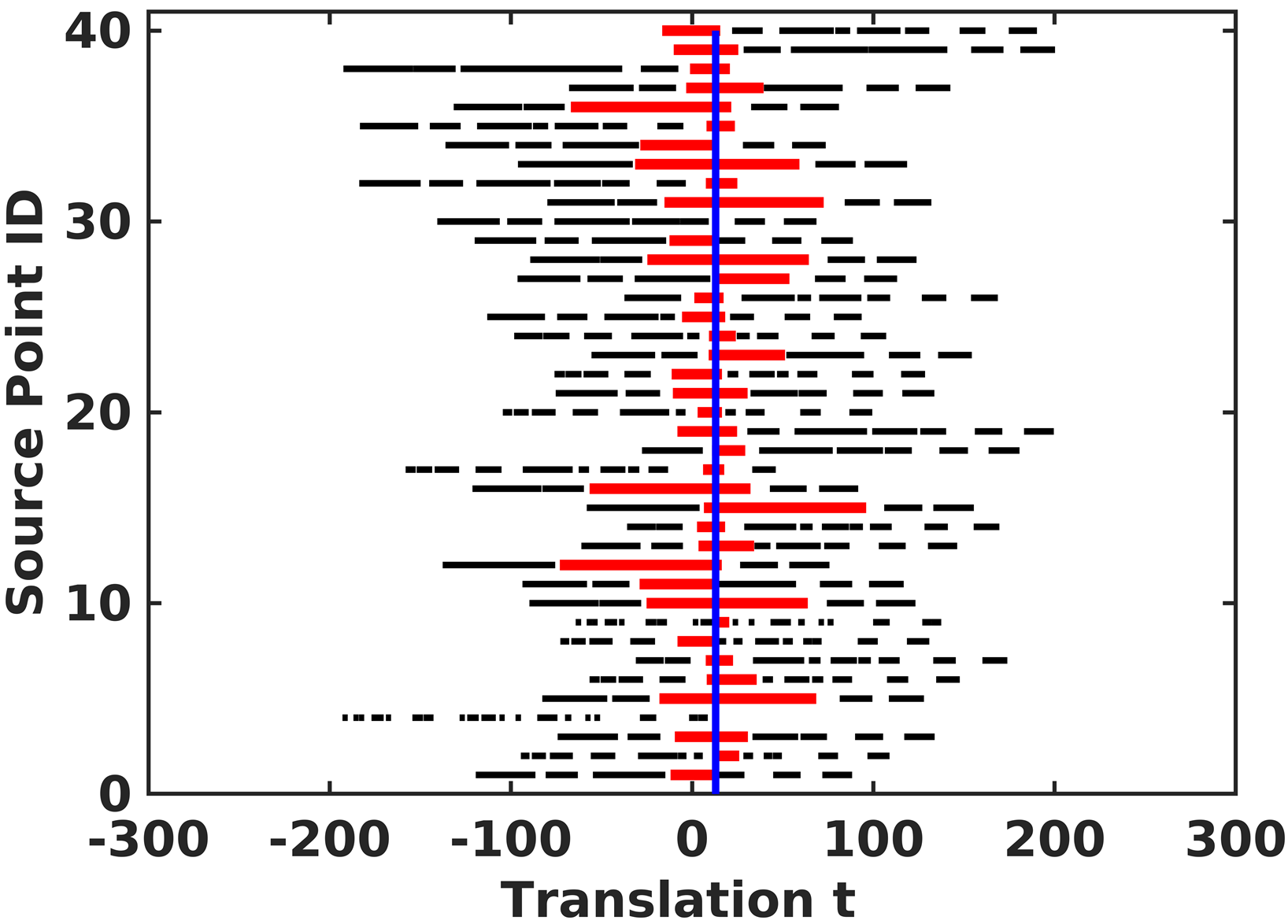}}
	\end{minipage}
    \\
     \footnotesize{(a) SPCR Upper for $\mathbb{S}^{2+}$: 33} & \footnotesize{(b) SPCR Upper for $\mathbb{S}^{2-}$: 38}
  \end{tabular}
\caption{The visualization results for the 24-th iteration of a representative SPCR test on synthetic data. The black line segments are the intervals after \textit{interval merging}, and the red line segments are the intervals crossed by the blue probe with the max-stabbing number. After interval merging, interval stabbing is utilized to calculate the upper bounds. Notably, the blue probe resulting from interval stabbing can only penetrate at most one interval for each point $\bm{p}_i$. The final SPCR upper bound for $\mathbb{S}^{2}$ is 38.} 
\label{fig_4}
\end{figure}

\begin{proposition}[Bound convergence]\label{Proposition6}
When the sub-branch of the square-shaped domain $\mathbb{D}$ collapses to a single point whose center is $\bm{d}^c\in\mathbb{R}^{2}$ and half-side length is zero, we can have 
\begin{equation}
\overline{E}_j(\mathbb{D}) = \underline{E}_j(\mathbb{D}).
\end{equation}
\end{proposition}
\begin{proof}
The rigorous proof can be found in Appendix B.
\end{proof}

Based on the upper and lower bounds in Proposition~\ref{Proposition4} and Proposition~\ref{Proposition5}, the proposed 2D BnB search algorithm for solving decoupled sub-problems is outlined in Appendix C. We employ the depth-first search strategy\cite{meh} to implement the proposed BnB algorithm. As we indicated in Section \ref{Parametrization}, although the initial solution domain $\mathbb{S}^{2}$ is mapped to two identical 2D-disks, only one disk domain is branched, since the bounds of $\mathbb{S}^{2+}$ and $\mathbb{S}^{2-}$ can be computed separately in the same disk domain, as shown in Fig.~\ref{fig_3}. During each iteration, the branch with maximal upper bound is partitioned into four sub-branches since the current parameter space is only 2-dimensional. Then, the branch list is updated, and the upper and lower bounds for each sub-branch are estimated. The sub-branches that do not have a better solution than the best-so-far solution are eliminated. As the number of iterations increases, the gap between the upper and lower bounds gradually decreases. Until the gap reduces to zero, the proposed BnB algorithm obtains the optimal solutions ($\bm{r}_j^*$, $t_j^*$) as well as the maximized consensus set of the sub-problem. As shown in \cite{straub2017efficient,liu2018efficient,chen2022deterministic}, existing decoupling-based methods usually solve sub-problems sequentially. However, we can solve three sub-problems in an arbitrary order with the proposed algorithm and obtain the final results \(\bm{R}^*\) and \(\bm{t}^*\) by SVD refinement.
\subsection{Simultaneous Pose and Correspondence Registration}\label{Challenge}
This section extends the proposed correspondence-based registration method to address the challenging simultaneous pose and correspondence registration (SPCR) problem. The SPCR problem inherently poses greater complexity than the correspondence-based problem. Formally, given the source point cloud $\mathcal{P}=\{\bm{p}_i\}_{i=1}^{M}$ and the target point cloud $\mathcal{Q}=\{\bm{q}_k\}_{k=1}^{N}$, we define the inlier set maximization objective function for the SPCR problem as
\begin{equation}
S_j(\bm{r}_j,t_j|\mathcal{P},\mathcal{Q},\epsilon)= \sum_{i=1}^{M} \mathop{\max}_{k} \mathbb{I}\left(\left| \bm{r}_j^\mathrm{T}\bm{p}_i+t_j-q_k^j \right| \leq\epsilon\right), \label{objs}
\end{equation}
where $j=X,Y,Z$. Similar to \cite{campbell2018globally,9485090}, this formulation implies that for each point $\bm{p}_i$ in $\mathcal{P}$, as long as a sufficiently close point $\bm{q}_k$ in $\mathcal{Q}$ exists, then it contributes 1 to the objective function. Additionally, the main difference between this formulation and the one under the assumption of all-to-all correspondences is the number of counted inliers. In the following, the interval merging technique is introduced to keep the number of counted inliers from exceeding $M$.

The upper and lower bounds for the SPCR objective~(\ref{objs}) are slightly different from those of correspondence-based registration, given by the following propositions. In addition, the optimization of objective~(\ref{objs}) is also based on BnB.

\begin{proposition}[SPCR Upper bound for $\mathbb{S}^{2}$]\label{Proposition7}
Given a sub-branch of the square-shaped domain $\mathbb{D}$, whose center is $\bm{d}^c\in\mathbb{R}^{2}$ and half-side length is $\sigma$, the SPCR upper bound for $\mathbb{S}^{2+}$ can be set as
\begin{subequations}
\begin{align}
&\overline{S}_j^+(\mathbb{D}) = \mathop{\max}_{t_j}\sum_{i=1}^{M} \mathop{\max}_{k}\mathbb{I}\left( t_j\in\left[t_j^{ik-},t_j^{ik+}\right]\right),\label{mergea}\\
&t_j^{ik-}= -\epsilon- \left\| \bm{p}_i \right\|\cos\left(\mathop{\max}\left\{\angle\left(\bm{r}_j^c,\bm{p}_i\right)-\sqrt{2}\sigma, 0\right\}\right) +q_k^j,\\
&t_j^{ik+}= \epsilon- \left\| \bm{p}_i \right\|\cos\left(\mathop{\min}\left\{\angle\left(\bm{r}_j^c,\bm{p}_i\right)+ \sqrt{2}\sigma, \pi\right\}\right) +q_k^j.
\end{align}
\end{subequations}
The SPCR upper bound for $\mathbb{S}^{2-}$ can be set as
\begin{subequations}
\begin{align}
&\overline{S}_j^-(\mathbb{D}) = \mathop{\max}_{t_j}\sum_{i=1}^{M} \mathop{\max}_{k}\mathbb{I}\left( t_j\in\left[t_j^{ik-},t_j^{ik+}\right]\right),\label{mergeb}\\
&t_j^{ik-}= -\epsilon+ \left\| \bm{p}_i \right\|\cos\left(\mathop{\min}\left\{\angle\left(\bm{r}_j^c,\bm{p}_i\right)+ \sqrt{2}\sigma, \pi\right\}\right) +q_k^j,\\
&t_j^{ik+}= \epsilon+ \left\| \bm{p}_i \right\|\cos\left(\mathop{\max}\left\{\angle\left(\bm{r}_j^c,\bm{p}_i\right)-\sqrt{2}\sigma, 0\right\}\right) +q_k^j.
\end{align}
\end{subequations}
The final SPCR upper bound for $\mathbb{S}^{2}$ can be set as
\begin{equation}
\overline{S}_j(\mathbb{D}) = \mathop{\max}\left\{\overline{S}_j^+(\mathbb{D}), \overline{S}_j^-(\mathbb{D})\right\}.
\end{equation}
\end{proposition}
\begin{proof}
The proof is similar to the proofs of Proposition~\ref{Proposition2}, Proposition~\ref{Proposition3}, and Proposition~\ref{Proposition4}, hence we omit it.
\end{proof}

In Proposition~\ref{Proposition7}, the challenge lies in finding a solution to the problem of $\mathop{\max}_{k}\mathbb{I}\left( t_j\in\left[t_j^{ik-},t_j^{ik+}\right]\right)$ while ensuring its maximum value does not exceed $1$. Given that there are $N$ intervals for each point $\bm{p}_i$, and these intervals may overlap, directly applying interval stabbing to these $M\times N$ intervals is unfeasible. Accordingly, the \textit{interval merging} technique\cite{de1997computational} can be employed to solve the problem of $\mathop{\max}_{k}\mathbb{I}\left( t_j\in\left[t_j^{ik-},t_j^{ik+}\right]\right)$. After interval merging, the max-stabbing probe can penetrate at most one interval for each point $\bm{p}_i$, meaning each point $\bm{p}_i$ contributes a maximum of 1 to the upper bound functions (\ref{mergea}) and (\ref{mergeb}). In other words, a point $\bm{q}_k$ which is close enough to point $\bm{p}_i$ is found. The complete interval merging algorithm for all points $\{\bm{p}_i\}_{i=1}^{M}$ is presented in Appendix C. Specifically, interval merging is executed one time for each point $\bm{p}_i$, and then a total of $M$ times for point cloud $\mathcal{P}$. Subsequently, interval stabbing is employed on the merged intervals to compute upper bounds. An example of the visualization results of interval merging and stabbing is given in Fig.~\ref{fig_4}. Similarly, when computing the SPCR lower bound, we employ another indicator function to solve this “multi-interval” problem, as shown in Eq. (\ref{lb}) and Eq. (\ref{ld}) of the following Proposition~\ref{Proposition8}.

\begin{proposition}[SPCR Lower bound for $\mathbb{S}^{2}$]\label{Proposition8}
Given a sub-branch of the square-shaped domain $\mathbb{D}$, whose center is $\bm{d}^c\in\mathbb{R}^{2}$ and half-side length is $\sigma$, the SPCR lower bound can be set as
\begin{subequations}
\begin{align}
&\underline{S}_j(\mathbb{D}) = \mathop{\max}\left\{\underline{S}_j^+(\mathbb{D}), \underline{S}_j^-(\mathbb{D})\right\},\\
&\underline{S}_j^+(\mathbb{D}) = \sum_{i=1}^{M} \mathbb{I}\left(\mathcal{M}_i^{j+}>0\right),\label{lb}\\
&\mathcal{M}_i^{j+}=\sum_{k=1}^{N} \mathbb{I}\left(\left| \left(\bm{r}_j^c\right)^\mathrm{T}\bm{p}_i+\overline{t}_j^+-q_k^j \right| \leq\epsilon\right),\\
&\underline{S}_j^-(\mathbb{D}) = \sum_{i=1}^{M} \mathbb{I}\left(\mathcal{M}_i^{j-}>0\right),\label{ld}\\
&\mathcal{M}_i^{j-}=\sum_{k=1}^{N} \mathbb{I}\left(\left| -\left(\bm{r}_j^c\right)^\mathrm{T}\bm{p}_i+\overline{t}_j^--q_k^j \right| \leq\epsilon\right),
\end{align}
\end{subequations}
where $\overline{t}_j^+$ is the max-stabbing position of the SPCR upper bound for $\mathbb{S}^{2+}$, and $\overline{t}_j^-$ is the max-stabbing position of the SPCR upper bound for $\mathbb{S}^{2-}$.
\end{proposition}
\begin{proof}
The proof is similar to the proof of Proposition~\ref{Proposition5}, hence we omit it.
\end{proof}

\begin{proposition}[SPCR bound convergence]\label{Proposition9}
When the sub-branch of the square-shaped domain $\mathbb{D}$ collapses to a single point whose center is $\bm{d}^c\in\mathbb{R}^{2}$ and half-side length is zero, we can have 
\begin{equation}
\overline{S}_j(\mathbb{D}) = \underline{S}_j(\mathbb{D}).
\end{equation}
\end{proposition}
\begin{proof}
The proof is similar to the proof of Proposition~\ref{Proposition6}, hence we omit it.
\end{proof}

To improve the efficiency, we can only solve the first sub-problem (e.g., $\max S_X(\bm{r}_X,t_X|\mathcal{P},\mathcal{Q},\epsilon)$) using the extended BnB-based SPCR approach. Then we can solve the second sub-problem (e.g., $\max E_Y(\bm{r}_Y,t_Y|\mathcal{K},\epsilon)$) and third sub-problem (e.g., $\max E_Z(\bm{r}_Z,t_Z|\mathcal{K},\epsilon)$) by Algorithm 1 in Appendix C. This is because we can obtain the set of candidate inlier correspondences $\mathcal{K}$ after solving the first correspondence-free sub-problem, which is implicitly determined by the residual projection constraint. Notably, partial outliers occasionally satisfy this constraint and cannot be removed. However, the proposed correspondence-based method can be applied to robustly address the two remaining sub-problems and generate the final consensus set.

\section{Experiments}\label{Experiments}
This section presents a comprehensive comparison of the proposed method with SOTA correspondence-based methods on both synthetic and real-world datasets. Additionally, we evaluate the extended method against existing SPCR methods, specifically on synthetic data. We implement the proposed method in Matlab 2019b and conduct all experiments on a laptop with an i7-9750H CPU and 16GB RAM. 

\begin{figure*}
  \centering
  \begin{tabular}{  c  c  c  }
 
    \begin{minipage}[b]{0.32\textwidth}
		\centering
		\raisebox{-.5\height}{\includegraphics[width=\linewidth]{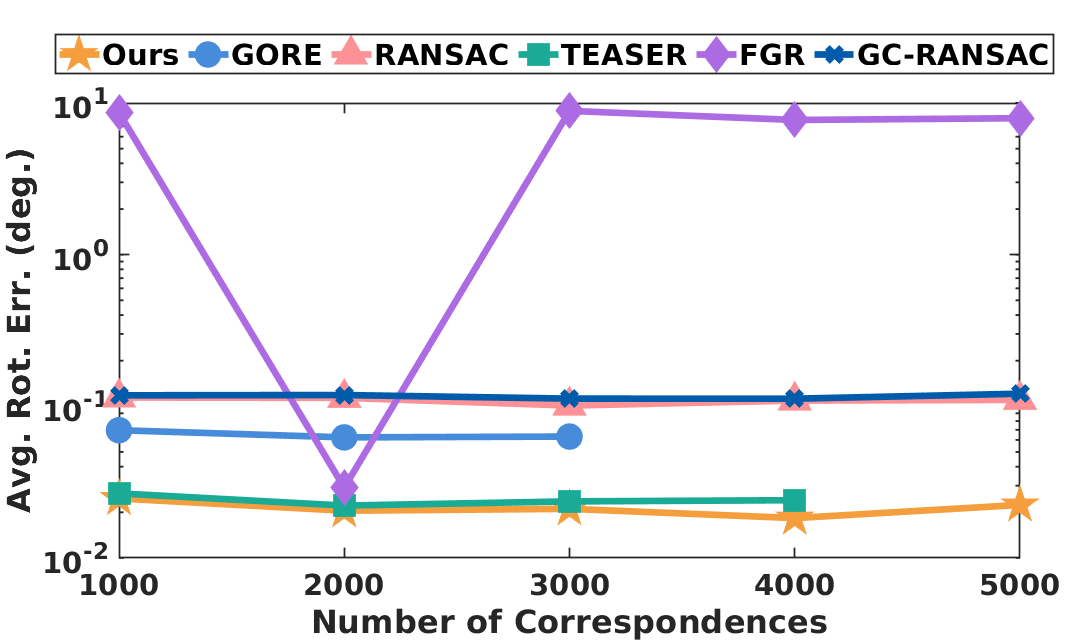}}
	\end{minipage}
    & \begin{minipage}[b]{0.32\textwidth}
		\centering
		\raisebox{-.5\height}{\includegraphics[width=\linewidth]{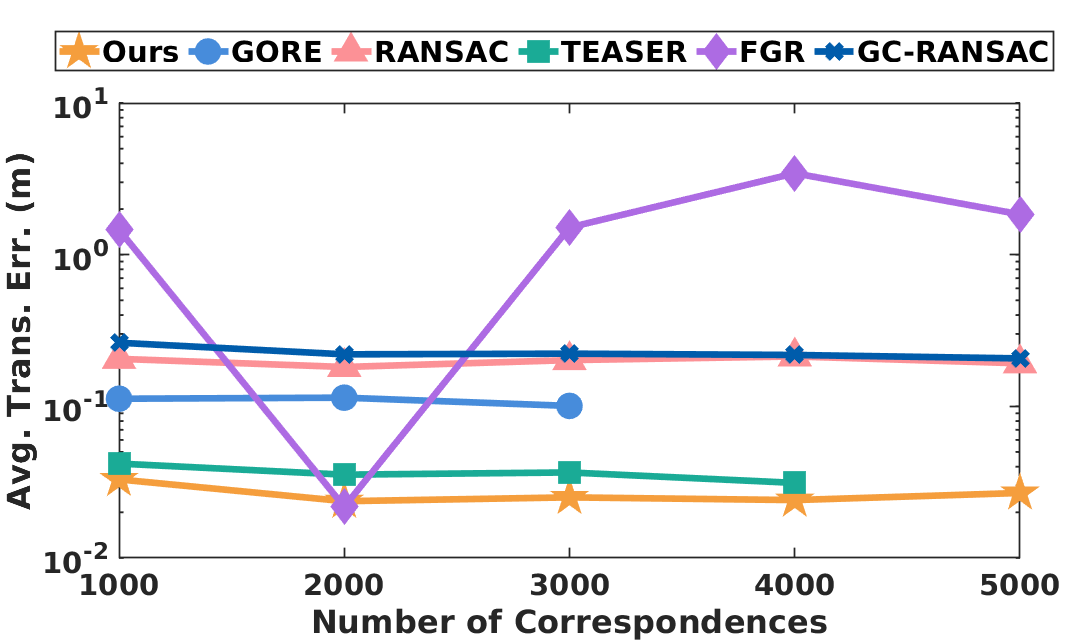}}
	\end{minipage}
    & \begin{minipage}[b]{0.32\textwidth}
		\centering
		\raisebox{-.5\height}{\includegraphics[width=\linewidth]{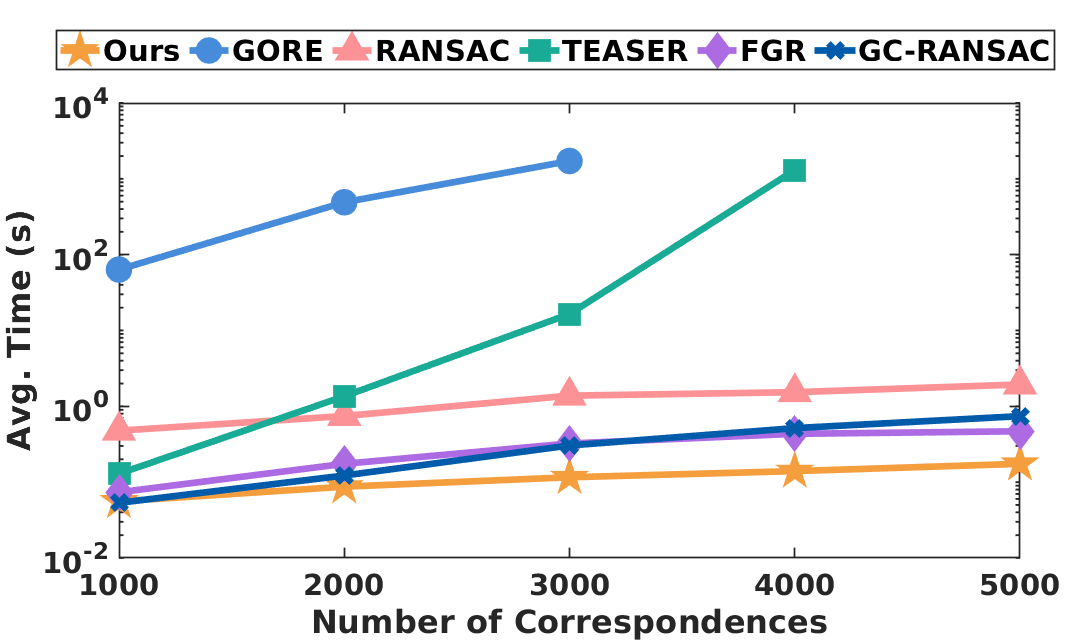}}
	\end{minipage}
    \\
     \footnotesize{(a) Average rotation error} & \footnotesize{(b) Average translation error} & \footnotesize{(c) Average running time}
  \end{tabular}
  \caption{Controlled experiments with \(N=\{1000,2000,\dots,5000\}\). The results include average rotation errors, average translation errors, and average running times.}
  \label{fig_5}
\end{figure*}

\begin{figure*}
  \centering
  \begin{tabular}{  c  c  c  }
 
    \begin{minipage}[b]{0.32\textwidth}
		\centering
		\raisebox{-.5\height}{\includegraphics[width=\linewidth]{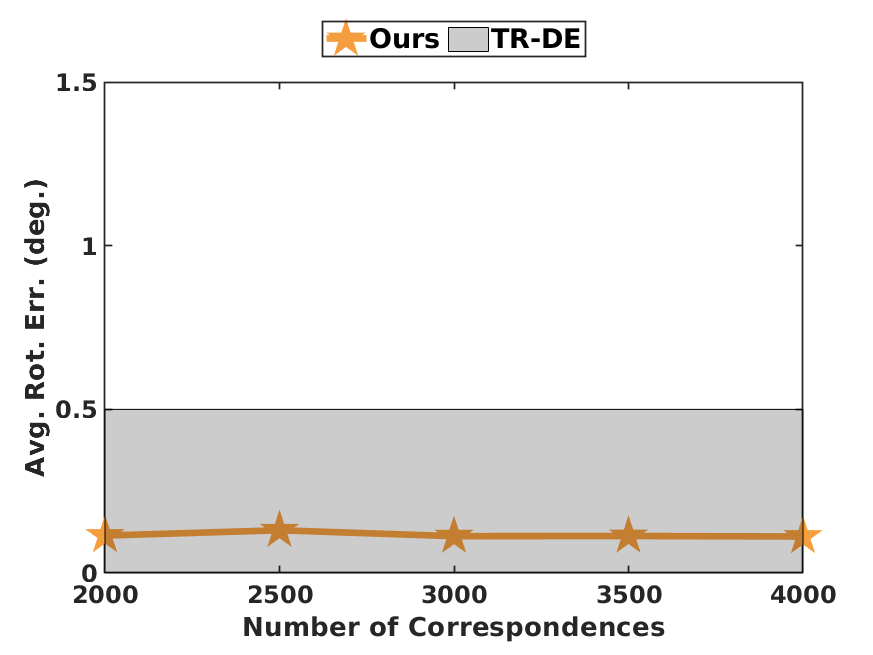}}
	\end{minipage}
    & \begin{minipage}[b]{0.32\textwidth}
		\centering
		\raisebox{-.5\height}{\includegraphics[width=\linewidth]{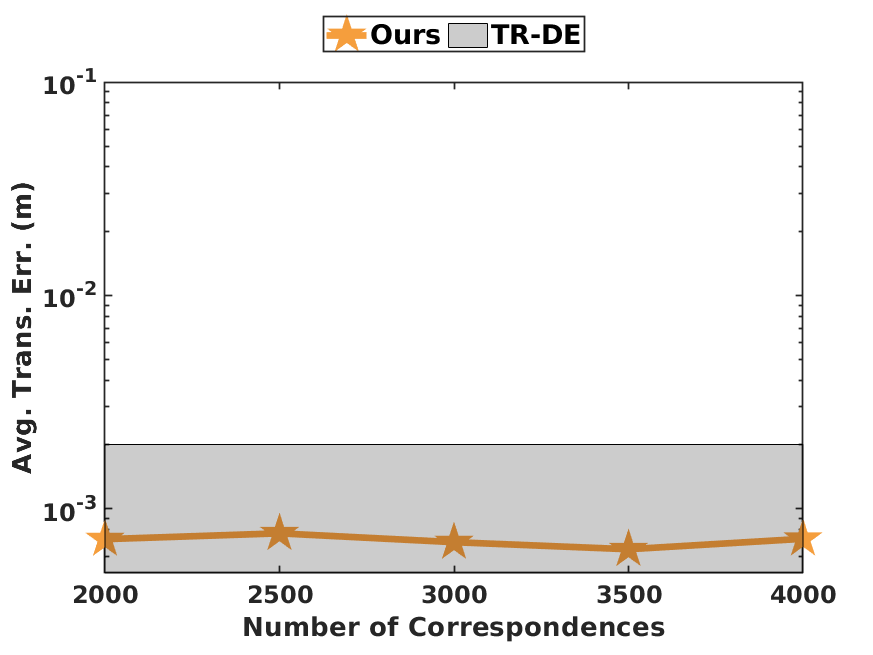}}
	\end{minipage}
    & \begin{minipage}[b]{0.32\textwidth}
		\centering
		\raisebox{-.5\height}{\includegraphics[width=\linewidth]{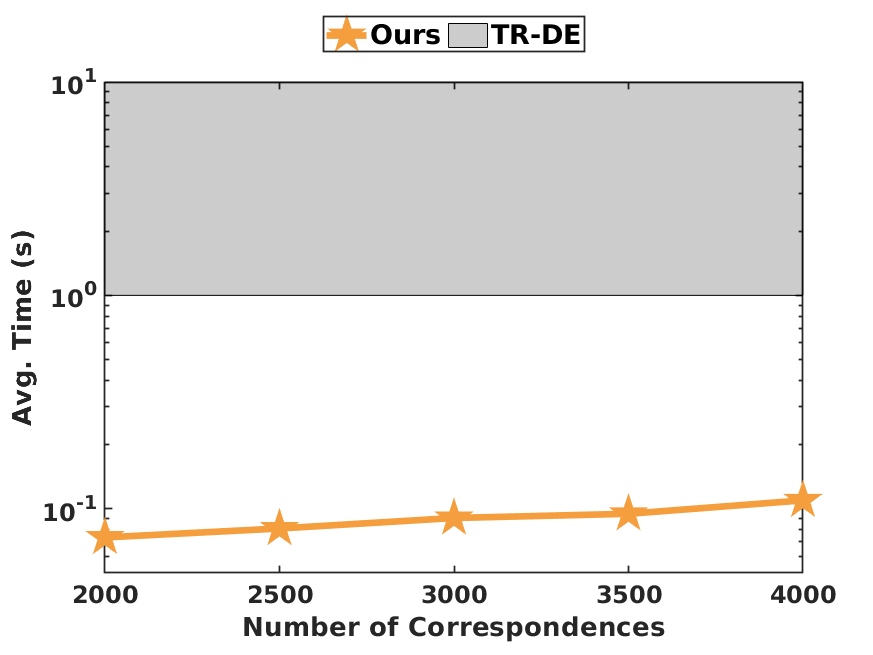}}
	\end{minipage}
 \\
    \footnotesize{(a) Average rotation error} & \footnotesize{(b) Average translation error} & \footnotesize{(c) Average running time}
  \end{tabular}
  \caption{Controlled experiments with the same experimental conditions as TR-DE\cite{chen2022deterministic}. We use the gray rectangular region to approximately represent the results of TR-DE given in \cite{chen2022deterministic}. The results include average rotation errors, average translation errors, and average running times.}
  \label{fig_6}
\end{figure*}
\begin{table*}\footnotesize
\setlength{\tabcolsep}{7pt} 
\renewcommand{\arraystretch}{1.2} 
\caption{Controlled experiments with the extremely high number of correspondences. The results include average rotation error ($\degree$) $|$ average translation error ($\si{\metre}$) $|$ average running time (s). Bolded and underlined fonts indicate the first two best values.
\label{table1}}
\centering
\begin{tabular}{c c c c c c c}
\hline
{\makecell[c]{Number of \\ correspondences}} & {$10k$} & {$20k$} & {$50k$} & {$100k$} & {$200k$} & {$500k$}\\\hline
{GORE\cite{bustos2017guaranteed}} & {$>$1 hour} & ~ & ~ & ~ & ~ & ~\\
{TEASER\cite{yang2020teaser}} & {out-of-memory} & ~ & ~ & ~ & ~ & ~\\
{RANSAC\cite{fischler1981random}} & {0.121$|$0.152$|$4.506} & 0.086$|$0.185$|$6.732 & 0.073$|$0.153$|$12.51 & 0.079$|$0.137$|$30.13 & 0.080$|$0.143$|$92.37 & 4.961$|$4.883$|$315.7\\
{GC-RANSAC\cite{barath2021graph}} & 0.111$|$0.176$|$2.636 & 0.100$|$0.200$|$9.976 & 122.4$|$141.3$|$20.64 & 139.3$|$142.0$|$20.64 & 126.3$|$138.4$|$\underline{20.64} & 133.4$|$148.5$|$\underline{20.64}\\
{FGR\cite{zhou2016fast}} & \underline{0.021}$|$\textbf{0.010}$|$\underline{1.540} & \underline{0.024}$|$\textbf{0.013}$|$\underline{2.477} & \underline{0.037}$|$\textbf{0.022}$|$\underline{6.346} & \underline{0.031}$|$\textbf{0.018}$|$\underline{12.13} & \underline{0.024}$|$\textbf{0.018}$|$23.97 & \underline{0.047}$|$\textbf{0.025}$|$68.74 \\
{Ours} & \textbf{0.016}$|$\underline{0.017}$|$\textbf{0.397} & \textbf{0.022}$|$\underline{0.028}$|$\textbf{0.649} & \textbf{0.025}$|$\underline{0.025}$|$\textbf{1.649} & \textbf{0.025}$|$\underline{0.028}$|$\textbf{2.963} & \textbf{0.023}$|$\underline{0.027}$|$\textbf{6.111} & \textbf{0.018}$|$\textbf{0.025}$|$\textbf{15.56} \\\hline
\end{tabular}
\end{table*}
\subsection{Experimental Setting}\label{Experimental Setting}
We denote the proposed method as \textbf{Ours}. The compared methods for correspondence-based registration are as follows,
\begin{itemize}
    \item GORE\cite{bustos2017guaranteed}: A guaranteed outlier removal registration method based on BnB and pose decoupling. 
    \item RANSAC\cite{fischler1981random}: A typical consensus maximization registration approach. The maximum number of iterations is set to $10^4$. 
    \item TEASER\cite{yang2020teaser}: A certifiable decoupling-based registration method with a robust cost function.
    \item FGR\cite{zhou2016fast}: A fast registration method with a robust cost function.
    \item GC-RANSAC\cite{barath2021graph}: A variant of RANSAC-based registration method with improvements in local optimization. The maximum number of iterations is set to $10^4$.
    \item TR-DE\cite{chen2022deterministic}: A deterministic point cloud registration method based on BnB and pose decoupling. 
    \item SC$^2$-PCR\cite{chen2022sc2}: A robust registration method that proposes a second-order spatial compatibility measurement to identify inliers.
    \item MAC\cite{zhang20233d}: A robust registration method with maximal cliques.
    \item DGR\cite{choy2020deep}: A learning-based outlier rejection method employing the sparse convolutional network. 
    \item PointDSC\cite{bai2021pointdsc}: A learning-based outlier rejection method utilizing the spatial consistency. 
    \item Hunter\cite{10246849}: A learning-based robust transformation estimation method utilizing the high-order consistency. 
\end{itemize}

Besides, the compared methods for SPCR are as follows,
\begin{itemize}
    \item GO-ICP\cite{7368945}: A 6-DOF global optimal registration method based on BnB. 
    \item GO-ICPT: A variant of GO-ICP with outlier trimming.
    \item ICP\cite{121791}: A typical EM-type method implemented by \textit{pcregistericp} function in MATLAB.
    \item CPD\cite{myronenko2010point}: A GMM-based probabilistic registration approach.
    \item GMMReg\cite{jian2010robust}: A robust and general GMM-based registration method.
\end{itemize}

Similar to \cite{yang2020teaser,chen2022deterministic,bai2021pointdsc}, the evaluation metrics for point cloud registration in this study include 1) rotation error $E_{\bm{R}}$, 2) translation error $E_{\bm{t}}$, 3) running time, 4) success rate $SR$, and 5) $F1$-score. The error definitions are as follows:
\begin{subequations}
\begin{align}
    &E_{\bm{R}}=\arccos{\left(\frac{Tr(\bm{R}_{gt}^{-1}\bm{R}^*)-1}{2}\right)},
\\
    &E_{\bm{t}}=\|\bm{t}_{gt}-\bm{t}^*\|,
\end{align}
\end{subequations}
where \(\bm{t}_{gt}\) and \(\bm{R}_{gt}\) are the ground truth, \(\bm{t}^*\) and \(\bm{R}^*\) are the estimated solutions, and $Tr(\cdot)$ is the trace of a matrix. The successful cases must satisfy the predefined threshold for $E_{\bm{R}}$ and $E_{\bm{t}}$. Besides, the definition of $F1$-score is given in \cite{bai2021pointdsc}.

\subsection{Synthetic Data Experiments}\label{Synthetic}

In this section, we conduct various experiments on synthetic data to compare the performance of the proposed method with SOTA correspondence-based and correspondence-free registration methods.

\begin{figure*}
  \centering
  \begin{tabular}{  c  c  c  }
 
    \begin{minipage}[b]{0.32\textwidth}
		\centering
		\raisebox{-.5\height}{\includegraphics[width=\linewidth]{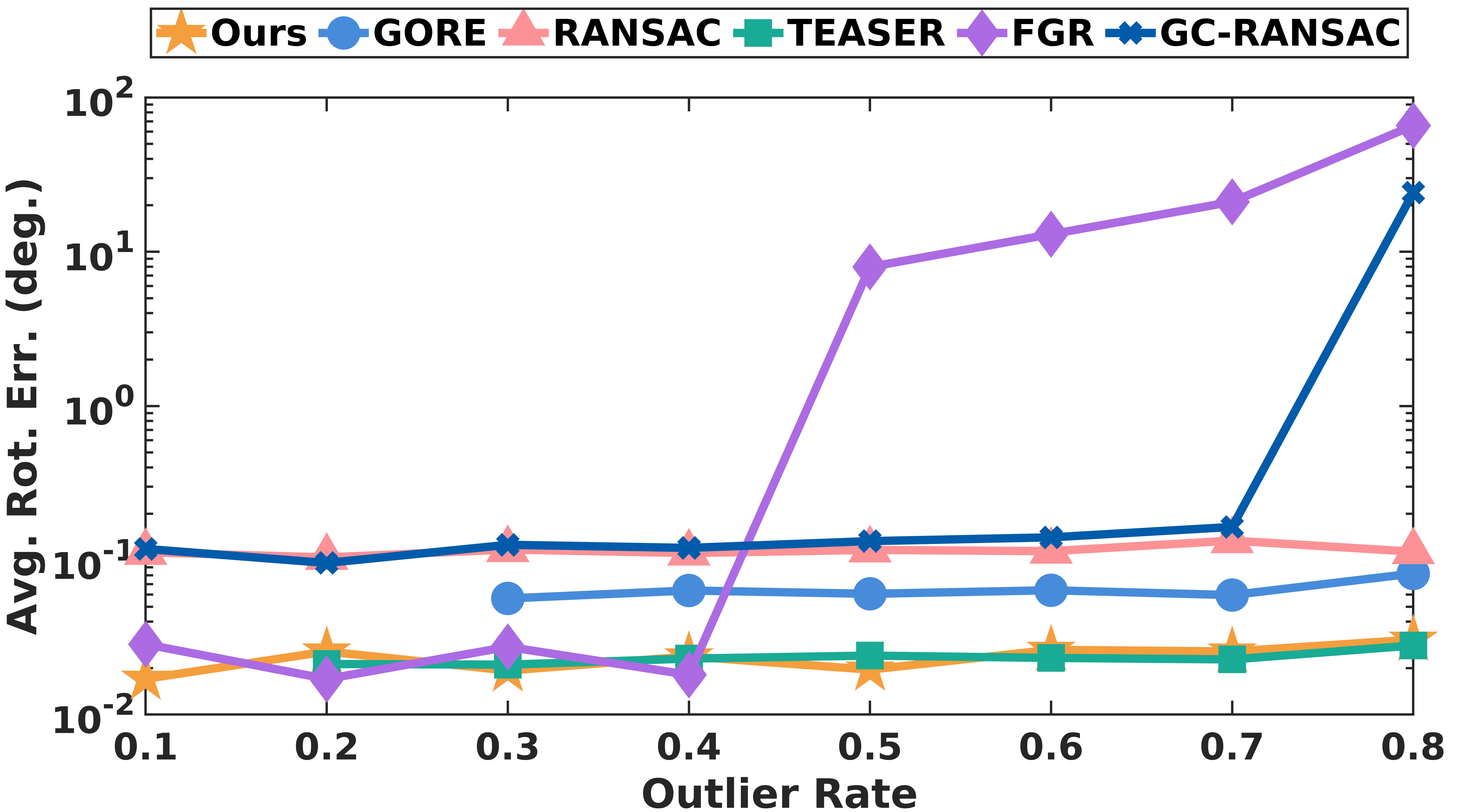}}
	\end{minipage}
    & \begin{minipage}[b]{0.32\textwidth}
		\centering
		\raisebox{-.5\height}{\includegraphics[width=\linewidth]{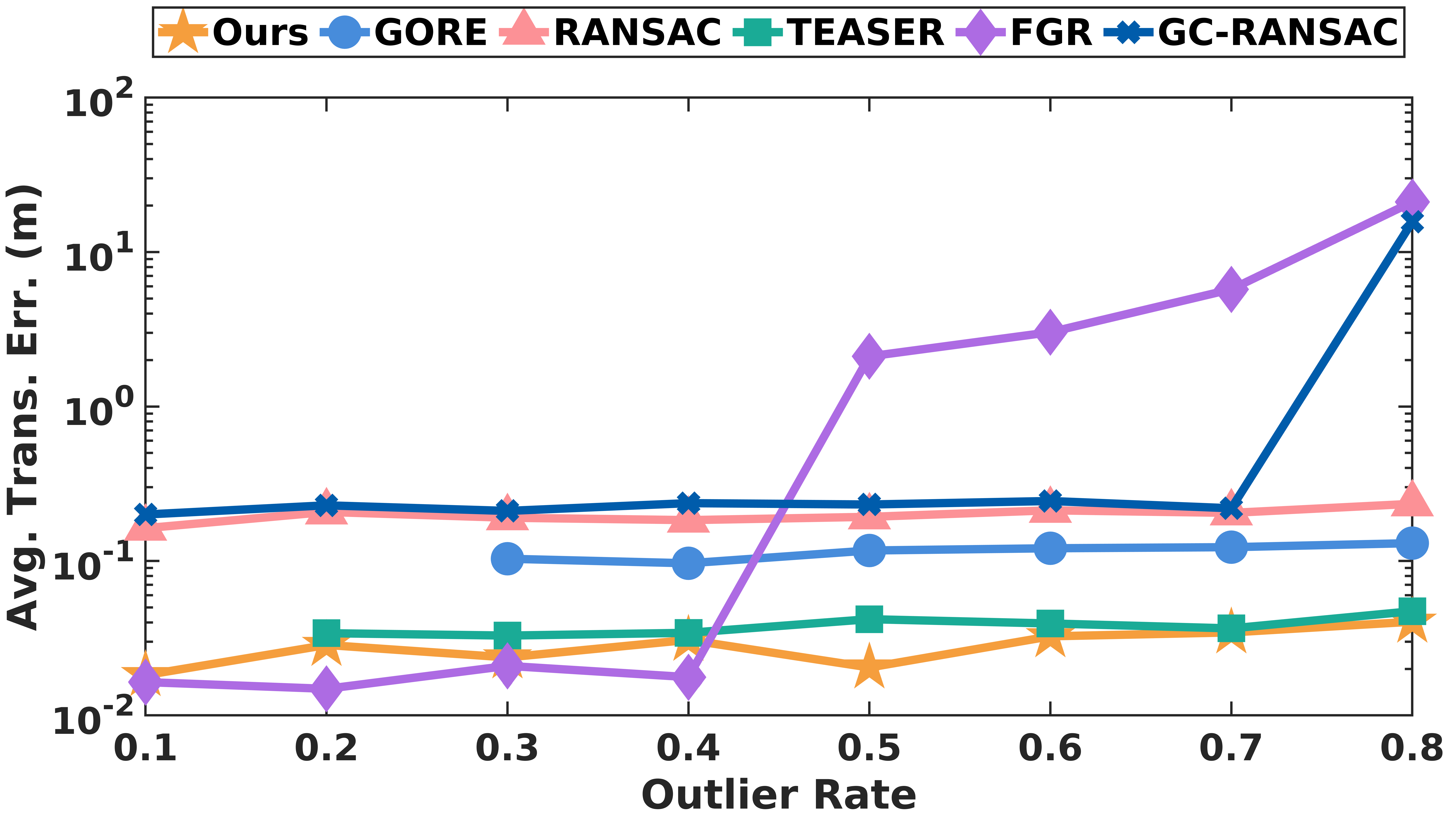}}
	\end{minipage}
    & \begin{minipage}[b]{0.32\textwidth}
		\centering
		\raisebox{-.5\height}{\includegraphics[width=\linewidth]{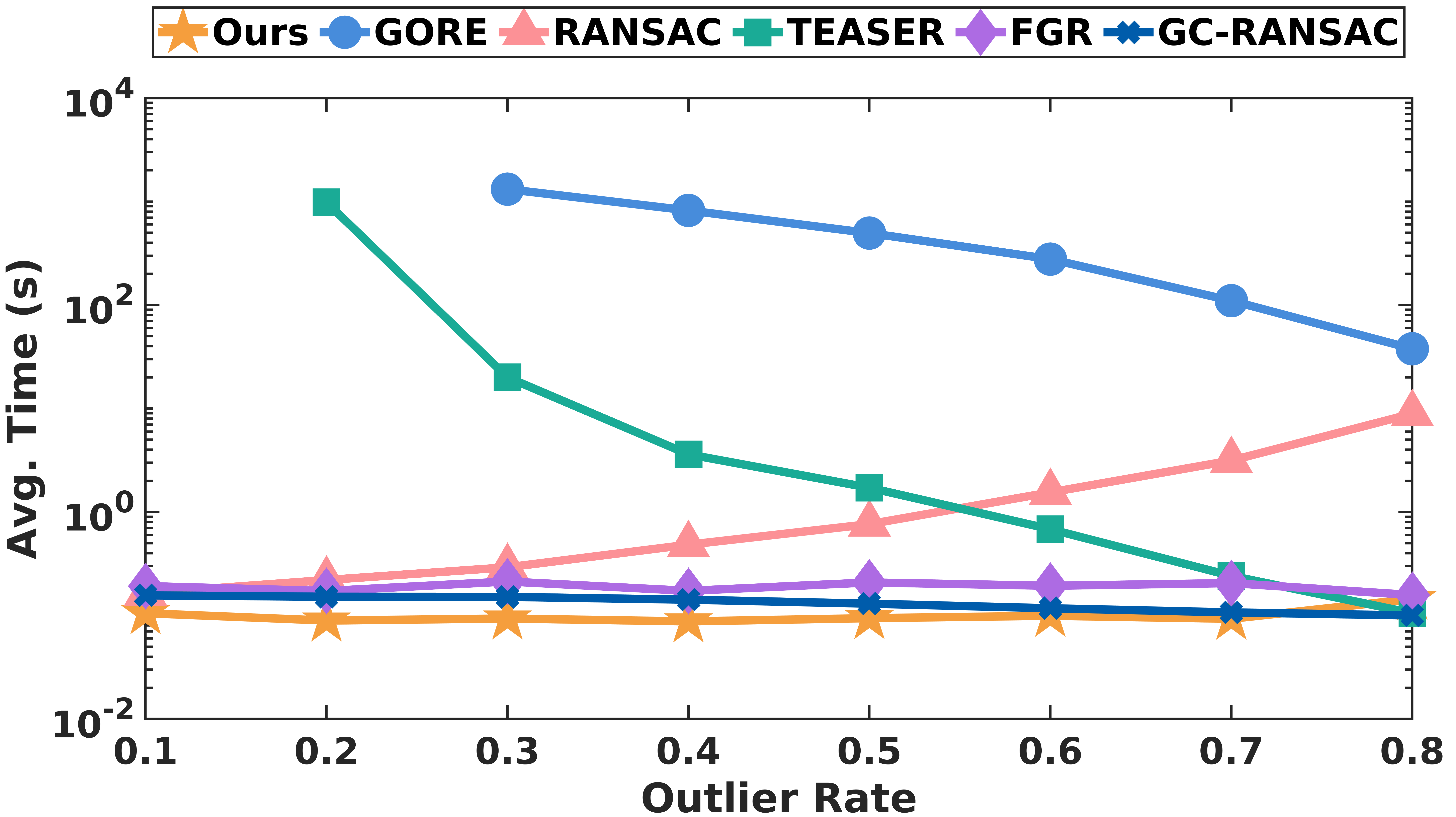}}
	\end{minipage}
 \\
     \footnotesize{(a) Average rotation error} & \footnotesize{(b) Average translation error} & \footnotesize{(c) Average running time}
  \end{tabular}
  \caption{Controlled experiments with \(\eta=\{0.1,0.2,\dots,0.8\}\). The results include average rotation errors, average translation errors, and average running times.}
  \label{fig_7}
\end{figure*}

\begin{figure*}
  \centering
  \begin{tabular}{  c  c  c  }
 
    \begin{minipage}[b]{0.32\textwidth}
		\centering
		\raisebox{-.5\height}{\includegraphics[width=\linewidth]{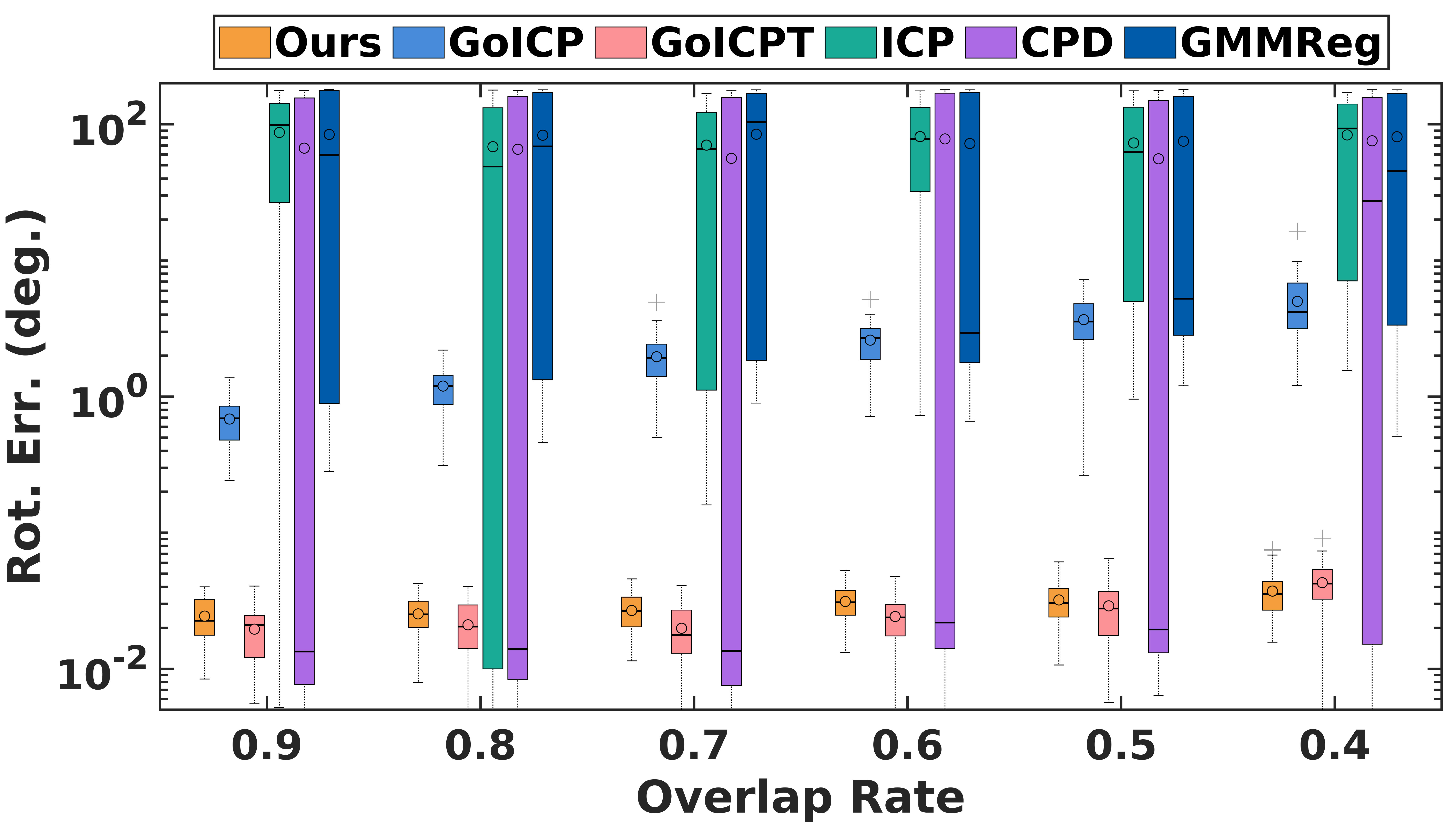}}
	\end{minipage}
    & \begin{minipage}[b]{0.32\textwidth}
		\centering
		\raisebox{-.5\height}{\includegraphics[width=\linewidth]{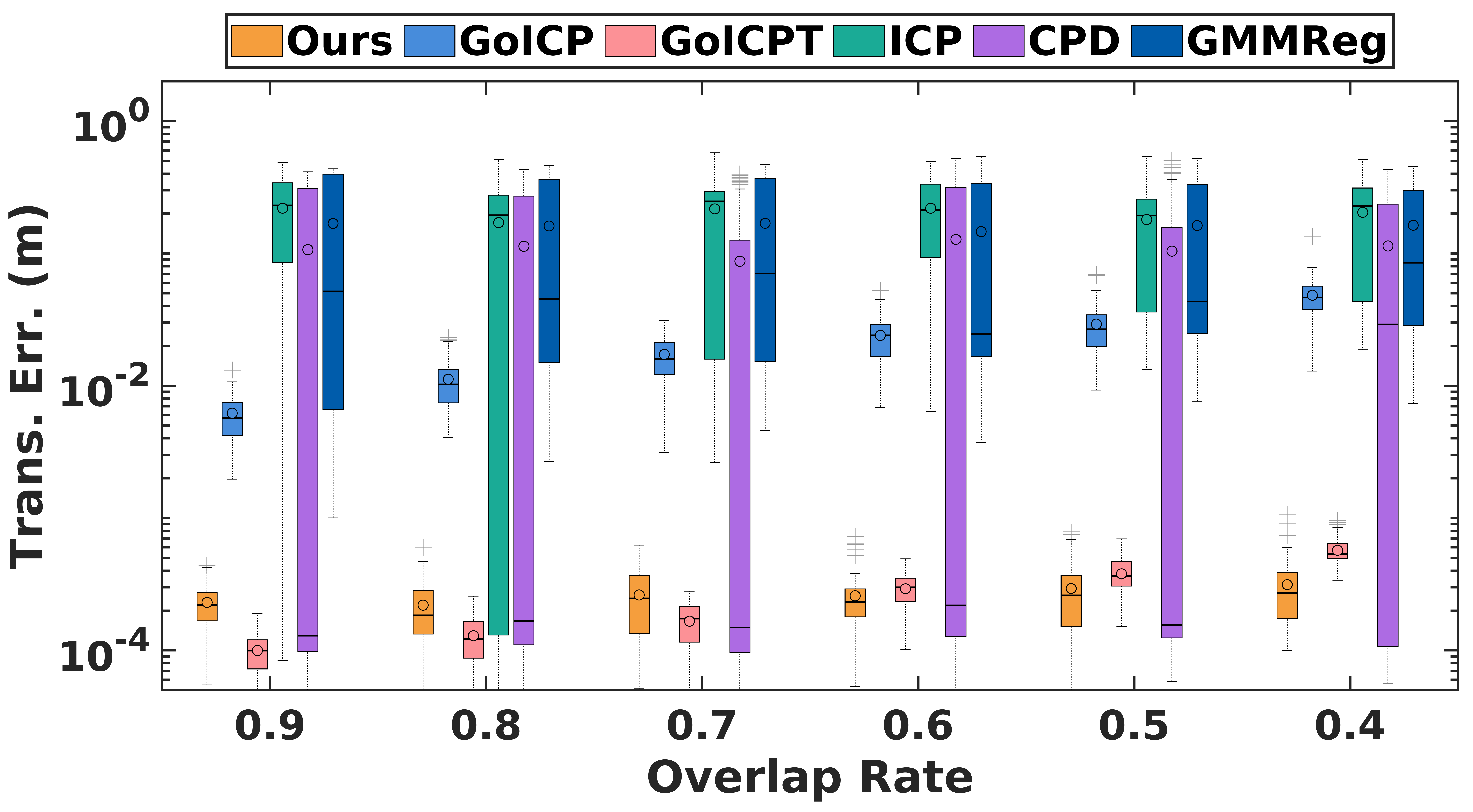}}
	\end{minipage}
    & \begin{minipage}[b]{0.32\textwidth}
		\centering
		\raisebox{-.5\height}{\includegraphics[width=\linewidth]{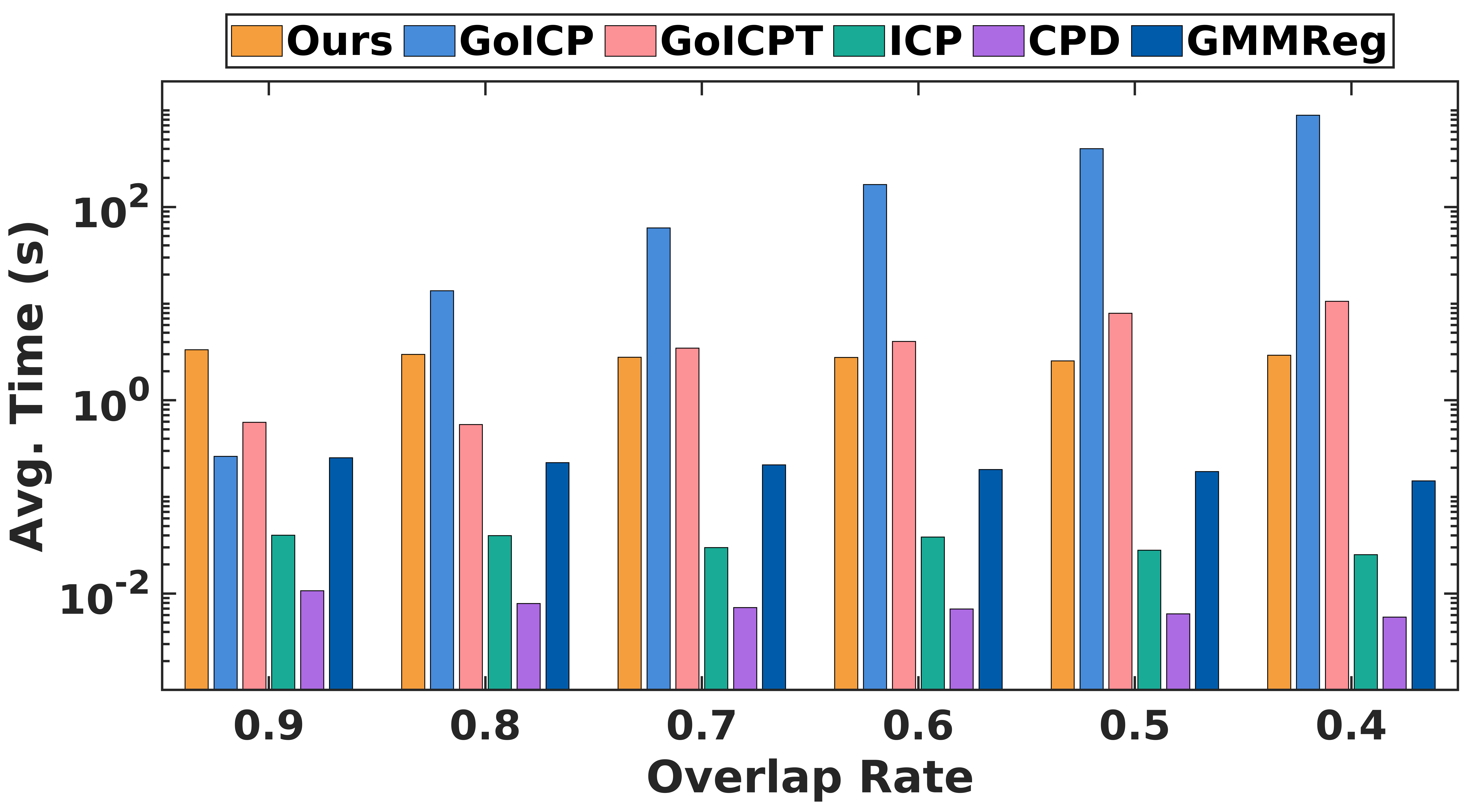}}
	\end{minipage}
 \\
     \footnotesize{(a) Rotation error} & \footnotesize{(b) Translation error} & \footnotesize{(c) Average running time}
  \end{tabular}
  \caption{Controlled SPCR experiments with \(\rho=\{0.9,0.8,\dots,0.4\}\). The results include rotation errors, translation errors, and average running times.}
  \label{fig_8}
\end{figure*}

\subsubsection{Data generation}

First, we randomly generate the source point cloud $\mathcal{P}$ in the cube \([-100,100]^3\). The source point cloud is transformed by a random rotation \(\bm{R}_{gt}\in \mathbb{SO}(3)\) and a random translation \(\bm{t}_{gt}\in[-100,100]^3\) to generate the target point cloud $\mathcal{Q}$. Then, a portion of points in the target point cloud is replaced by arbitrarily generated points to simulate outliers. The outlier rate $\eta$ is the ratio of these replaced points to all points. Besides, zero-mean Gaussian noise with standard deviation $\sigma$ is added to the target point cloud. Notably, the inlier threshold in each synthetic data experiment is set according to the standard deviation of the noise. Moreover, synthetic data experiments are conducted in Appendix E to theoretically assess the sensitivity of the proposed method to the inlier threshold.

\subsubsection{Efficiency and accuracy experiments}

This section presents three sets of experiments comparing the efficiency and accuracy of Ours with GORE, RANSAC, TEASER, FGR, GC-RANSAC, and TR-DE. Rotation errors, translation errors, and time costs are recorded for each experiment group. The first group focuses on experiments with a regular number of correspondences. We randomly generate \(N=\{1000,2000,\dots,5000\}\) correspondences with a noise level of $\sigma=0.5$ and an outlier rate of $\eta=0.5$. The experiment is repeated $50$ times for each setting, and the average results are depicted in Fig.~\ref{fig_5}. It is worth noting that results are not reported when the running time exceeds $1800$ seconds. Among the deterministic methods, GORE and TEASER exhibit relatively high accuracy. However, their time costs increase significantly as the number of correspondences grows, with TEASER being the fastest in this regard. FGR, on the other hand, demonstrates occasional unsuccessful results but shows high efficiency. RANSAC and GC-RANSAC suffer from lower accuracy due to sampling uncertainty. Nevertheless, they exhibit relatively high efficiency at the regular outlier rate ($\eta=0.5$). In contrast, Ours outperforms all other methods in terms of both efficiency and accuracy. When $N$ reaches $4000$, Ours is approximately $10^4$ times faster than GORE and TEASER. This may be explained by the reason that even after outlier rejection, a significant number of candidate inlier correspondences are still retained when dealing with a large number of correspondences. Consequently, the optimization process for GORE and TEASER becomes slower.

Since the code of TR-DE\cite{chen2022deterministic} is not released publicly, we set the same experimental conditions as TR-DE to compare the performance, which is the second group of experiments. Specifically, the source point cloud is randomly generated within the unit cube, and the experiment is conducted with \(N=\{2000,2500,\dots,4000\}\), $\sigma=0.005$, and, $\eta=0.55$. We also conduct $200$ independent trials for each setting and record the average experiment results, as shown in Fig.~\ref{fig_6}. We use the gray rectangular region to approximately represent the results of TR-DE given in \cite{chen2022deterministic}. We can observe that Ours is about $10$ times faster than TR-DE while keeping comparable accuracy.

To further investigate the potential efficiency advantages of Ours, we conduct the third group of experiments, specifically focusing on extremely high numbers of correspondences: \(N=\{10k,20k,50k,100k,200k,500k\}\) (where $k$ denotes one thousand). The remaining settings are consistent with those of the first group. Table~\ref{table1} presents the average rotation errors, average translation errors, and average time costs of each method. The running time of GORE exceeds one hour starting from $N=10k$ thus its results are not reported. Furthermore, TEASER demands a substantial amount of memory space, which renders it unable to operate efficiently under such extreme experimental conditions. As $N$ increases to $500k$, RANSAC yields numerous unsatisfactory solutions and incurs a time cost of up to $315.7$s. Additionally, GC-RANSAC fails to converge to the correct result after $N$ reaches $50k$ due to early termination. In comparison to FGR, Ours delivers more accurate rotation estimates but slightly less accurate translation estimates. However, experimental results indicate that the number of correspondences has a relatively minor impact on the efficiency of our method. For instance, when the number of correspondences increases from $10k$ to $500k$, Ours is approximately $8$ to $20$ times faster than RANSAC and roughly $4$ times faster than FGR. Overall, the proposed method exhibits superior efficiency while maintaining competitive accuracy compared to SOTA approaches. 

\subsubsection{Robustness experiments}

In this section, we conduct a group of controlled experiments to compare the robustness of Ours with GORE, RANSAC, TEASER, FGR, and GC-RANSAC. We randomly generate \(N=2000\) correspondences with varying outlier rates $(\eta=\{0.1,0.2,\dots,0.8\})$ and a noise level of $\sigma=0.5$. The experiment is repeated $50$ times for each setting. The average rotation errors, average translation errors, and average time costs for each method are reported in Fig.~\ref{fig_7}. Results beyond a running time of $1800$ seconds are not recorded in this group of experiments. Comparing the registration errors demonstrates that Ours, GORE, RANSAC, and TEASER are robust against up to $80\%$ outlier rate. RANSAC has relatively higher registration errors than Ours, GORE, and TEASER. Moreover, the running time of RANSAC increases significantly with an increase in the outlier rate. In contrast, both GORE and TEASER display a significant decrease in running times as the outlier rate increases due to a corresponding reduction in the number of inliers. This indicates that, for GORE and TEASER, the time required for outlier removal is considerably smaller compared to the time spent on the optimization part. Consequently, they exhibit lower efficiency at regular outlier rates (e.g., $\eta\leq0.5$). On the other hand, despite the high efficiency exhibited by both the deterministic FGR and the non-deterministic GC-RANSAC, they do not perform well when confronted with high outlier rates (e.g., $\eta\geq0.7$). In contrast, Ours stands out as one of the fastest and most robust methods. 

\subsubsection{Challenging SPCR experiments}\label{Challenging}

In this section, we evaluate the performance of our extended simultaneous pose and correspondence registration (SPCR) method against GoICP, GoICPT, ICP, CPD, and GMMReg using the Bunny dataset\cite{curless1996volumetric}. The Bunny dataset consists of $35947$ points and is pre-normalized to fit within the cube $[-1,1]^3$, as required by GoICP\cite{7368945}. Similar to \cite{yang2020teaser}, we down-sample the Bunny dataset to $M=100$ points, which serve as the source point cloud $\mathcal{P}$. To generate the target point cloud $\mathcal{Q}$, we apply a random rotation and translation to the source point cloud. Additionally, we randomly remove a certain proportion of points from $\mathcal{Q}$ to simulate partial overlap between $\mathcal{P}$ and $\mathcal{Q}$. The visualization results for a pair of synthetic data are shown in Fig.~\ref{fig_1}(d-1), where the bolded points represent the down-sampled point clouds. Furthermore, we add zero-mean Gaussian noise with $\sigma=0.001$ to the source point cloud $\mathcal{P}$. The registration experiment is repeated $50$ times for each overlap rate in $\rho=\{0.9,0.8,\dots,0.4\}$. Notably, the trimming fraction of GoICPT is set to be identical to the overlap rate. 

The registration errors and average running times for each approach are presented in Fig.~\ref{fig_8}. Notably, the running times of building distance transform (DT)\cite{7368945} for GoICP and GoICPT are not recorded and approximately take $23$s on average. During repeated experiments, the local methods, ICP, CPD, and GMMReg, tend to converge to local optima, resulting in incorrect results with large registration errors. However, their efficiency remains a notable advantage. In contrast, the global methods GoICP and its variant GoICPT demonstrate greater robustness than these local methods. In particular, with a precisely tuned trimming ratio, GoICPT achieves remarkably high accuracy across all overlap rates. Nevertheless, these global methods suffer from relatively slow running times, which increase more rapidly than our proposed method. Consequently, when the overlap ratio is low (e.g., $\rho\leq0.7$), Ours is faster than GoICP and GoICPT. As a global method, Ours also falls short in efficiency compared to the local methods. However, Ours is more robust than local methods such as ICP, CPD, and GMMReg. Furthermore, as depicted in Fig.~\ref{fig_1}(d), Ours exhibits greater robustness than ICP and higher efficiency than GoICP on a randomly generated pair of Bunny data ($\rho=0.6$). These experiments illustrate the potential practicality of our proposed approach in addressing the challenging SPCR problem and its strength in terms of robustness and efficiency. 

\subsection{Real-World Data Experiments}

In this section, to assess the performance of the proposed method on real-world data, we conduct experiments using the Bremen dataset\cite{borrmann2013thermal}, ETH dataset\cite{theiler2014keypoint}, and KITTI dataset\cite{geiger2012we}. These datasets depict challenging outdoor scenarios, with the former two captured using terrestrial LiDAR and the latter collected from onboard LiDAR.

\subsubsection{Bremen dataset experiments}\label{Bremen}

\begin{figure}

\centering
\begin{tabular}{  c  }
 
    \begin{minipage}[b]{0.85\columnwidth}
		\centering
		\raisebox{-.5\height}{\includegraphics[width=\columnwidth]{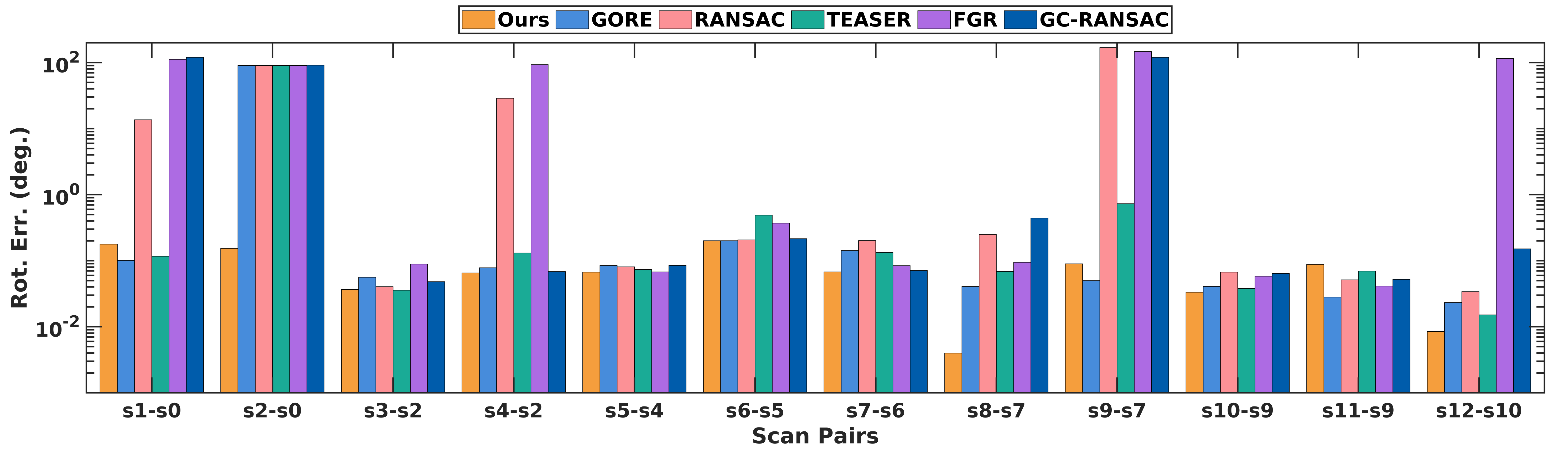}}
	\end{minipage}\\
 \footnotesize{(a) Rotation error}
 \\
     \begin{minipage}[b]{0.85\columnwidth}
		\centering
		\raisebox{-.5\height}{\includegraphics[width=\columnwidth]{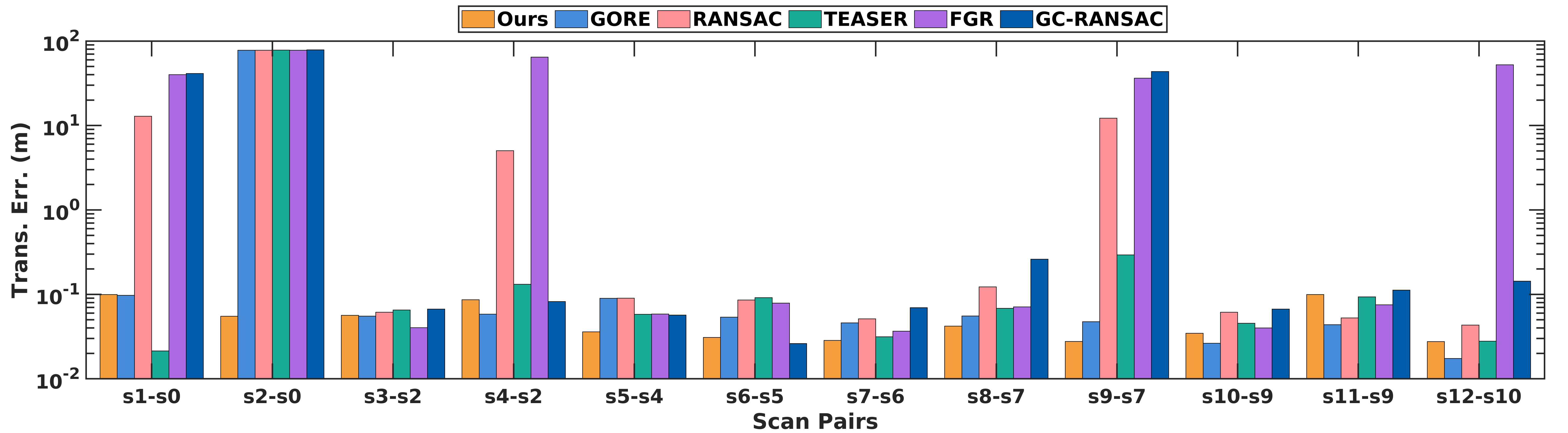}}
	\end{minipage}\\
 \footnotesize{(b) Translation error}
 \\
     \begin{minipage}[b]{0.85\columnwidth}
		\centering
		\raisebox{-.5\height}{\includegraphics[width=\columnwidth]{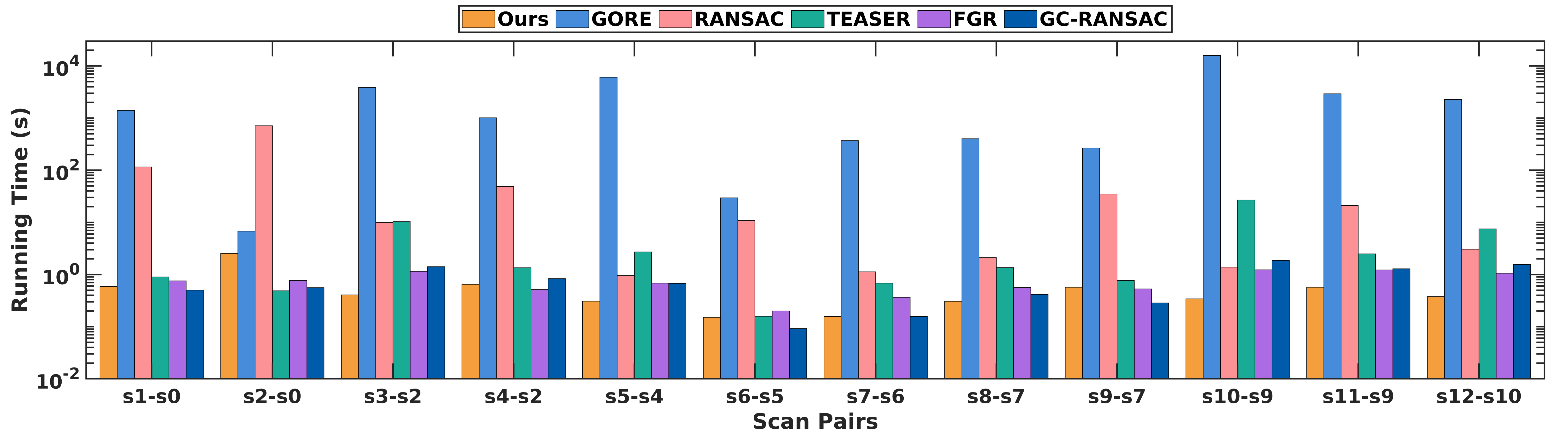}}
	\end{minipage}\\
 \footnotesize{(c) Running time}
  \end{tabular}
\caption{Experiment results on the Bremen dataset\cite{borrmann2013thermal} with the FPFH~\cite{rusu2009fast} descriptor. The results include rotation errors, translation errors, and running times.}
\label{fig_9}
\end{figure}

\begin{figure}
\centering
\begin{tabular}{  c  }
 
    \begin{minipage}[b]{0.9\columnwidth}
		\centering
		\raisebox{-.5\height}{\includegraphics[width=0.98\columnwidth]{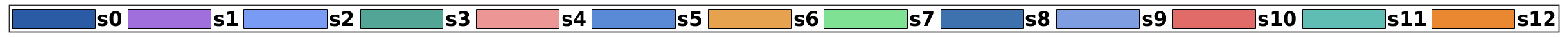}}
	\end{minipage}\\
     \begin{minipage}[b]{0.8\columnwidth}
		\centering
		\raisebox{-.5\height}{\includegraphics[width=0.98\columnwidth]{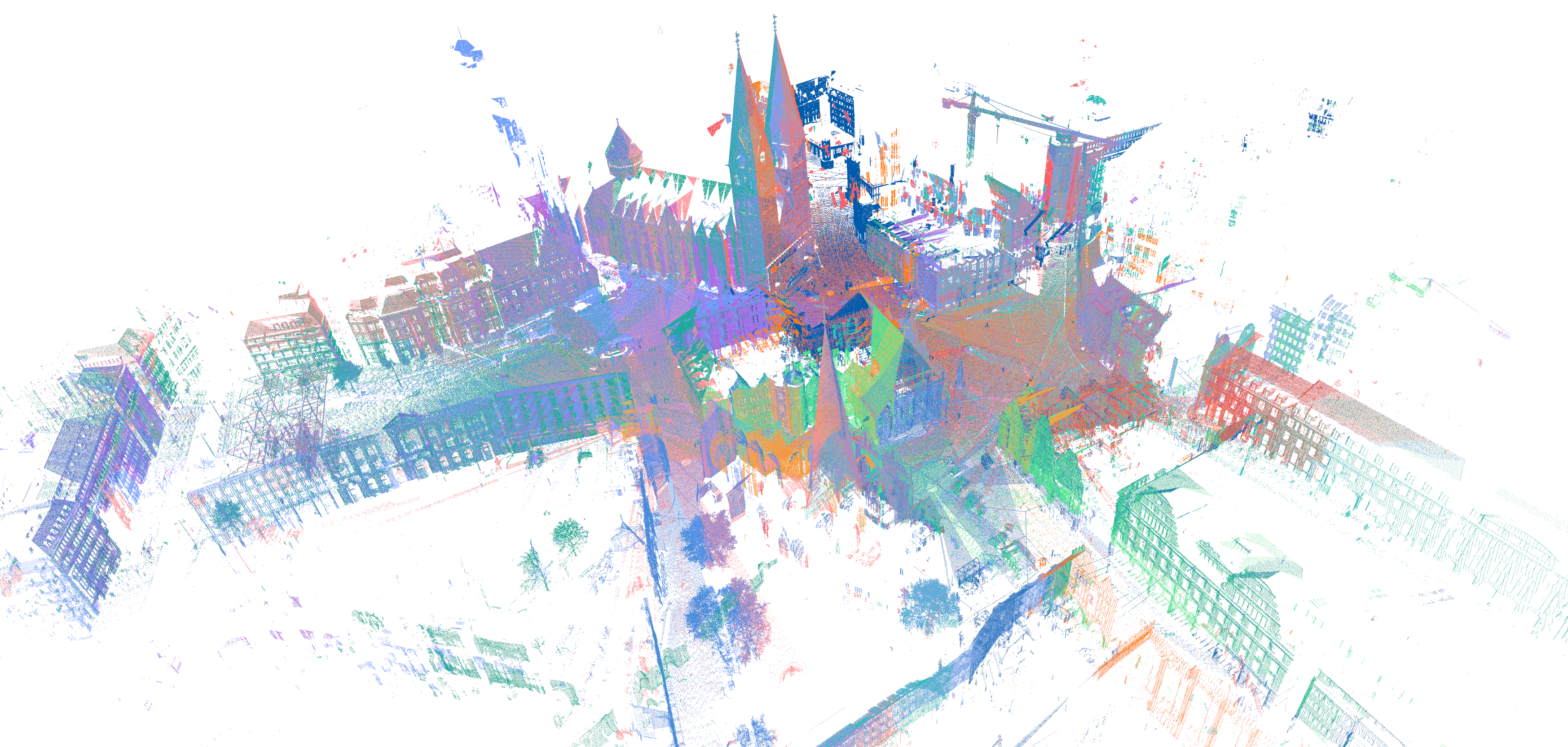}}
	\end{minipage}

  \end{tabular}

\caption{Complete registration results of Ours on the Bremen dataset\cite{borrmann2013thermal}, where different scans are indicated by different colors. The pair-wise point cloud registration is conducted for all $12$ scan pairs.}
\label{fig_10}
\end{figure}

\begin{figure}
\centering
\begin{tabular}{  c  }
 
    \begin{minipage}[b]{0.85\columnwidth}
		\centering
		\raisebox{-.5\height}{\includegraphics[width=\columnwidth]{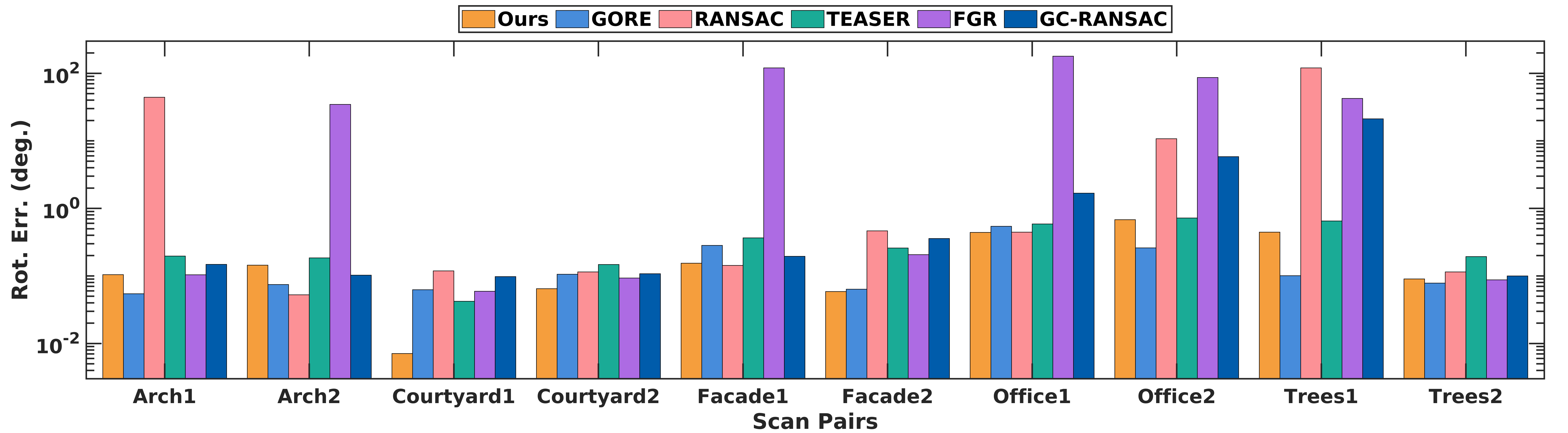}}
	\end{minipage}\\
 \footnotesize{(a) Rotation error}
 \\
     \begin{minipage}[b]{0.85\columnwidth}
		\centering
		\raisebox{-.5\height}{\includegraphics[width=\columnwidth]{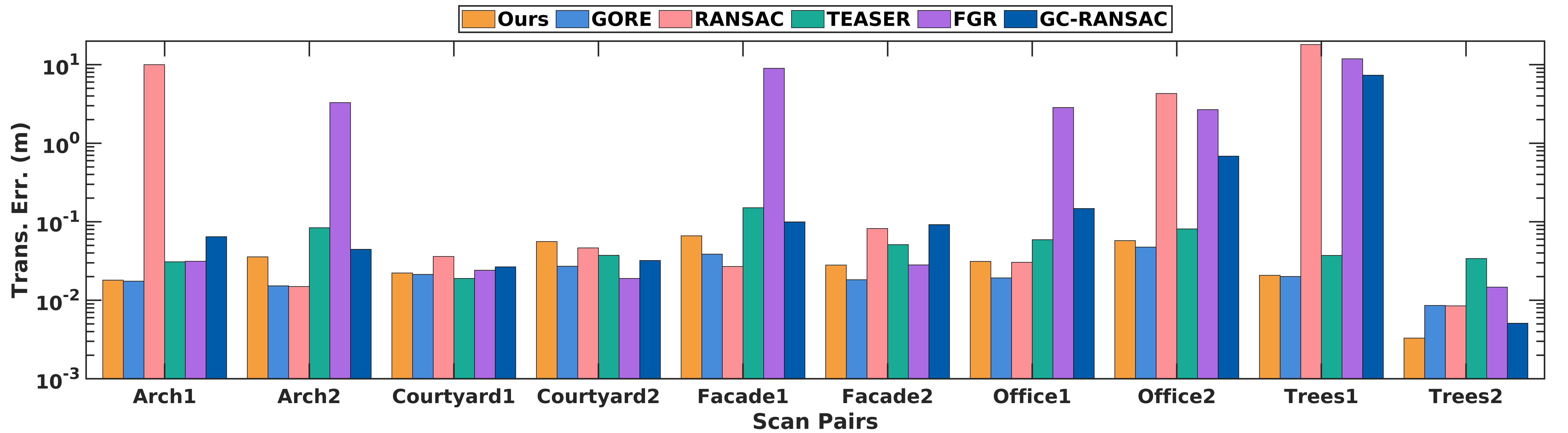}}
	\end{minipage}\\
 \footnotesize{(b) Translation error}
 \\
     \begin{minipage}[b]{0.85\columnwidth}
		\centering
		\raisebox{-.5\height}{\includegraphics[width=\columnwidth]{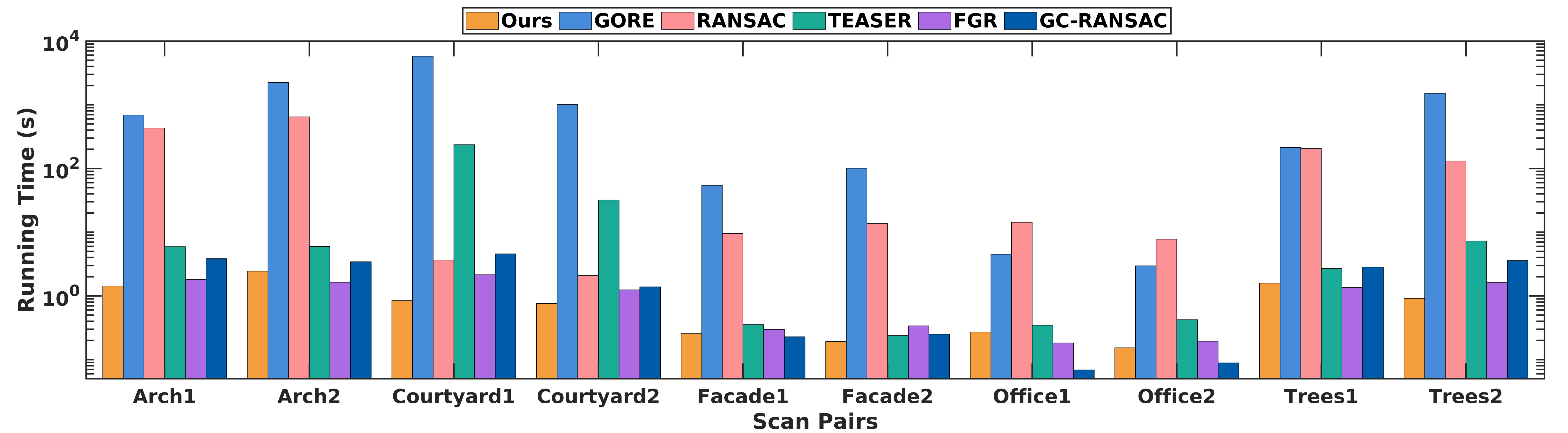}}
	\end{minipage}\\
 \footnotesize{(c) Running time}
  \end{tabular}
\caption{Experiment results on the ETH dataset\cite{theiler2014keypoint} with the FPFH~\cite{rusu2009fast} descriptor. The results include rotation errors, translation errors, and running times.}
\label{fig_11}
\end{figure}

The Bremen dataset\cite{borrmann2013thermal} is a large-scale outdoor dataset with $13$ LiDAR scans. The challenge of aligning the Bremen dataset lies in the large number of repeated features from buildings and streets. Similar to~\cite{9373914,10091912}, we initially down-sample the scans using the voxel grid algorithm\cite{rusu20113d}. Subsequently, we extract ISS\cite{zhong2009intrinsic} keypoints and calculate FPFH\cite{rusu2009fast} descriptors for each keypoint. Through K-nearest neighbor search, we generate the set of putative correspondences $\mathcal{K}$. The ground-truth pose for each scan is provided within the dataset. Since the proposed method is only for pair-wise registration, we construct $12$ scan pairs to register all scans. Appendix D provides detailed information for each scan pair from the Bremen dataset. The down-sampling resolution for the Bremen dataset is set to $0.15\si{\metre}$, which also determines the inlier threshold. With several thousand correspondences, the outlier rate ranges from approximately $90\%$ to $99\%$ for the Bremen dataset. We employ the proposed method (Ours), as well as GORE, RANSAC, TEASER, FGR, and GC-RANSAC, to register these scan pairs. 

The rotation error, translation error, and running time of each method for each scan pair are shown in Fig.~\ref{fig_9}. Notably, when dealing with the registration of the scan pair s2-s0, all compared methods, except for Ours, fail due to the exceptionally high outlier rate ($99.64\%$). GORE and TEASER demonstrate successful alignment with relatively high accuracy for the remaining scan pairs. Despite this, GORE exhibits the highest time cost among all methods, even when the number of correspondences is small or the outlier rate is low, which is consistent with the findings from synthetic data experiments. For instance, in the case of the pair s10-s9, which only has a $90.67\%$ outlier rate, GORE requires over $3$ hours for alignment, while TEASER takes up to $26.84$ seconds. In contrast, Ours achieves registration in a mere $0.342$ seconds. Furthermore, Fig.~\ref{fig_1}(a) shows another registration case for the scan pair s8-s7, where Ours not only achieves better accuracy but also is approximately $10^3$ times faster than GORE and about $4$ times faster than TEASER. Overall, Ours shows higher efficiency compared to GORE and TEASER, which exhibit similar levels of robustness and accuracy as Ours.

On the other hand, non-deterministic RANSAC demonstrates unstable performance, occasionally generating unsatisfactory solutions with significant registration errors, as observed in pairs s1-s0, s2-s0, s4-s2, and s9-s7. Moreover, RANSAC is also time-consuming in these practical scenarios with high outlier rates and a large number of correspondences. FGR, while fast for all scan pairs, often converges to erroneous results. Although GC-RANSAC outperforms RANSAC in terms of stability and efficiency, it still struggles to register all scan pairs successfully. In contrast, Ours exhibits remarkable robustness to the repeated features of outdoor point clouds, achieving a $100\%$ registration success rate on the Bremen dataset. The complete visualization result of the proposed registration method on the Bremen dataset is given in Fig.~\ref{fig_10}.

\subsubsection{ETH dataset experiments}
The ETH dataset\cite{theiler2014keypoint} is a challenging large-scale LiDAR dataset that encompasses five distinct scenarios: Arch, Courtyard, Facade, Office, and Trees. The average overlap rates of these scenarios are $30-40\%, 40-70\%, 60-70\%, \textgreater80\%, \approx50\%$, respectively, as reported in \cite{10091912}. Its registration difficulty comes from low overlap and duplicate features. To ensure the generality of the registration algorithm, we select two scan pairs from each scenario for our registration experiments. The ETH dataset provides ground truth information regarding the relative pose, enabling accurate evaluation. We follow the same data preparation strategy outlined in the last section to establish the initial correspondence set $\mathcal{K}$. For the ETH dataset, the down-sampling resolution and the inlier threshold are both set to $0.1\si{\metre}$. Detailed information about the ETH dataset can be found in Appendix D. The outlier rate in the ETH dataset ranges from approximately $86\%$ to $99\%$, with the number of correspondences varying from around $1k$ to $15k$. To evaluate the registration performance, we compare Ours, GORE, RANSAC, TEASER, FGR, and GC-RANSAC using a total of 10 scan pairs from the ETH dataset.

\begin{figure}
 \centering
 \begin{tabular}{c c}
    \begin{minipage}[b]{0.45\columnwidth}
    \centering
    \raisebox{-.5\height}{\includegraphics[width=\linewidth]{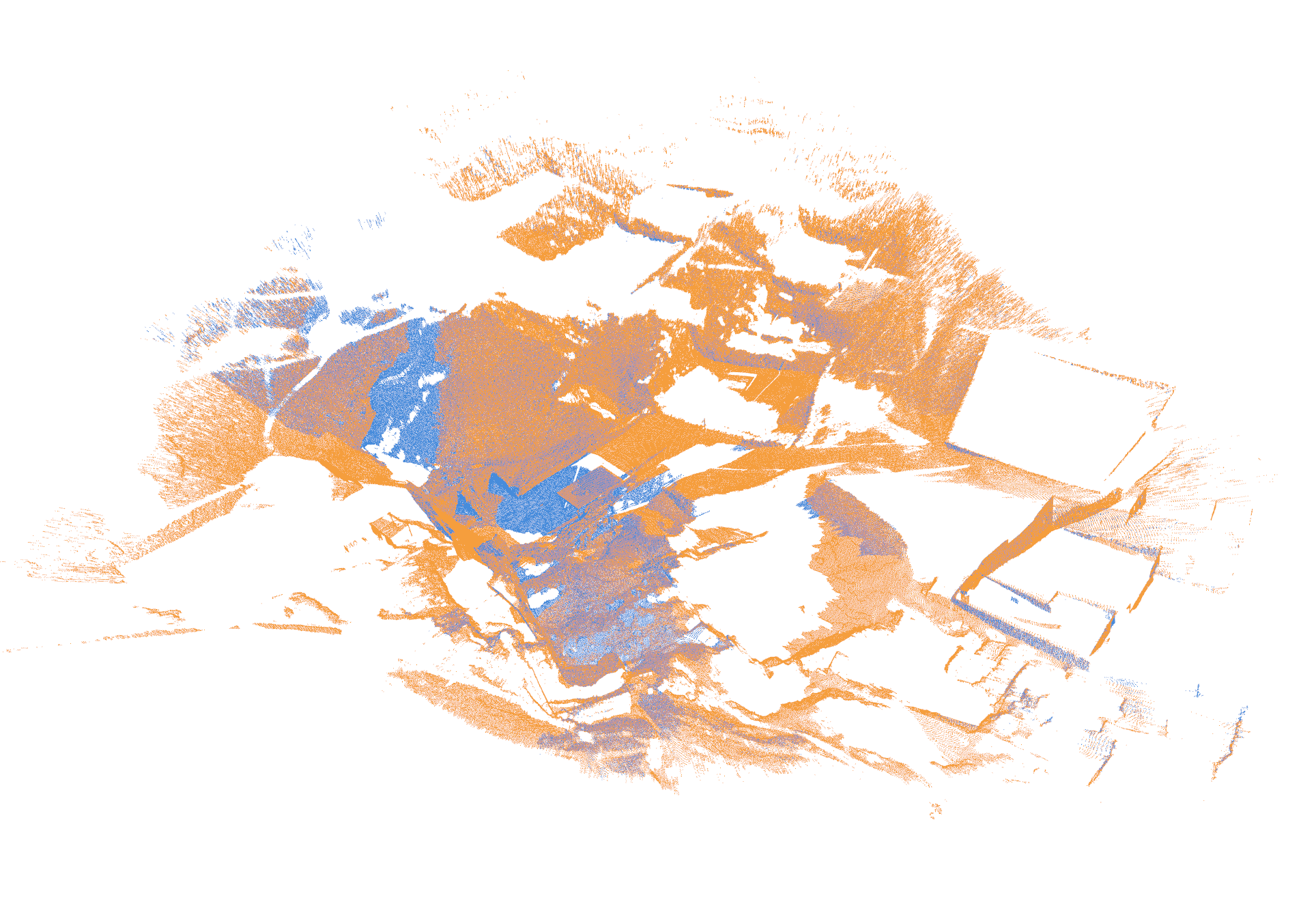}}
    \end{minipage}   

 &  \begin{minipage}[b]{0.45\columnwidth}
    \centering
    \raisebox{-.5\height}{\includegraphics[width=\linewidth]{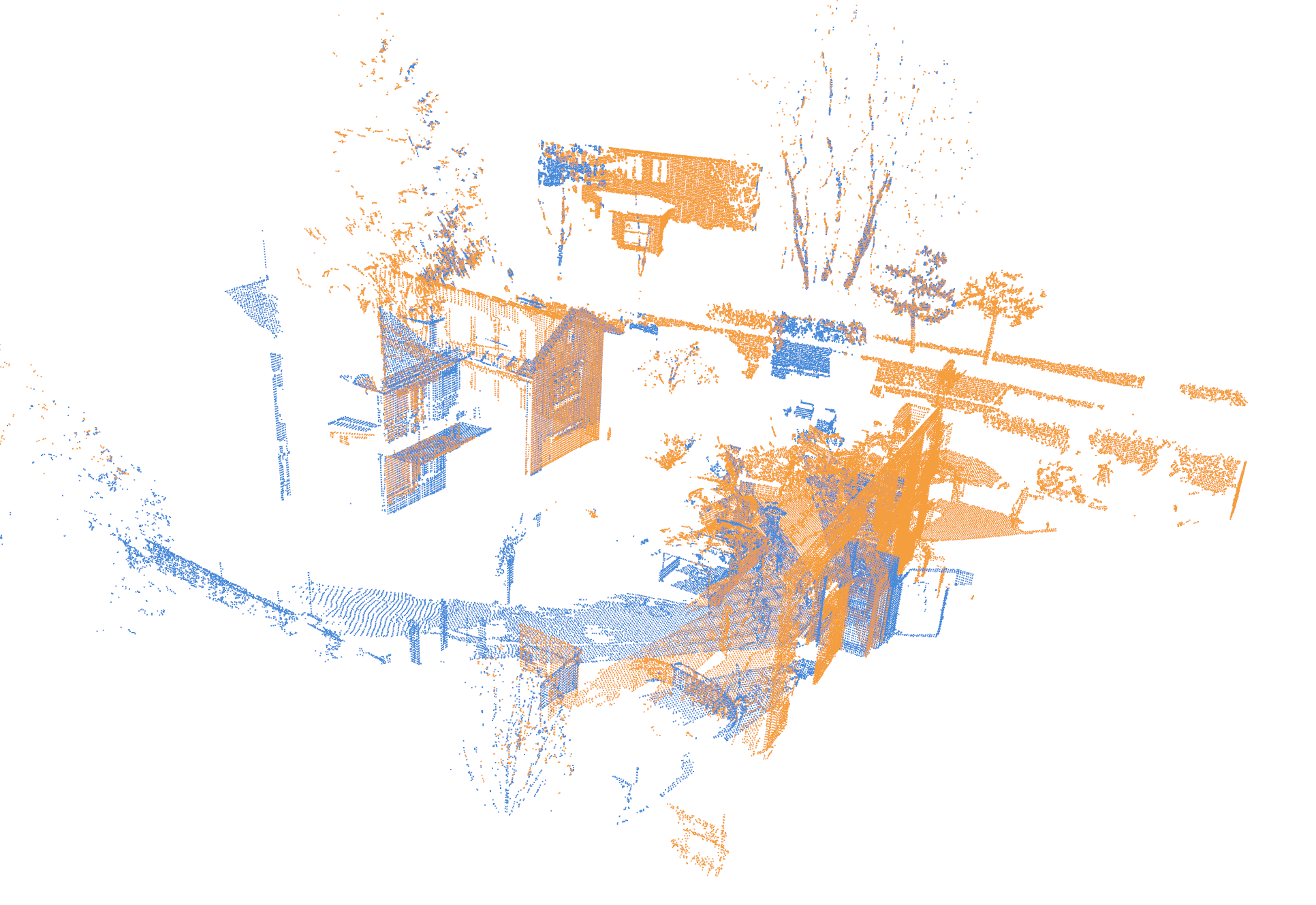}}
    \end{minipage}
 \\ \footnotesize(a) Courtyard1
 &  \footnotesize(b) Facade1
 \\ 
    \begin{minipage}[b]{0.45\columnwidth}
    \centering
    \raisebox{-.5\height}{\includegraphics[width=\linewidth]{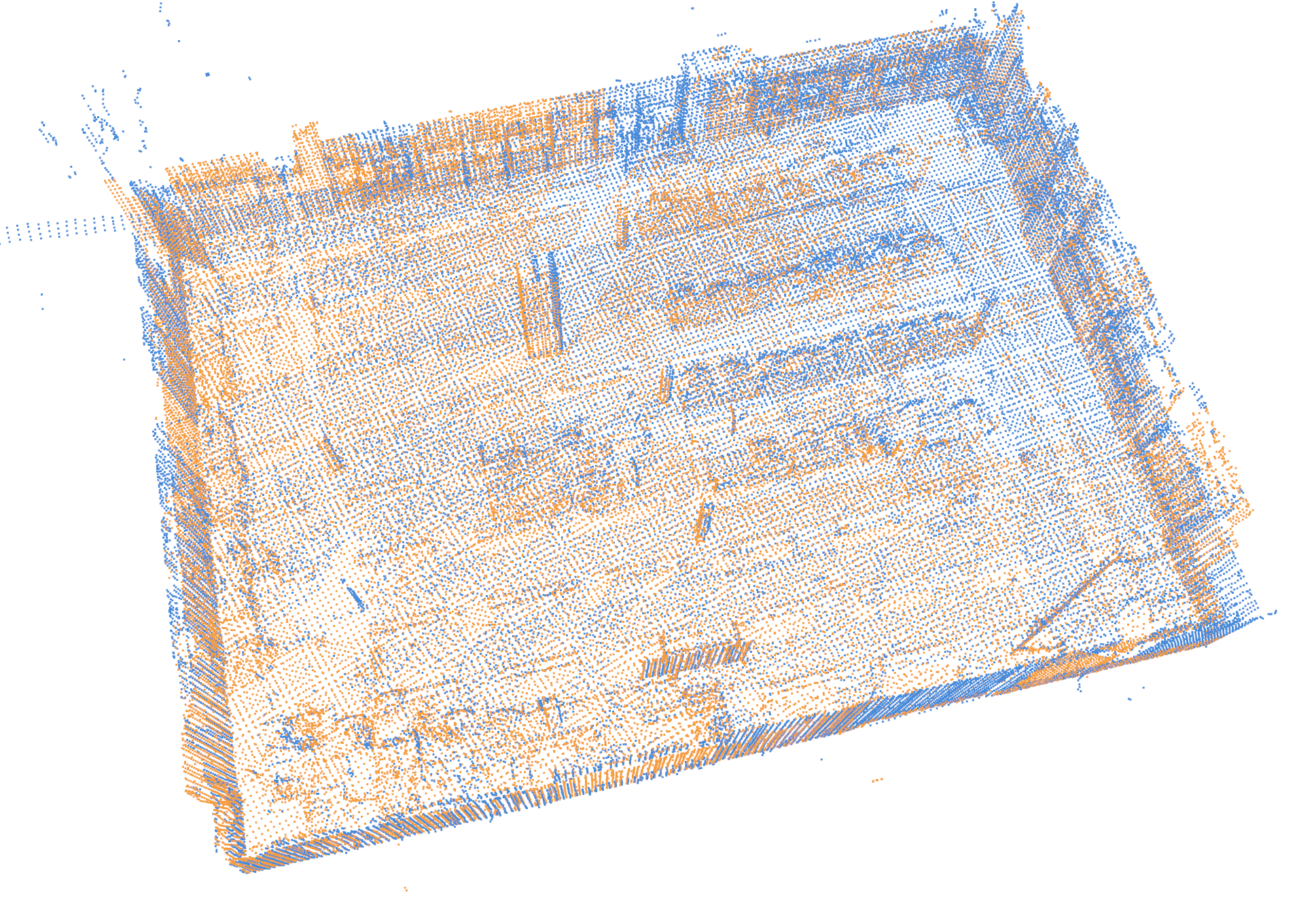}}
    \end{minipage}   
 &  \begin{minipage}[b]{0.45\columnwidth}
    \centering
    \raisebox{-.5\height}{\includegraphics[width=\linewidth]{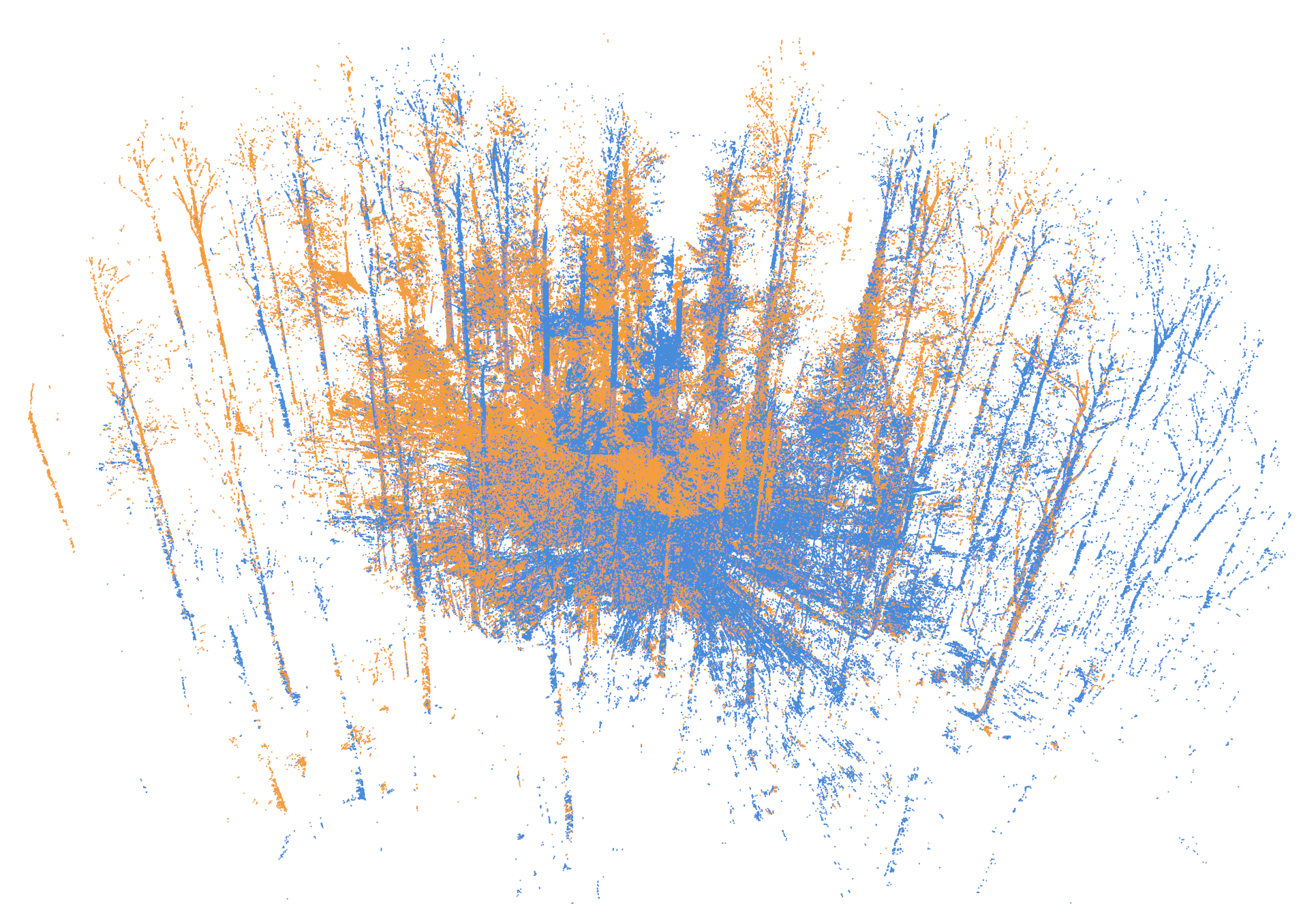}}
    \end{minipage}
 \\ \footnotesize(c) Office1
 &  \footnotesize(b) Trees1

 \end{tabular}
 \caption{Qualitative results of the proposed method on the ETH dataset\cite{theiler2014keypoint}, including four scan pairs: (a) Courtyard1, (b) Facade1, (c) Office1, and (d) Trees1. The aligned source point cloud is blue, and the target point cloud is yellow.}
 \label{fig_12}
\end{figure}

Fig.~\ref{fig_11} reports the rotation error, translation error, and running time for each method evaluated on the ETH dataset. Ours, GORE, and TEASER achieve remarkable robustness over all five scenes, successfully registering all scan pairs. GORE exhibits better accuracy overall compared to Ours, although it is time-consuming. Nevertheless, the registration errors achieved by Ours are still acceptable for practical applications. While the overall accuracy of TEASER is lower than that of Ours and GORE, its time cost increases significantly when dealing with a large number of inliers. For instance, Ours is approximately $280$ times faster than TEASER in aligning the scan pair Courtyard1 with an outlier rate of $86.55\%$, and about $42$ times faster than TEASER in aligning the scan pair Courtyard2 with an outlier rate of $90.62\%$. Another registration case for the scan pair Arch1 (with an outlier rate of $98.45\%$) is illustrated in Fig.~\ref{fig_1}(b), where Ours achieves the lowest translation error and is roughly $480$ times faster than GORE and about $4$ times faster than TEASER. Qualitative results of the proposed method for the remaining four scenarios are provided in Fig.~\ref{fig_12}. Similar to the results in the last section, FGR and GC-RANSAC demonstrate relatively high efficiency but exhibit instability when registering scan pairs with high outlier rates, such as Office2 and Trees1. RANSAC is not only time-consuming on the ETH dataset but also prone to producing incorrect registration results. In summary, benefiting from the proposed pose decoupling strategy, our registration method is more efficient than the SOTA methods while maintaining comparable robustness against low overlap and high outlier rates. 

\subsubsection{KITTI dataset experiments}
\begin{table}\footnotesize
\setlength{\tabcolsep}{5pt} 
\renewcommand{\arraystretch}{1.2} 
\caption{Experiment results on the KITTI dataset\cite{geiger2012we} with FCGF~\cite{choy2019fully} descriptors. Bolded and underlined fonts indicate the first two best values.
\label{table4}}
\centering
\begin{tabular}{c c c c c c}
\hline
Method & {$SR$(\%)} & {$E_{\bm{R}}$(\degree)} & {$E_{\bm{t}}$(cm)} & {$F1$(\%)} & {Time(s)}\\
\hline
\textit{i) Traditional} &~&~&~&~&~\\
{RANSAC\cite{fischler1981random}} & 96.40 & 0.36 & 21.12 & 84.77 & 2.56\\
{TEASER\cite{yang2020teaser}}  & 95.50 & 0.33 & 22.38 & 85.77 & 31.5\\ 
{FGR\cite{zhou2016fast}} & 96.94 & 0.34 & 19.69 & 85.80 & 0.99\\
{GC-RANSAC\cite{barath2021graph}} & 97.48 & \underline{0.32} & 20.68 & 85.42 & 1.16\\
{TR-DE\cite{chen2022deterministic}} & \textbf{98.20} & 0.38 & \textbf{18.00} & \underline{85.99} & 3.01\\
{SC$^2$-PCR\cite{chen2022sc2}} & \textbf{98.20} & 0.33 & 20.95 & 85.90 & \underline{0.31}\\
{MAC\cite{zhang20233d}} & 97.84 & 0.34 & \underline{19.34} & - & -\\
\hline
\textit{ii) Deep learned} &~&~&~&~&~\\
{DGR\cite{choy2020deep}} & 95.14 & 0.43 & 23.28 & 73.60 & 0.86\\
{PointDSC\cite{bai2021pointdsc}} & 97.84 & 0.33 & 20.99 & 85.29 & \underline{0.31} \\
{Hunter\cite{10246849}} & \textbf{98.20} & \textbf{0.29} & 20.35 & - & \textbf{0.14} \\
\hline
{Ours} & \textbf{98.20} & \underline{0.32} & 20.05 & \textbf{86.40} & 0.62\\\hline
\end{tabular}
\end{table}

KITTI\cite{geiger2012we} is an outdoor driving scenarios dataset captured with an onboard 64-beam LiDAR. The point clouds are contaminated with large noise as they are collected in a dynamic scene. Following the data preparation strategy in~\cite{chen2022deterministic,bai2021pointdsc,choy2019fully}, we evaluate the performance of the proposed method on the KITTI dataset. The initial correspondences are generated using the learning-based descriptor FCGF\cite{choy2019fully}, and the inlier threshold is set to $0.6\si{\metre}$. For successful registration, we set the thresholds for rotation error ($E_{\bm{R}}$) and translation error ($E_{\bm{t}}$) to $5\degree$ and $0.6\si{\metre}$, respectively. In addition to comparing the performance of Ours against traditional methods such as RANSAC, TEASER, FGR, GC-RANSAC, TR-DE, SC$^2$-PCR, and MAC, we also compare it with learning-based outlier rejection methods, including DGR, PointDSC, and Hunter. Notably, the learning-based descriptor FCGF outperforms traditional descriptors, resulting in a relatively low outlier rate for FCGF-based correspondences (approximately $58.7\%$ on average). Consequently, GORE is significantly slow on the KITTI dataset, so we do not report its results. 

As shown in Table~\ref{table4}, all methods achieve a success rate exceeding $95\%$ owing to the low outlier rate of FCGF-based correspondences. Among these methods, Ours attains the best success rate of $98.20\%$, as well as TR-DE, SC$^2$-PCR, and Hunter. Although our method is not the fastest, it performs with competitive efficiency compared to SC$^2$-PCR and learning-based methods. Moreover, Ours is approximately $5$ times faster than the SOTA BnB-based TR-DE, about $4$ times faster than the non-deterministic RANSAC, and approximately $50$ times faster than the deterministic TEASER. It is worth mentioning that the most efficient method is the learning-based registration method, Hunter. However, learning-based methods often require additional training procedures and may perform well only on the datasets they were trained on. Additionally, Ours exhibits the second best rotation accuracy and the best $F1$-score. In Fig.~\ref{fig_1}(c), qualitative results of registering a selected pair from the KITTI dataset is provided, where Ours has better accuracy and efficiency than both FGR and GC-RANSAC. In general, when compared to SOTA methods, including learning-based estimators, our method showcases competitive performance in both efficiency and robustness. This underscores the effectiveness of the proposed pose decoupling strategy and the deterministic BnB-based search method.

In addition, an evaluation and verification of the orthogonality of the estimated coarse results is conducted in Appendix F. The results reveal that the proposed method can partially assess the quality of the estimation results by checking the orthogonality and determinant of the coarse rotation matrix.

\section{Conclusion}\label{Conclusion}
In this paper, we present an efficient and deterministic point cloud registration method, leveraging a novel pose decoupling strategy. By utilizing $L_\infty$ residual projections, we successfully decouple the initial registration problem into three sub-problems, resulting in improved efficiency. Furthermore, we introduce a step-wise deterministic search strategy based on branch and bound for these sub-problems. Specifically, we define the inlier set maximization objective function and derive the novel upper bound based on the interval stabbing technique. Benefit from interval stabbing, we can further reduce the dimensionality of the branching space, thus accelerating the BnB search while maintaining robustness. Interestingly, thanks to its significant robustness, our proposed method can be extended to solve the challenging SPCR problem by introducing the interval merging technique. Extensive experiments conducted on both synthetic and real-world datasets demonstrate the competitive performance of our proposed method in terms of efficiency and robustness when compared to SOTA approaches. 

\ifCLASSOPTIONcaptionsoff
  \newpage
\fi

\bibliographystyle{IEEEtran}
\bibliography{reference}

\end{document}